\RequirePackage{fix-cm}
\documentclass[twocolumn]{svjour3}

\smartqed  

\usepackage{times}

\usepackage{epsfig}
\usepackage{graphicx}
\usepackage[belowskip=-10pt,aboveskip=2pt,font=small]{caption}
\usepackage{dblfloatfix}

\usepackage{amsmath}
\usepackage{nccmath}
\usepackage{amssymb}

\usepackage{multirow}
\usepackage{rotating}
\usepackage{booktabs}
\usepackage{slashbox} 
\usepackage{algpseudocode}
\usepackage{algorithm}

\usepackage{enumerate}
\usepackage[olditem,oldenum]{paralist}

\usepackage[pagebackref=true,breaklinks=true,letterpaper=true,colorlinks,bookmarks=false,citecolor=blue,linkcolor=green]{hyperref}

\usepackage{color}

\usepackage[utf8]{inputenc} 
\usepackage[T1]{fontenc}    
\usepackage[pagebackref=true,breaklinks=true,colorlinks,bookmarks=false,citecolor=blue]{hyperref} 
\usepackage{url}            
\usepackage{amsfonts}       
\usepackage{nicefrac}       
\usepackage{microtype}      
\usepackage{subcaption}
\usepackage[font=footnotesize,labelfont=footnotesize]{caption}
\usepackage{comment}
\usepackage{mysymbols}
\belowdisplayskip \abovedisplayskip

\newlength{\sectionReduceTop}
\newlength{\sectionReduceBot}
\newlength{\subsectionReduceTop}
\newlength{\subsectionReduceBot}
\newlength{\abstractReduceTop}
\newlength{\abstractReduceBot}
\newlength{\captionReduceTop}
\newlength{\captionReduceBot}
\newlength{\subsubsectionReduceTop}
\newlength{\subsubsectionReduceBot}

\newlength{\eqnReduceTop}
\newlength{\eqnReduceBot}

\newlength{\horSkip}
\newlength{\verSkip}

\newlength{\figureHeight}
\definecolor{orange}{RGB}{225, 90, 0}
\definecolor{teal}{RGB}{5, 210, 150}
\definecolor{yellow}{RGB}{220, 210, 10}
\definecolor{purple}{RGB}{100, 0, 205}

\newcommand{\gb}{Guided Backpropagation}
\newcommand{\cgb}{Guided Grad-CAM}
\newcommand{\dec}{Deconvolution}
\newcommand{\cdec}{Deconvolution Grad-CAM}
\newcommand{\gcam}{Grad-CAM}
\newcommand{\refsec}[1]{Sec.~\ref{#1}}
\newcommand{\secref}[1]{Sec.~\ref{#1}}
\newcommand{\reffig}[1]{Fig.~\ref{#1}}
\newcommand{\figref}[1]{Fig.~\ref{#1}}

\newcommand{\reftab}[1]{Table.~\ref{#1}}
\newcommand{\rp}[1]{\textcolor{black}{#1}}
\newcommand{\rpi}[1]{\textcolor{black}{#1}}

\newcommand{\iccv}[1]{\textcolor{black}{#1}}
\newcommand{\ijcv}[1]{\textcolor{black}{#1}}

\newcommand{\rpr}[1]{\textcolor{black}{#1}}
\newcommand{\ad}[1]{\textcolor{black}{#1}}
\newcommand{\review}[1]{\textcolor{black}{#1}}
\newcommand{\mac}[1]{\textcolor{black}{#1}} 
\newcommand{\rama}[1]{\textcolor{black}{#1}}
\newcommand{\viceversa}{\textit{vice versa}\/}
\newcommand{\what}[1]{\emph{what}}
\newcommand{\where}[1]{\emph{where}}
\newcommand{\whatcent}[1]{\emph{what}-centric}
\newcommand{\wherecent}[1]{\emph{where}-centric}
\newcommand{\Whatcent}[1]{\emph{What}-centric}
\newcommand{\Wherecent}[1]{\emph{Where}-centric}
\newcommand{\para}[1]{\textbf{#1}.}
\usepackage[toc,page]{appendix}
\renewcommand\appendixtocname{Appendix}
\renewcommand\appendixpagename{Appendix}

\begin{document}
\twocolumn
\title{Grad-CAM: Visual Explanations from Deep Networks\\
via Gradient-based Localization
}

\author{Ramprasaath R. Selvaraju
        \and Michael Cogswell
        \and Abhishek Das
        \and Ramakrishna Vedantam
        \and Devi Parikh
        \and Dhruv Batra
}


\institute{Ramprasaath R. Selvaraju \at
              Georgia Institute of Technology, Atlanta, GA, USA\\
              \email{ramprs@gatech.edu}           
           \and
           Michael Cogswell \at
              Georgia Institute of Technology, Atlanta, GA, USA\\
              \email{cogswell@gatech.edu}
              \and
           Abhishek Das \at
              Georgia Institute of Technology, Atlanta, GA, USA\\
              \email{abhshkdz@gatech.edu}
              \and
           Ramakrishna Vedantam \at
              Georgia Institute of Technology, Atlanta, GA, USA\\
              \email{vrama@gatech.edu}
              \and
           Devi Parikh \at
              Georgia Institute of Technology, Atlanta, GA, USA\\
              Facebook AI Research, Menlo Park, CA, USA\\
              \email{parikh@gatech.edu}
              \and
           Dhruv Batra \at
              Georgia Institute of Technology, Atlanta, GA, USA\\
              Facebook AI Research, Menlo Park, CA, USA\\
              \email{dbatra@gatech.edu}
}


\date{}



\maketitle

\makeatletter
\def\blfootnote{\xdef\@thefnmark{}\@footnotetext}
\makeatother
\begin{abstract}

\label{sec:abstract}
We propose a technique for producing `visual explanations' for decisions from a large class of Convolutional Neural Network (CNN)-based models, making them more transparent and explainable.

Our approach -- Gradient-weighted Class Activation Mapping (Grad-CAM), uses the gradients of any target concept (say `dog' in a classification network or a sequence of words in captioning network) flowing into the final convolutional layer to produce a coarse localization map highlighting the important regions in the image for predicting the concept.

Unlike previous approaches, \gcam{} is applicable to a wide variety of CNN model-families: 
(1) CNNs with fully-connected layers (\eg VGG), 
(2) CNNs used for structured outputs (\eg captioning), 
(3) CNNs used in tasks with multi-modal inputs (\eg visual question answering) or reinforcement learning, all \emph{without architectural changes or re-training}.
We combine \gcam{} with existing fine-grained visualizations to create a high-resolution class-discriminative visualization, \cgb{},
and apply it to image classification,
image captioning, and visual question answering (VQA) models, including ResNet-based architectures.

In the context of image classification models, our visualizations (a) lend insights into failure modes of these models
(showing that seemingly unreasonable predictions have reasonable explanations),
(b) \mac{outperform previous methods on the ILSVRC-15 weakly-supervised localization task,}
(c) are robust to adversarial perturbations,
(d) are more faithful to the underlying model,
and (e) help achieve model generalization by identifying dataset bias.

For image captioning and VQA, our visualizations show that
even non-attention based models learn to localize discriminative regions of input image.

We devise a way to identify important neurons through \gcam{}
and combine it with neuron names~\cite{netdissect} to provide textual explanations for model decisions.
Finally, we design and conduct human studies to measure if \gcam{} explanations help users
establish appropriate trust in predictions from deep networks and show that \gcam{} helps untrained users
successfully discern a `stronger' deep network from a `weaker' one even when both make identical predictions.
\iccv{Our code is available at {\small \url{https://github.com/ramprs/grad-cam/}}, along with a demo
on CloudCV~\cite{agrawal2015cloudcv}\footnote{\url{http://gradcam.cloudcv.org}}, and a video at \footnotesize{\url{youtu.be/COjUB9Izk6E}}.}

\end{abstract}

\vspace{-15pt}
\section{Introduction}
Deep neural models based on Convolutional Neural Networks (CNNs)
have enabled unprecedented breakthroughs in a variety of computer vision tasks, from image
classification~\cite{krizhevsky_nips12,he_cvpr15}, object detection~\cite{girshick2014rcnn}, semantic segmentation~\cite{long2015fcn} to image captioning~\cite{vinyals_cvpr15,chen2015microsoft,fang2015captions,johnson_cvpr16}, visual question answering~\cite{antol2015vqa,gao2015you,malinowski_iccv15,ren_nips15}
and more recently, visual dialog~\cite{visdial,guesswhat,visdial_rl} and
embodied question answering~\cite{embodiedqa,gordon2017iqa}.
While these models enable superior performance, their lack of decomposability into
\emph{individually intuitive} components makes them hard to interpret~\cite{lipton_arxiv16}.
Consequently, when today's intelligent systems fail, they often fail spectacularly disgracefully
without warning or explanation, leaving a user staring at an incoherent output, wondering why the system did what it did.  %

\emph{Interpretability matters.}
In order to build trust in intelligent systems and move towards their meaningful
integration into our everyday lives,
it is clear that we must build `transparent' models that have the ability to
explain \emph{why they predict what they \rama{predict}}.
Broadly speaking, this transparency and ability to explain is useful at three different stages of Artificial Intelligence (AI) evolution.
First, when AI is significantly weaker than humans and not yet reliably deployable (\eg visual question
answering ~\cite{antol2015vqa}), the goal of transparency and explanations is to identify the failure
modes~\cite{agrawal2016analyzing,hoiem2012diagnosing}, thereby helping researchers focus their
efforts on the most fruitful research directions.
Second, when AI is on par with humans and reliably deployable (\eg, image classification~\cite{karpathy_imagenet}
trained on sufficient data), the goal
is to establish appropriate trust \rama{and confidence in} users.
Third, when AI is significantly stronger than humans (\eg chess or Go~\cite{silver2016mastering}),
the goal of explanations is \rama{in} machine teaching~\cite{JohnsCVPR2015} -- \ie, a
machine teaching a human about how to make better decisions.%

\begin{figure*}[t!]
	\centering
	\begin{subfigure}[t]{0.161\textwidth}
		\centering
		\includegraphics[width=\textwidth]{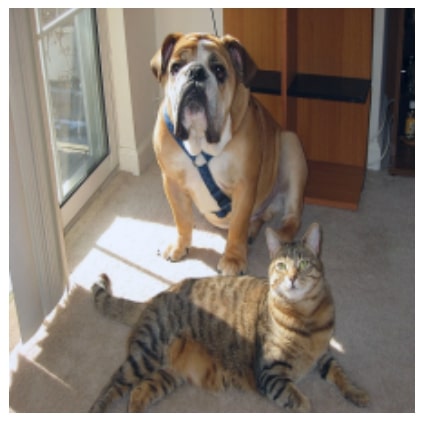}
        \caption{\scriptsize{Original Image}}
	\end{subfigure}
	\begin{subfigure}[t]{0.161\textwidth}
		\centering
		\includegraphics[width=\textwidth]{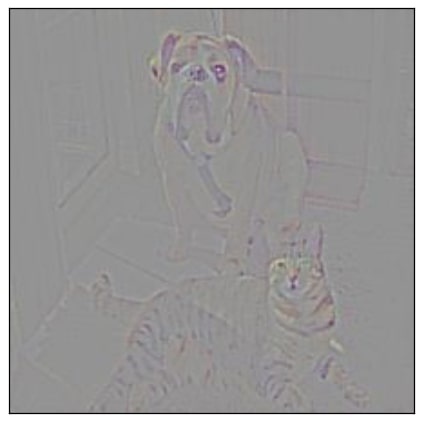}
        \caption{\scriptsize{Guided Backprop `Cat'}}
        \label{fig:teaser_gb_cat}
	\end{subfigure}
	\begin{subfigure}[t]{0.161\textwidth}
		\centering
		\includegraphics[width=\textwidth]{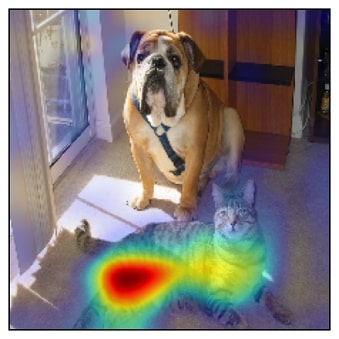}
        \caption{\scriptsize{Grad-CAM `Cat'}}
        \label{fig:teaser_gcam_cat}
	\end{subfigure}
	\begin{subfigure}[t]{0.161\textwidth}
		\centering
		\includegraphics[width=0.995\textwidth]{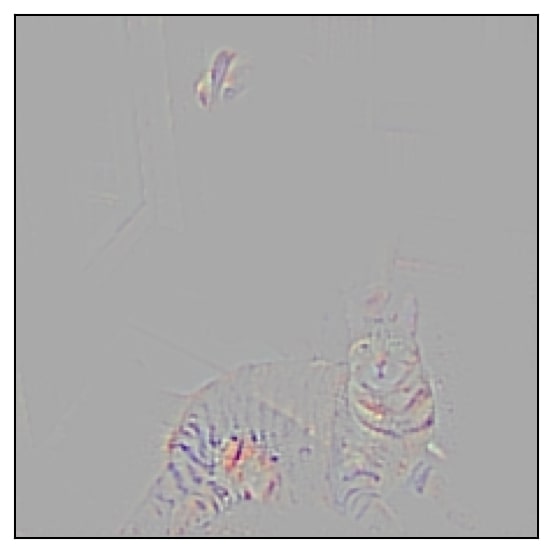}
        \caption{\hspace{-1.5pt}\scriptsize{Guided Grad-CAM `Cat'}}
        \label{fig:teaser_gbgcam_cat}
	\end{subfigure}
	\begin{subfigure}[t]{0.161\textwidth}
		\centering
		\includegraphics[width=\textwidth]{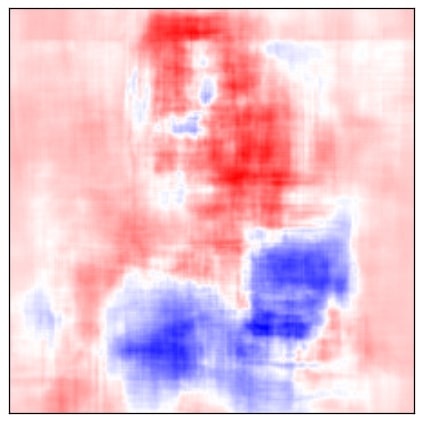}
        \caption{\scriptsize{Occlusion map `Cat'}}
        \label{fig:teaser_cat_occlusion}
	\end{subfigure}
	\begin{subfigure}[t]{0.161\textwidth}
		\centering
		\includegraphics[width=0.995\textwidth]{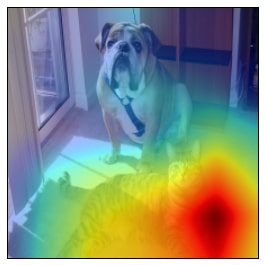}
		\caption{\scriptsize{ResNet \gcam{} `Cat'}}
        \label{fig:teaser_cat_occlusion}
	\end{subfigure}\\
    \vspace*{10pt}
	\begin{subfigure}[t]{0.161\textwidth}
		\centering
        \includegraphics[width=\textwidth]{figures/teaser/original.jpg}
        \caption{\scriptsize{Original Image}\hspace{-12pt}}
	\end{subfigure}
	\begin{subfigure}[t]{0.161\textwidth}
		\centering
        \includegraphics[width=\textwidth]{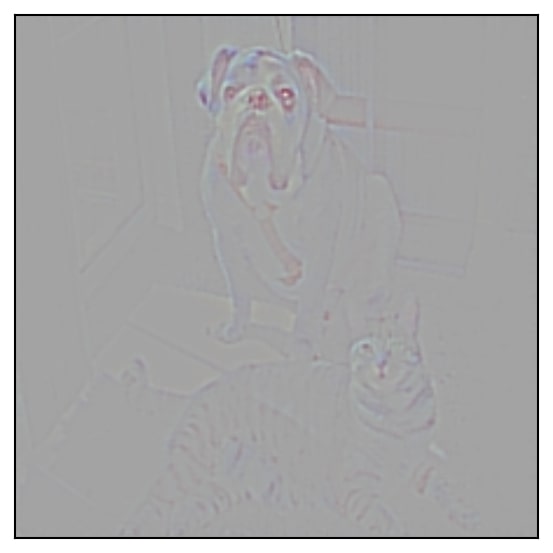}
		\caption{\scriptsize{Guided Backprop `Dog'}\hspace{-12pt}}
        \label{fig:teaser_gb_dog}
	\end{subfigure}
	\begin{subfigure}[t]{0.161\textwidth}
		\centering
        \includegraphics[width=\textwidth]{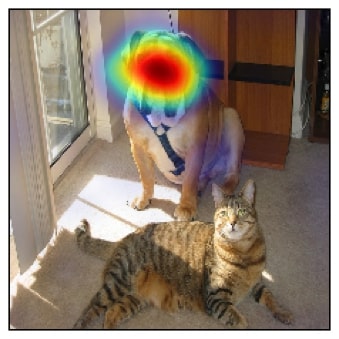}
        \caption{\scriptsize{Grad-CAM `Dog'}\hspace{-12pt}}
        \label{fig:teaser_gcam_dog}
	\end{subfigure}
	\begin{subfigure}[t]{0.161\textwidth}
		\centering
        \includegraphics[width=1.01\textwidth]{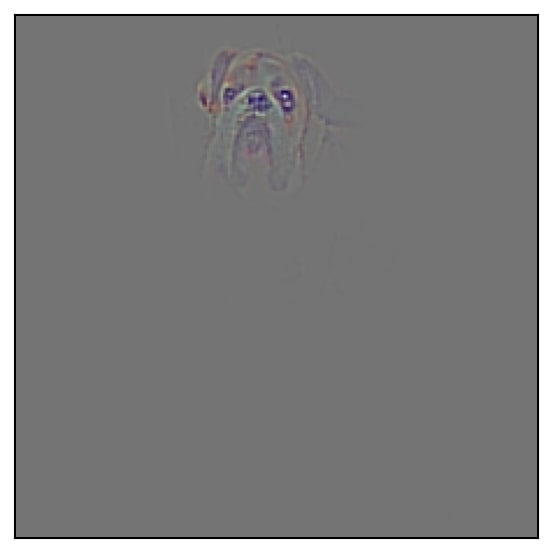}
        \caption{\hspace{-1.9pt}\scriptsize{Guided Grad-CAM `Dog'}\hspace{-12pt}}
        \label{fig:teaser_gbgcam_dog}
	\end{subfigure}
	\begin{subfigure}[t]{0.161\textwidth}
		\centering
        \includegraphics[width=\textwidth]{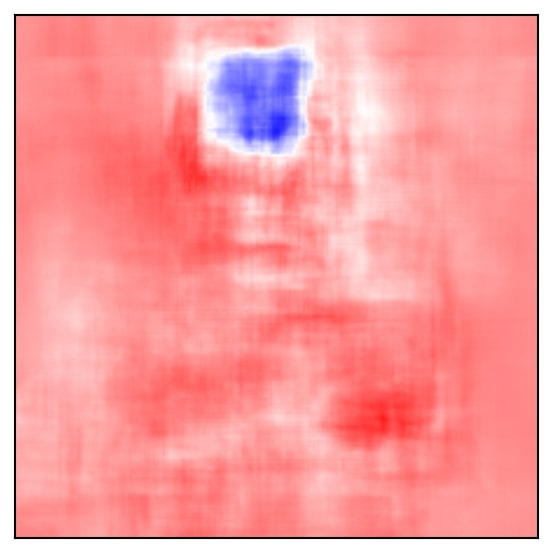}
        \caption{\scriptsize{Occlusion map `Dog'}\hspace{-12pt}}
        \label{fig:teaser_dog_occlusion}
	\end{subfigure}
	\begin{subfigure}[t]{0.161\textwidth}
		\centering
		\includegraphics[width=0.995\textwidth]{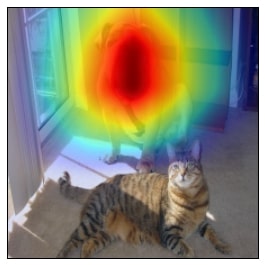}
		\caption{\hspace{-1.9pt}\scriptsize{ResNet \gcam{} `Dog'}}
        \label{fig:teaser_gcam_resnet}
	\end{subfigure}
    \vspace{14pt}
	\caption{
		(a) Original image with a cat and a dog. (b-f) Support for the cat category according to various visualizations \rp{for VGG-16 and ResNet}.
		(b) Guided Backpropagation~\cite{springenberg_arxiv14}: highlights all contributing features. (c, f) \gcam{} (Ours): localizes class-discriminative regions, (d) Combining (b) and (c) gives Guided Grad-CAM, which gives high-resolution class-discriminative visualizations. %
		Interestingly, the localizations achieved by our \gcam{} technique, (c) are very similar to results from occlusion sensitivity (e), while being orders of magnitude cheaper to compute. (f, l) are \gcam{} visualizations for ResNet-18 layer.
		Note that in (c, f, i, l), red regions corresponds to high score for class, while in (e, k), blue corresponds to evidence for the class.
		Figure best viewed in color.
	  }
    \label{fig:teaser}
\end{figure*}

There typically exists a trade-off between accuracy and simplicity or interpretability.
Classical rule-based or expert systems~\cite{jackson_expertsys} are highly interpretable but not very accurate (or robust).
Decomposable pipelines where each stage is hand-designed are thought to be more interpretable as each individual component assumes a natural intuitive explanation.
By using deep models, we sacrifice interpretable modules for uninterpretable ones that achieve greater performance through greater abstraction (more layers) and tighter integration (end-to-end training). Recently introduced deep residual networks (ResNets)~\cite{he_cvpr15} are over 200-layers deep and have shown state-of-the-art performance in several challenging tasks. \rama{
Such complexity makes these models hard to interpret.}
As such, deep models are beginning to explore the spectrum between interpretability and accuracy.

\rpr{Zhou~\etal~\cite{zhou_cvpr16} recently proposed a technique called Class Activation Mapping (CAM) for identifying discriminative regions used by a restricted class of image classification CNNs which do not contain any fully-connected layers.
  In essence, this work trades off model complexity and performance for more transparency into the working of the model.
In contrast, we make existing state-of-the-art deep models interpretable without altering their architecture, thus \rama{avoiding the interpretability \vs accuracy trade-off}.
Our approach is a generalization of CAM~\cite{zhou_cvpr16} and is applicable to a significantly broader range of CNN model families: (1) CNNs with fully-connected layers (\eg VGG), (2) CNNs used for structured outputs (\eg captioning), (3) CNNs used in tasks with multi-modal inputs (\eg VQA) or reinforcement learning, without requiring architectural changes or re-training. %
}

\noindent \textbf{What makes a good visual explanation?}
Consider image classification~\cite{imagenet_cvpr09} -- a `good’ visual explanation from the model for justifying any target category should be (a) class-discriminative (\ie localize the category in the image) and (b) high-resolution (\ie capture fine-grained detail).

Fig.~\hyperlink{page.2}{1} shows outputs from a number of visualizations for the `tiger cat' class (top) and `boxer' (dog) class (bottom).
\mac{Pixel-space gradient visualizations such as \gb{}~\cite{springenberg_arxiv14}  and \dec{}~\cite{zeiler_eccv14}
are high-resolution and highlight fine-grained details in the image, but are not class-discriminative
(\figref{fig:teaser_gb_cat} and \figref{fig:teaser_gb_dog} are very similar).}

\mac{In contrast, localization approaches like CAM or our proposed method Gradient-weighted Class Activation Mapping (\gcam{}),
are highly class-discriminative
(the `cat' explanation exclusively highlights the `cat' regions but not `dog' regions in \figref{fig:teaser_gcam_cat}, and \viceversa{} in \figref{fig:teaser_gcam_dog}).
}

In order to combine the best of both worlds, we show that it is possible to fuse existing pixel-space gradient visualizations with \gcam{} to create \cgb{} visualizations that are both high-resolution and class-discriminative. As a result, important regions of the image which correspond to any decision of interest are visualized in high-resolution detail even if the image contains \rp{evidence for multiple possible} \rama{concepts}, as shown in Figures \hyperlink{page.2}{1d} and \hyperlink{page.2}{1j}.
When visualized for `tiger cat', \cgb{} not only highlights the cat regions, but also highlights the stripes on the cat, which is important for predicting that particular variety of cat.

\noindent To summarize, our contributions are as follows:

\noindent \textbf{(1)}
\mac{We introduce \gcam{}, a class-discriminative localization technique that generates
visual explanations for \emph{any} CNN-based network without requiring architectural changes or re-training.
We evaluate \gcam{} for localization (\secref{sec:localization}), and faithfulness to model (\secref{sec:occ}), where it outperforms baselines.}

\noindent \textbf{(2)}
\mac{We apply \gcam{} to existing top-performing classification, captioning (\secref{sec:nic}), and VQA (\secref{sec:vqa}) models.}
For image classification, our visualizations lend insight into failures of current CNNs (\secref{sec:diagnose}), showing that seemingly unreasonable predictions have reasonable explanations.
For captioning and VQA, our visualizations expose
that common \mac{CNN + LSTM models are often surprisingly good at}
localizing discriminative image regions despite not being trained on
grounded image-text pairs.

\noindent \textbf{(3)}
We show a proof-of-concept of how interpretable \gcam{} visualizations
help in diagnosing failure modes by uncovering biases in datasets.
This is important not just for generalization, but also for fair and
bias-free outcomes as more and more decisions are made by algorithms in society.

\noindent \textbf{(4)}
We present Grad-CAM visualizations for ResNets~\cite{he_cvpr15} applied to
image classification and VQA (\secref{sec:vqa}).

\ijcv{\noindent \textbf{(5)}
We use neuron importance from \gcam{} and neuron names from~\cite{netdissect} and obtain textual explanations for model decisions (\secref{sec:text_exp}).}

\noindent \textbf{(6)}
We conduct human studies (\secref{sec:human_evaluation}) \mac{that show \cgb{} explanations are class-discriminative and not only help humans establish trust,
but also help untrained users successfully discern a `stronger' network from a `weaker' one,
\emph{even when both make identical predictions.}}

\rpi{\noindent \textbf{Paper Organization: }
The rest of the paper is organized as follows. 
In section 3 we propose our approach \gcam{} and \cgb{}. 
In sections 4 and 5 we evaluate the localization ability, class-discriminativeness, trustworthyness and faithfulness of \gcam{}. 
In section 6 we show certain use cases of \gcam{} such as diagnosing image classification CNNs and identifying biases in datasets. 
In section 7 we provide a way to obtain textual explanations with \gcam{}. 
In section 8 we show how \gcam{} can be applied to vision and language models -- image captioning and Visual Question Answering (VQA).}

\begin{figure*}
    \begin{center}
  \centering
  \includegraphics[width=\textwidth]{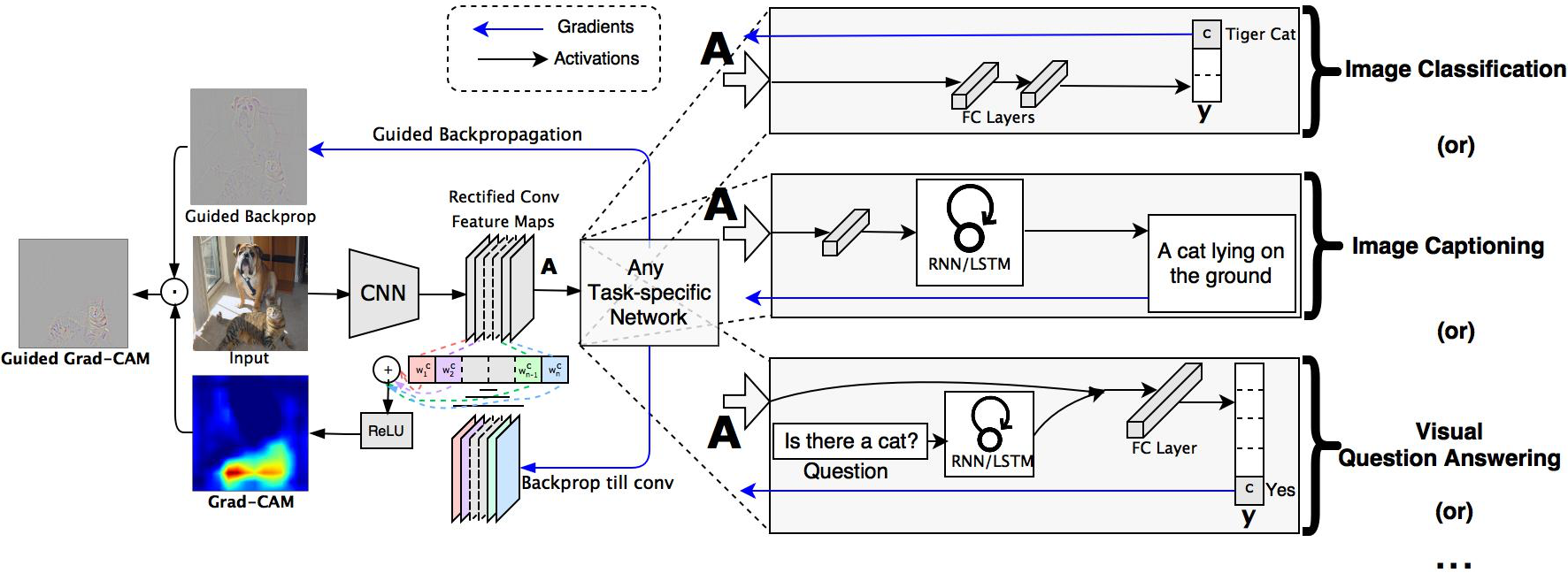}
	\caption{\scriptsize  \gcam{} overview: Given an image \mac{ and a class of interest (\eg, `tiger cat' or any other type of differentiable output) as input, we forward propagate the image through the CNN part of the model and then through task-specific computations to obtain a raw score for the category.}
        The gradients are set to zero for all classes except the desired class (tiger cat), which is set to 1.
		This signal is then backpropagated to the rectified convolutional \mac{feature maps of interest, which we combine to compute} the coarse \gcam{} localization (blue heatmap) \rp{which represents where the model has to look to make the particular decision}.
	Finally, we pointwise multiply the heatmap with guided backpropagation to get \cgb{} visualizations which are both high-resolution and concept-specific.
}
	\label{fig:approach}
\end{center}
\end{figure*}

\vspace{-15pt}
\section{Related Work}

Our work draws on recent work in CNN visualizations, model trust assessment, and weakly-supervised localization.


\noindent \para{Visualizing CNNs}
A number of previous works~\cite{simonyan_arxiv13,springenberg_arxiv14,zeiler_eccv14,Gan_2015_CVPR} have visualized CNN predictions by highlighting `important' pixels (\ie change in intensities of these pixels have the most impact on the prediction score).
Specifically, Simonyan \etal~\cite{simonyan_arxiv13} visualize partial derivatives of predicted class scores \wrt pixel intensities, while \gb{}~\cite{springenberg_arxiv14} and \dec{}~\cite{zeiler_eccv14} make modifications to `raw' gradients that result in qualitative improvements.
\mac{These approaches are compared in ~\cite{mahendran16eccv}.}
Despite producing fine-grained visualizations, these methods are not class-discriminative. Visualizations with respect to different classes are nearly identical (see Figures \hyperlink{page.2}{1b} and \hyperlink{page.2}{1h}).

Other visualization methods synthesize images to maximally activate a network unit~\cite{simonyan_arxiv13,erhan2009visualizing} or invert a latent representation~\cite{mahendran2016visualizing,dosovitskiy_cvpr16}.
Although these can be high-resolution and class-discriminative,
they are not specific
to a single input image and visualize a model overall.

\noindent\para{Assessing Model Trust}
Motivated by notions of interpretability~\cite{lipton_arxiv16} and assessing trust in models~\cite{lime_sigkdd16}, we evaluate \gcam{} visualizations in a manner similar to ~\cite{lime_sigkdd16} via human studies to show that they can be important tools for users to evaluate and place trust in automated systems.

\noindent\para{Aligning Gradient-based Importances}
 Selvaraju \etal \cite{niwt} proposed an approach that uses the gradient-based neuron importances introduced in our work, and maps it to class-specific domain knowledge from humans in order to learn classifiers for novel classes. In future work, Selvaraju \etal \cite{hint} proposed an approach to align gradient-based importances to human attention maps in order to ground vision and language models.

\noindent \para{Weakly-supervised localization}
Another relevant line of work is weakly-supervised localization in the context of CNNs, where the task is to localize objects in images using holistic image class labels only~\cite{cinbis2016weakly,oquab_cvpr14,oquab_cvpr15,zhou_cvpr16}.




Most relevant to our approach is the Class Activation Mapping (CAM) approach to localization~\cite{zhou_cvpr16}.
This approach modifies image classification CNN architectures replacing fully-connected layers with convolutional layers and global average pooling~\cite{lin2013network}, thus achieving class-specific feature maps. 
Others have investigated similar methods using global max pooling~\cite{oquab_cvpr15} and log-sum-exp pooling~\cite{pinheiro2015image}.

A drawback of CAM is that it requires feature maps to directly precede softmax layers, so it is only applicable to a particular kind of CNN architectures performing global average pooling over convolutional maps immediately prior to prediction (\ie conv feature maps $\rightarrow$ global average pooling $\rightarrow$ softmax layer).
Such architectures may achieve inferior accuracies compared to general networks on some tasks (\eg image classification) or may simply be inapplicable to any other tasks (\eg image captioning or VQA).
We introduce a new way of combining feature maps using the gradient signal that does not require \emph{any} modification in the network architecture.
This allows our approach to be applied to off-the-shelf CNN-based architectures,
including those for image captioning and visual question answering.
For a fully-convolutional architecture, CAM is a special case of \gcam{}.


Other methods approach localization by classifying perturbations of the input image.
Zeiler and Fergus~\cite{zeiler_eccv14} perturb inputs by occluding patches and classifying the occluded image, typically resulting in lower classification scores for relevant objects when those objects are occluded.
This principle is applied for localization in \cite{bazzani2016self}. 
Oquab \etal~\cite{oquab_cvpr14} classify many patches containing a pixel then average these patch-wise
scores to provide the pixel's class-wise score.
Unlike these, our approach achieves localization in one shot; it only requires a single forward and a partial backward pass per image and thus is typically an order of magnitude more efficient.
In recent work, Zhang~\etal~\cite{zhang2016top} introduce contrastive Marginal Winning Probability (c-MWP), a probabilistic Winner-Take-All formulation for modelling the top-down attention for neural classification models which can highlight discriminative regions. 
This is computationally more expensive than \gcam{} and only works for
image classification CNNs. 
Moreover, \gcam{} outperforms c-MWP in quantitative and qualitative evaluations (see \refsec{sec:localization} and \refsec{sec:sup_pointing}).

\vspace{-10pt}
\section{\gcam{}}\label{sec:approach}
A number of previous works have asserted that deeper representations in a CNN capture higher-level visual constructs~\cite{bengio2013representation,mahendran2016visualizing}.
Furthermore, convolutional layers naturally retain spatial information which is lost in fully-connected layers, so we can expect the last convolutional layers to have the best compromise between high-level semantics and detailed spatial information.
The neurons in these layers look for semantic class-specific information in the image (say object parts).
\gcam{} uses the gradient information flowing into the last convolutional layer
of the CNN to assign importance values to each neuron for a particular decision of interest.
\rp{
    Although our technique is fairly general in that it can be used to explain
    activations in any layer of a deep network, in this work, we focus on explaining
    output layer decisions only.
}

As shown in \reffig{fig:approach}, in order to obtain the class-discriminative localization map \gcam{}
$L_{\text{\gcam{}}}^c$ $\in \mathbb{R}^{u \times v}$ of width $u$ and height $v$ for any class $c$,
we first compute the gradient
of the score for class $c$, $y^c$ \mac{(before the softmax)}, with respect to feature map
activations $A^k$ of a convolutional layer, \ie $\frac{\del y^c}{\del A^k}$.
These gradients flowing back are global-average-pooled~\footnote{\rpi{Empirically we found global-average-pooling to work better than global-max-pooling as can be found in the Appendix. }}
\ad{
    over the width and height dimensions (indexed by $i$ and $j$ respectively)}
to obtain the neuron importance weights $\alpha{}_{k}^c$:
\begin{ceqn}
\begin{equation} \label{eq:alpha1}
    \alpha{}_{k}^c =
    \overbrace{
        \frac{1}{Z}\sum_{i}\sum_{j}
    }^{\text{global average pooling}}
    \hspace{-17pt}
    \underbrace{
        \vphantom{\sum_{i}\sum_{j}} \frac{\partial y^c}{\partial A_{ij}^{k}}
    }_{\text{gradients via backprop}}
\end{equation}
\end{ceqn}
\rpi{During computation of $\alpha{}_{k}^c$ while backpropagating gradients with respect to activations, the exact computation amounts to successive matrix products of the weight matrices and the gradient with respect to activation functions till the final convolution layer that the gradients are being propagated to. }
Hence, this weight $\alpha{}_{k}^c$ represents a \emph{partial linearization} of the deep network downstream from A,
and captures the `importance' of feature map $k$ for a target class $c$. 

We perform a weighted combination of forward activation maps, and follow it by a ReLU to obtain,
\begin{ceqn}
\begin{equation} \label{eq:gcam}
    L_{\text{\gcam{}}}^{c} = ReLU \underbrace{\left(\sum_k \alpha{}_{k}^{c} A^{k}\right)}_{\text{linear combination}}
\end{equation}
\end{ceqn}
Notice that this results in a coarse heatmap of the same size as the convolutional feature
maps ($14 \times 14$ in the case of last convolutional layers of VGG \cite{simonyan_arxiv14} and AlexNet \cite{krizhevsky_nips12} networks) \footnote{\rpi{We find that \gcam{} maps become progressively worse as we move to earlier convolutional layers as they have smaller receptive fields and only focus on less semantic local features.}}.
\mac{We apply a ReLU to the linear combination of maps because we are only interested in the features that have a \emph{positive} influence on the class of interest, \ie
pixels whose intensity should be \emph{increased} in order to increase $y^c$.
Negative pixels are likely to belong to other categories in the image.
As expected, without this ReLU, localization maps sometimes highlight more than
just the desired class and perform worse at localization.}
Figures \hyperlink{page.2}{1c, 1f} and \hyperlink{page.2}{1i, 1l} show \gcam{} visualizations for `tiger cat'
and `boxer (dog)' respectively. 
Ablation studies are available in ~\secref{sec:ablation}.

\mac{
    In general, $y^c$ need not be the class score produced by an image classification
    CNN. It could be any differentiable activation including words from a caption or
    answer to a question.
}

\vspace{-10pt}
\subsection{\gcam{} generalizes CAM}
\label{sec:generalization}
\ijcv{In this section, we \ad{discuss the connections between \gcam{} and Class Activation Mapping (CAM)~\cite{zhou_cvpr16},
and formally  prove that \gcam{} generalizes CAM} for a wide variety of CNN-based architectures.}
\noindent Recall that CAM produces a localization map for an image classification
CNN with a specific kind of architecture where global average pooled convolutional feature maps
are fed directly into softmax.
Specifically, let the penultimate layer produce $K$ feature maps,
$A^k \in \mathbb{R}^{u \times v}$, with each element indexed by $i,j$.
So $A_{ij}^k$ refers to the activation at location $(i,j)$ of the feature map $A^k$.
These feature maps are then spatially pooled using Global Average Pooling (GAP)
and linearly transformed to produce a score $Y^c$ for each class $c$,
\begin{ceqn}
\begin{equation} \label{eq:scores}
    Y^c = \sum_{k}
	\hspace{-20pt}
    \underbrace{
        \vphantom{\frac{1}{Z}\sum_i \sum_j} w^c_k
    }_{\vspace{-3pt}
	\text{class feature weights}}
    \hspace{-27pt}
    \overbrace{
        \frac{1}{Z}\sum_i \sum_j
    }^{\text{global average pooling}}
    \hspace{-15pt}
    \underbrace{
        \vphantom{\frac{1}{Z}\sum_i \sum_j} A^k_{ij}
    }_{\vspace{-5pt}\text{feature map}}
\end{equation}
\end{ceqn}

Let us define $F^k$ to be the global average pooled output,
\begin{ceqn}
\begin{equation}\label{eq:cam_gap}
    F^{k} = \frac{1}{Z} \sum_{i} \sum_{j} A_{ij}^k
\end{equation}
\end{ceqn}

CAM computes the final scores by,
\begin{ceqn}
\begin{equation}\label{eq:cam_scores}
    Y^c = \sum_{k} w_{k}^c \cdot F^{k}
\end{equation}
\end{ceqn}
where $w_{k}^c$ is the weight connecting the $k^{th}$ feature map with the $c^{th}$ class.
Taking the gradient of the score for class c ($Y^c$)  with respect to the feature map $F^k$ we get,
\begin{ceqn}
\begin{equation}\label{eq:cam_grad}
    \frac{\partial Y^c}{\partial F^k} = \frac{\frac{\partial Y^c}{\partial A_{ij}^k}}{\frac{\partial F^k}{\partial A_{ij}^k}}
\end{equation}
\end{ceqn}

Taking partial derivative of  \eqref{eq:cam_gap} \wrt $A_{ij}^k$, we can see that $\frac{\del F^k}{\del A^k_{ij}} = \frac{1}{Z}$. Substituting this in \eqref{eq:cam_grad}, we get,
\begin{ceqn}
\begin{equation}
    \frac{\partial Y^c}{\partial F^k} = \frac{\partial Y^c}{\partial A_{ij}^k} \cdot Z\\
\end{equation}
\end{ceqn}

From \eqref{eq:cam_scores} we get that, $ \frac{\partial Y^c}{\partial F^k} = w_{k}^c$. Hence,

\begin{ceqn}
\begin{equation}\label{eq:cam_weights}
    w_{k}^c = Z \cdot \frac{\partial Y^c}{\partial A_{ij}^k}\\
\end{equation}
\end{ceqn}

Summing both sides of \eqref{eq:cam_weights} over all pixels $(i,j)$,

\begin{ceqn}
\begin{equation}
    \sum_i \sum_j w^c_k  = \sum_i \sum_j Z \cdot \frac{\del Y^c}{\del A^k_{ij} }
\end{equation}
\end{ceqn}

Since $Z$ and $w^c_k$ do not depend on $(i,j)$, rewriting this as

\begin{ceqn}
\begin{equation}
    Z w^c_k  = Z \sum_i \sum_j \frac{\del Y^c}{\del A^k_{ij}}
\end{equation}
\end{ceqn}

Note that $Z$ is the number of pixels in the feature map (or $Z = \sum_i \sum_j \mathbf{1} $).
Thus, we can re-order terms and see that

\begin{ceqn}
\begin{equation}
    w^c_k  = \sum_i \sum_j \frac{\del Y^c}{\del A^k_{ij} }\\
\end{equation}
\end{ceqn}

Up to a proportionality constant ($1/Z$) that gets normalized-out during
visualization, the expression for $w^c_k$ is identical to $\alpha^c_k$ used by \gcam{} \eqref{eq:alpha1}.
Thus, \gcam{} is a strict generalization of CAM.
This generalization allows us to generate visual explanations from CNN-based
models that cascade convolutional layers with much more complex interactions,
such as those for image captioning and VQA (\refsec{sec:vqa}).

\vspace{-5pt}
\subsection{\cgb{}}
While \gcam{} is class-discriminative and localizes relevant image regions,
it lacks the ability to highlight fine-grained details like pixel-space gradient
visualization methods (\gb{}~\cite{springenberg_arxiv14}, \dec{}~\cite{zeiler_eccv14}).
\ad{\gb{} visualizes gradients with respect to the image where
    negative gradients are suppressed when backpropagating through ReLU layers.
    Intuitively, this aims to capture pixels detected by neurons, not
    the ones that suppress neurons.}
See Figure \hyperlink{page.2}{1c}, where
\gcam{} can easily localize the cat; however, it is unclear from the coarse
heatmap why the network predicts this particular instance as `tiger cat'.
In order to combine the best aspects of both, we fuse \gb{} and \gcam{} visualizations
via element-wise multiplication
($L_{\text{\gcam{}}}^c$ is first upsampled to the input image resolution using bilinear interpolation).
\figref{fig:approach} bottom-left illustrates this fusion.
This visualization is both high-resolution (when the class of interest is `tiger cat',
it identifies important `tiger cat' features like stripes, pointy ears and eyes) and
class-discriminative (it highlights the `tiger cat' but not the `boxer (dog)').
Replacing \gb{} with \dec{} gives similar results, but we found \dec{}
visualizations to have artifacts and \gb{} to be generally less noisy.

\subsection{Counterfactual Explanations}

Using a slight modification to \gcam{}, we can obtain explanations
that highlight support for regions that would make the network change its prediction. 
As a consequence, removing concepts occurring in those regions would make the
model more confident about its prediction. 
We refer to this explanation modality as counterfactual explanations.

\begin{figure}[ht!]
    \centering \begin{subfigure}[t]{0.158\textwidth} \centering
        \includegraphics[width=\textwidth]{figures/teaser/original.jpg}
        \caption{\scriptsize{Original Image}}
	\end{subfigure}
	\begin{subfigure}[t]{0.158\textwidth}
        \centering
        \includegraphics[width=\textwidth]{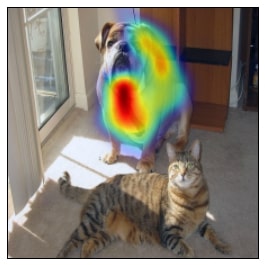}
        \caption{\scriptsize{Cat Counterfactual exp}}
		\label{fig:neg_exp_cat}
	\end{subfigure}
    \centering
	\begin{subfigure}[t]{0.158\textwidth}
        \centering
        \includegraphics[width=\textwidth]{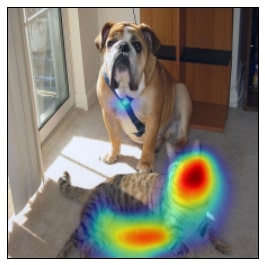}
        \caption{\ijcv{\scriptsize{\hspace{-1pt}Dog Counterfactual exp}}}%
		\label{fig:neg_exp_dog}
	\end{subfigure}
    \vspace{10pt}
    \caption{Counterfactual Explanations with \gcam{}}
    \label{fig:negexp}
\end{figure}

Specifically, we negate the gradient of $y^c$ (score for class $c$) with respect to feature maps $A$ of a convolutional layer.
Thus the importance weights $\alpha{}_{k}^{c}$ now become
\begin{ceqn}
\begin{equation} \label{eq:alpha_neg}
    \alpha{}_{k}^c =
    \overbrace{
        \frac{1}{Z}\sum_{i}\sum_{j}
    }^{\text{global average pooling}}
    \hspace{-17pt}
    \underbrace{
        \vphantom{\sum_{i}\sum_{j}} -\frac{\partial y^c}{\partial A_{ij}^{k}}
    }_{\text{Negative gradients}}
\end{equation}
\end{ceqn}
As in \eqref{eq:gcam}, we take a weighted sum of the forward activation maps, $A$, with weights $\alpha{}_{k}^c$, and follow it by a ReLU to obtain counterfactual explanations as shown in \reffig{fig:negexp}.

\vspace{-10pt}
\section{Evaluating Localization Ability of \gcam{}}
\subsection{Weakly-supervised Localization}\label{sec:localization}
In this section, we evaluate the localization capability of \gcam{} in the context of image classification.
The ImageNet localization challenge~\cite{imagenet_cvpr09} requires
approaches to provide bounding boxes in addition to classification labels.
Similar to classification, evaluation is performed for both the top-1 and top-5 predicted categories.

Given an image, we first obtain class predictions from our network and
then generate \gcam{} maps for each of the predicted classes and binarize them with a
threshold of 15\% of the max intensity.
This results in connected segments of pixels and we draw a bounding box around
the single largest segment.
Note that this is weakly-supervised localization -- the models were never exposed
to bounding box annotations during training.

We evaluate \gcam{} localization with off-the-shelf pretrained VGG-16~\cite{simonyan_arxiv14} \rpi{, AlexNet ~\cite{krizhevsky_nips12} and GoogleNet \cite{szegedy2016rethinking} 
(obtained from the Caffe~\cite{jia2014caffe} Zoo). }
Following ILSVRC-15 evaluation, we report both top-1 and top-5 localization errors
on the val set in \reftab{table:locres}.
\gcam{} localization errors are significantly better than those achieved by
c-MWP~\cite{zhang2016top} and Simonyan~\etal~\cite{simonyan_arxiv13}, which use grab-cut to
post-process image space gradients into heat maps.
\gcam{} for VGG-16 also achieves better top-1 localization error than CAM~\cite{zhou_cvpr16}, which requires a change
in the model architecture, necessitates re-training and thereby achieves worse classification errors (2.98\%
worse top-1), while \gcam{} does not compromise on classification performance.

\begin{table}[h!]
\vspace{-10pt}
\centering
    \resizebox{1.00\columnwidth}{!}{
        \begin{tabular}{l l l c c c c c c}
            & & \multicolumn{2}{c}{\textbf{Classification}} &~~~ & \multicolumn{2}{c}{\textbf{Localization}}\\
            \cmidrule{3-4}\cmidrule{6-7}
            & & \textbf{Top-$1$} & \textbf{Top-$5$} & & \textbf{Top-$1$} & \textbf{Top-$5$} \\
            \midrule
            \multirow{4}{*}{\rotatebox{90}{\tiny{\centering VGG-16}}} 

 &           Backprop~\cite{simonyan_arxiv13}   & $30.38$ & $10.89$ & & $61.12$ & $51.46$        \\
 &           c-MWP~\cite{zhang2016top}          & $30.38$ & $10.89$ & & $70.92$ & $63.04$        \\
&			\gcam{} (ours)                     & $30.38$ & $10.89$ & & $\mathbf{56.51}$ & $46.41$        \\
             \cmidrule{2-7}
&			CAM~\cite{zhou_cvpr16} & $33.40$ & $12.20$ & & $57.20$ & $\mathbf{45.14}$        \\
\midrule
\multirow{2}{*}{\rotatebox{90}{\tiny{\centering AlexNet}}} 
&           c-MWP~\cite{zhang2016top}          & $44.2$ & $20.8$ & & $92.6$ & $89.2$        \\
&			\gcam{} (ours)                     & $44.2$ & $20.8$ & & $68.3$ & $56.6$        \\
\midrule
\multirow{2}{*}{\rotatebox{90}{\tiny{\centering GoogleNet}}} 
&			\gcam{} (ours)                     & $31.9$ & $11.3$ & & $60.09$ & $49.34$        \\
&			CAM~\cite{zhou_cvpr16} 			   & $31.9$ & $11.3$ & & $60.09$ & $49.34$        \\
            \bottomrule
        \end{tabular}
    }
    \vspace{5pt}
    \caption{Classification and localization error \% on ILSVRC-15 val (lower is better) for VGG-16, AlexNet and GoogleNet. We see that \gcam{} achieves superior localization errors without compromising on classification performance.}
    \label{table:locres}
\end{table}

\vspace{-20pt}
\subsection{Weakly-supervised Segmentation}\label{sec:segmentation}

Semantic segmentation involves the task of assigning each pixel in the image an object class (or background class).
Being a challenging task, this requires expensive pixel-level annotation. 
The task of weakly-supervised segmentation involves segmenting objects with just image-level annotation, which can be obtained relatively cheaply from image classification datasets.
In recent work, Kolesnikov~\etal \cite{seed_eccv16} introduced a new loss function for training weakly-supervised image segmentation models.
Their loss function is based on three principles --
1) to seed with weak localization cues, encouraging segmentation network to match these cues,
2) to expand object seeds to regions of reasonable size based on information about which classes can occur in an image,
3) to constrain segmentations to object boundaries that alleviates the problem of imprecise boundaries already at training time.
They showed that their proposed loss function, consisting of the above three losses leads to better segmentation. 

\begin{figure}[htp]
\vspace{-17pt}
 \centering
 \includegraphics[width=1\linewidth]{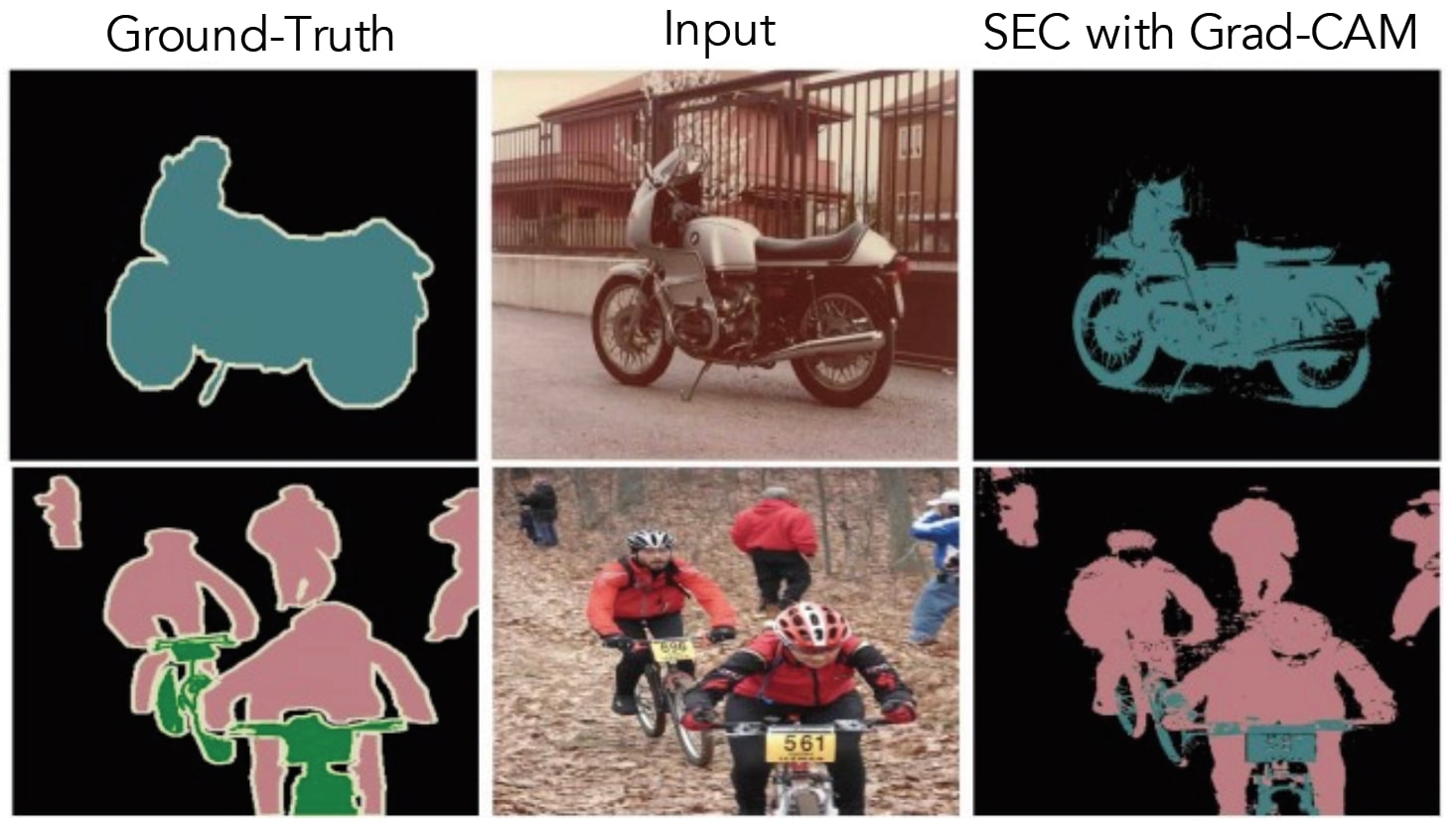}
 \caption{PASCAL VOC 2012 Segmentation results with \gcam{} as seed for SEC~\cite{seed_eccv16}.}
 \label{fig:segmentation_qual}
 \vspace{-10pt}
\end{figure}

However, their algorithm is sensitive to the choice of weak localization seed,
without which the network fails to localize objects correctly.
In their work, they used CAM maps from a VGG-16 based network which are used as object seeds for weakly localizing foreground classes.
We replaced the CAM maps with \gcam{} obtained from a standard VGG-16 network and obtain a Intersection over Union (IoU) score of 49.6 (compared to 44.6 obtained with CAM) on the PASCAL VOC 2012 segmentation task. \reffig{fig:segmentation_qual} shows some qualitative results.

\begin{figure*}[ht]
    \centering
   \begin{subfigure}[b]{0.25\textwidth}
       \centering
       \includegraphics[width=0.55\linewidth]{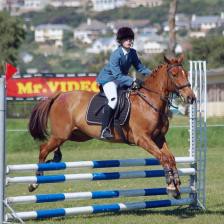}
       \vspace{0.72in}
       \caption{\scriptsize{Raw input image. Note that this is not a part of the tasks (b) and (c)}}
		\label{fig:hs_clsdisc}
	\end{subfigure}
    \unskip\ \vrule\
   \hspace{0.05in}
    \begin{subfigure}[b]{0.25\textwidth}
        \centering
        \includegraphics[width=0.90\linewidth]{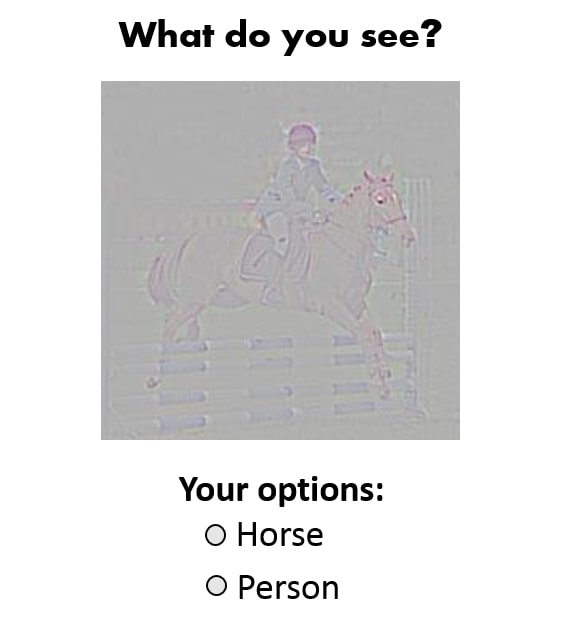}
        \vspace{0.183in}
        \caption{\scriptsize{AMT interface for evaluating the class-discriminative property}}
		\label{fig:hs_clsdisc}
	\end{subfigure}
    \unskip\ \vrule\ \hspace{0.05in}
	\begin{subfigure}[b]{0.45\textwidth}
        \centering
        \includegraphics[width=0.95\linewidth]{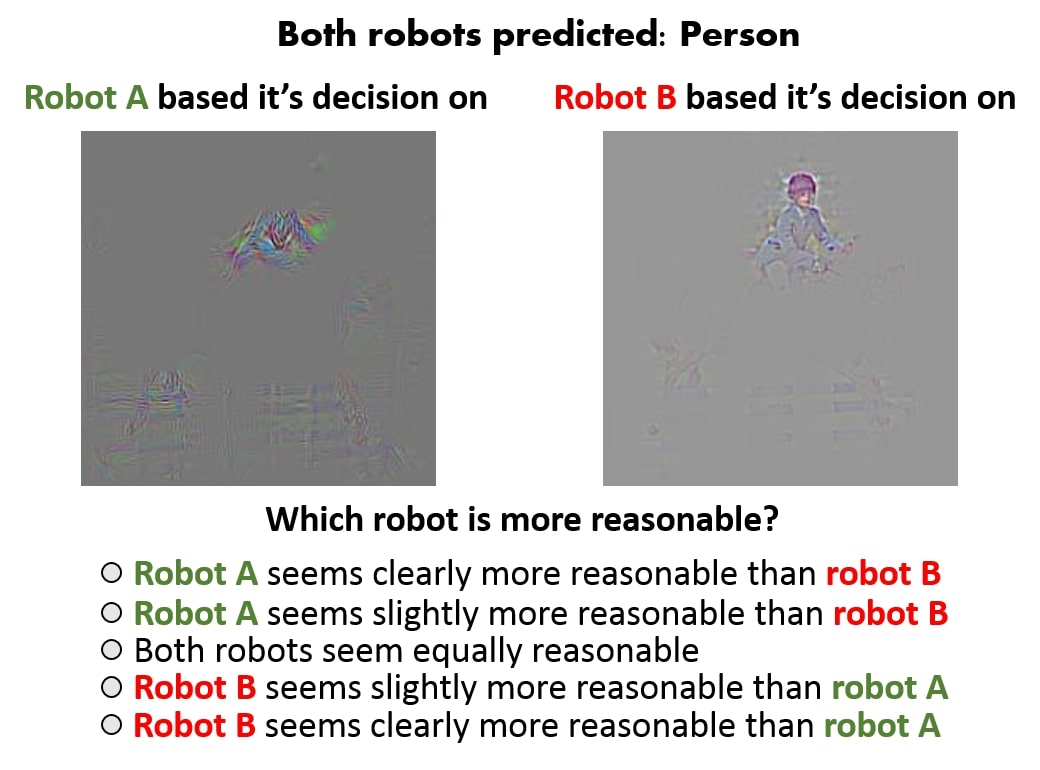}
        \vspace{0.01in}
        \caption{\scriptsize{AMT interface for evaluating if our visualizations instill trust in an end user}}%
		\label{fig:hs_trust}
	\end{subfigure}
    \vspace{12pt}
	\caption{\rp{AMT interfaces for evaluating different visualizations for class discrimination (b) and trustworthiness (c). \cgb{} outperforms baseline approaches (Guided-backprop and \dec) showing that our visualizations are more class-discriminative and help humans place trust in a more accurate classifier.}}
\label{fig:human_studies}
\end{figure*}

\vspace{-15pt}
\subsection{Pointing Game}\label{sec:pointing_game}
\vspace{-2pt}

Zhang \etal~\cite{zhang2016top} introduced the Pointing Game experiment to evaluate the discriminativeness of
different visualization methods for localizing target objects in scenes.
Their evaluation protocol first cues each visualization technique with the ground-truth object label
and extracts the maximally activated point on the generated heatmap. 
It then evaluates if the point lies within one of the annotated instances of the target object category,
thereby counting it as a hit or a miss.

The localization accuracy is then calculated as \\$Acc = \frac{\#Hits}{\#Hits+\#Misses}$.
However, this evaluation only measures precision of the visualization technique.
We modify the protocol to also measure recall --
we compute localization maps for top-5 class predictions from the
CNN classifiers\footnote{We use GoogLeNet finetuned on COCO, as provided by ~\cite{zhang2016top}.}
and evaluate them using the pointing game setup with an additional option to
reject any of the top-5 predictions from the model if the maximally activated
point in the map is below a threshold,
\rp{\ie if the visualization correctly rejects the predictions which are absent
from the ground-truth categories, it gets that as a hit.}
We find that \gcam{} outperforms c-MWP~\cite{zhang2016top} by a significant
margin (70.58\% \vs 60.30\%).
Qualitative examples comparing c-MWP~\cite{zhang2016top} and \gcam{} on %
can be found in \secref{sec:sup_pointing}\footnote{
c-MWP~\cite{zhang2016top} highlights arbitrary regions for predicted but non-existent categories, unlike \gcam{} maps which typically do not.}.

\vspace{-10pt}
\section{Evaluating Visualizations}\label{sec:human_evaluation}

\ad{In this section, we describe the human studies and experiments we conducted to understand
    the interpretability \vs faithfulness tradeoff of our approach to model predictions.}
Our first human study evaluates the main premise of our approach -- are \gcam{} visualizations more
class discriminative than previous techniques?
\mac{Having established that, we turn to understanding whether it can lead an end user to trust
    the visualized models appropriately.
    For these experiments, we compare VGG-16 and AlexNet finetuned on PASCAL VOC 2007
    \texttt{train} and visualizations evaluated on \texttt{val}.}

\vspace{-15pt}
\subsection{Evaluating Class Discrimination} \label{sec:class_disc}

\mac{In order to measure whether \gcam{} helps distinguish between classes,
we select images from the PASCAL VOC 2007 val set, which contain exactly $2$ annotated categories
and create visualizations for each one of them.}
For both VGG-16 and AlexNet CNNs, we obtain \rp{category-specific} visualizations using four techniques:
\dec{}, \gb{}, and \gcam{} versions of each of these methods (\cdec{} and \cgb{}).
We show these visualizations to 43 workers on Amazon Mechanical Turk (AMT) and ask
them ``Which of the two object categories is depicted in the image?''
(shown in \figref{fig:human_studies}).

Intuitively, a good prediction explanation is one that produces discriminative visualizations for the class of interest.
The experiment was conducted using all 4 visualizations for 90 image-category pairs (\ie 360 visualizations); 9 ratings were collected for each image,
evaluated against the ground truth and averaged to obtain the accuracy in \reftab{tab:eval_vis}.
When viewing \cgb{}, human subjects can correctly identify the category being visualized in $61.23$\%
of cases (compared to $44.44$\% for \gb{}; thus, \gcam{} improves human performance by $16.79$\%).
Similarly, we also find that \gcam{} helps make \dec{} more class-discriminative
(from $53.33$\% $\rightarrow$ $60.37$\%). \cgb{} performs the best among all methods.
Interestingly, our results indicate that \dec{} is more class-discriminative
than \gb{} ($53.33$\% \vs $44.44$\%),
although \gb{} is more aesthetically pleasing.
To the best of our knowledge, our evaluations are the first to quantify \mac{this subtle difference.}

\begin{table}[h!]
\vspace{-15pt}
\centering
\resizebox{.48\textwidth}{!}{
    \begin{tabular}{c p{3.0cm} p{1.7cm} p{3.1cm}}\toprule
        \textbf{Method} & \textbf{Human Classification Accuracy} & \textbf{Relative Reliability} & \textbf{Rank Correlation \;\;\; w/ Occlusion} \\
        \midrule
        \gb{}  & 44.44 & +1.00 & 0.168 \\
        \cgb{} & 61.23 & +1.27 & 0.261 \\
        \bottomrule
    \end{tabular}
}
    \vspace{2pt}
    \caption{\review{Quantitative Visualization Evaluation.
\cgb{} enables humans to differentiate between visualizations of different classes (Human Classification Accuracy) and 
pick more reliable models (Relative Reliability). It also accurately reflects the behavior of the model (Rank Correlation w/ Occlusion).}}
\label{tab:eval_vis}
\end{table}

\vspace{-15pt}
\subsection{Evaluating Trust}
Given two prediction explanations, we evaluate which seems more trustworthy.
We use AlexNet and VGG-16 to compare \gb{} and \cgb{} visualizations, noting
that VGG-16 is known to be more reliable than AlexNet with an accuracy of $79.09$ mAP
(\vs $69.20$ mAP) on PASCAL classification.
In order to tease apart the efficacy of the visualization from the accuracy of the model being visualized, we consider only those instances where \emph{both} models made the same prediction as ground truth. %
Given a visualization from AlexNet and one from VGG-16, and the predicted object category, \rp{54 AMT workers} were instructed to rate the reliability of the models relative to each other on a scale of clearly more/less reliable (+/-$2$), slightly more/less reliable (+/-$1$), and equally reliable ($0$).
This interface is shown in \figref{fig:human_studies}.
To eliminate any biases, VGG-16 and AlexNet were assigned to be `model-1' with approximately equal probability.
Remarkably, as can be seen in \reftab{tab:eval_vis}, we find that human subjects are able to identify the more accurate
classifier (VGG-16 over AlexNet) \emph{\iccv{simply from the prediction explanations,
despite both models making identical predictions.}}
With \gb{}, humans assign VGG-16 an average score of $1.00$ which means that it is
slightly more reliable than AlexNet, while \cgb{} achieves a higher score of $1.27$
which is closer to saying that VGG-16 is clearly more reliable.
Thus, our visualizations can help users place trust in a model that generalizes better,
just based on individual prediction explanations.

\vspace{-20pt}
\subsection{Faithfulness \vs Interpretability}\label{sec:occ}

Faithfulness of a visualization to a model is its ability to accurately explain the function learned by the model.
Naturally, there exists a trade-off between the interpretability and faithfulness of a visualization --
a more faithful visualization is typically less interpretable and \viceversa.
In fact, one could argue that a fully faithful explanation is the entire description of the model, which in the case of deep models is not interpretable/easy to visualize.
We have verified in previous sections that our visualizations are reasonably interpretable.
We now evaluate how faithful they are to the underlying model.
One expectation is that our explanations should be locally accurate, \ie in the vicinity of the input data point, our explanation should be faithful to the model~\cite{lime_sigkdd16}.

For comparison, we need a reference explanation with high local-faithfulness.
One obvious choice for such a visualization is image occlusion~\cite{zeiler_eccv14}, where we measure the difference in CNN scores when patches of the input image are masked. %
\rp{Interestingly, patches which change the CNN score are also patches to which \gcam{} and \cgb{} assign high intensity, achieving rank correlation $0.254$ and $0.261$ (\vs $0.168$, $0.220$
and $0.208$ achieved by \gb{}, c-MWP and CAM respectively) averaged over 2510
images in the PASCAL 2007 val set. This shows that \gcam{} is more faithful to
the original model compared to prior methods.
Through localization experiments and human studies, we
see that \gcam{} visualizations are \emph{more interpretable}, and through
correlation with occlusion maps, we see that \gcam{} is \emph{more faithful} to the model.}

\vspace{-75pt}
\section{Diagnosing image classification CNNs with \gcam{}}\label{sec:benefits}

\ijcv{In this section we further demonstrate the use of \gcam{} in analyzing failure modes of image classification CNNs, understanding the effect of adversarial noise, and identifying and removing biases in datasets, in the context of VGG-16 pretrained on imagenet.}

\vspace{-8pt}
\subsection{Analyzing failure modes for VGG-16}\label{sec:diagnose}
\vspace{-5pt}

\begin{figure}[ht!]
    \begin{center}
    \begin{subfigure}[b]{0.23\linewidth}
        \centering
        \includegraphics[width=1\linewidth]{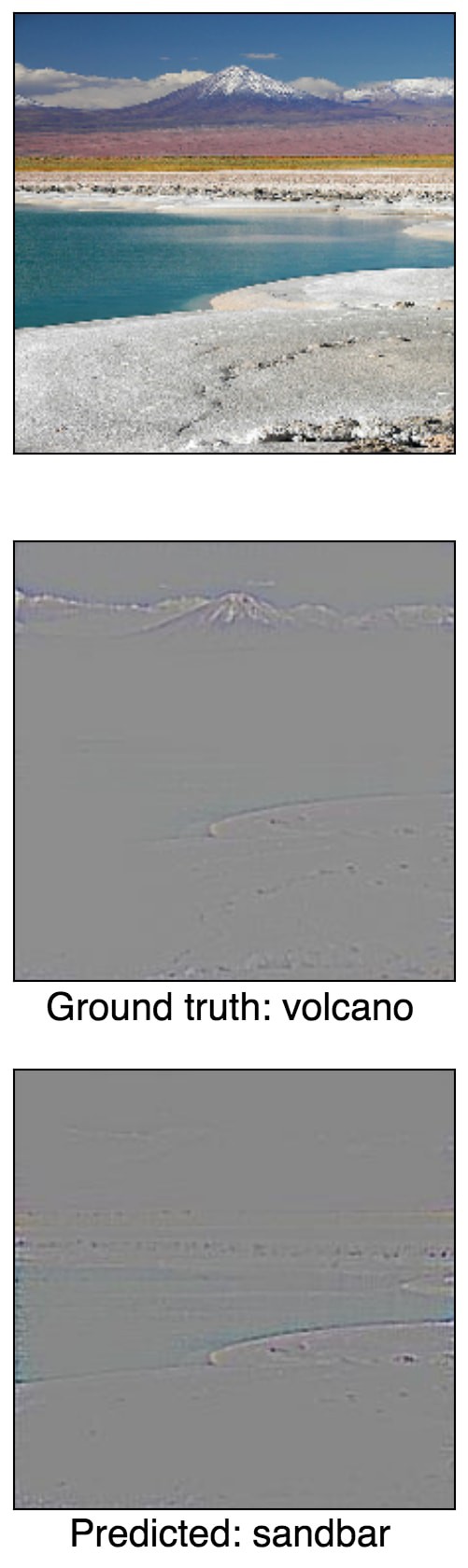}
        \caption{}
        \label{fig:failure_volcano}
    \end{subfigure}
    ~
    \begin{subfigure}[b]{0.23\linewidth}
        \centering
        \includegraphics[width=1\linewidth]{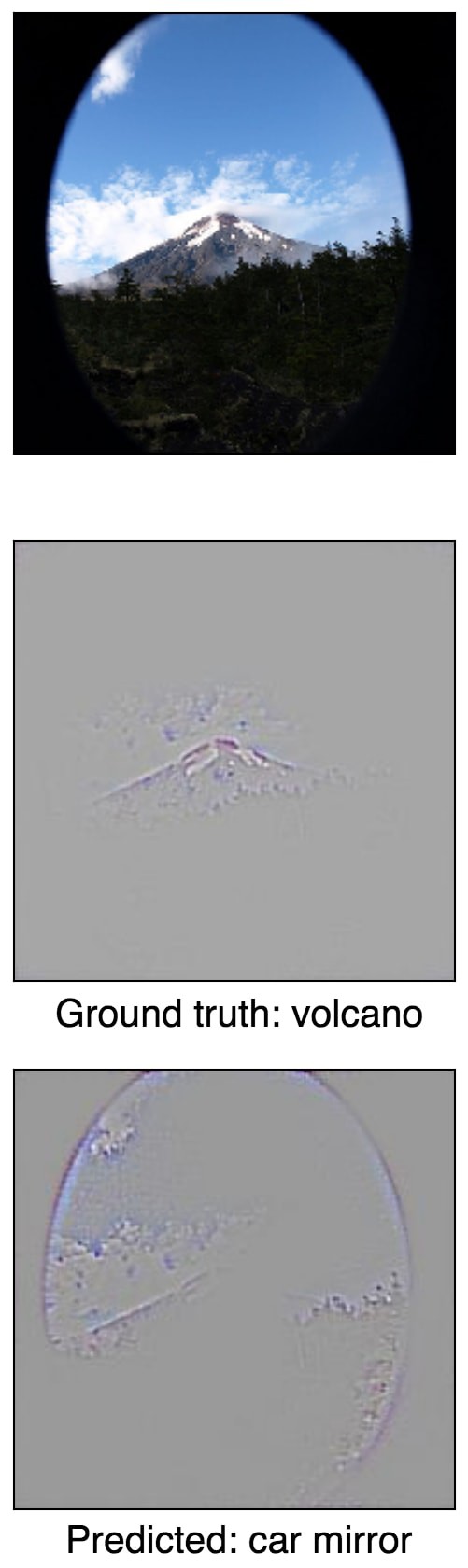}
        \caption{}
        \label{fig:failure_mirror}
    \end{subfigure}
    \begin{subfigure}[b]{0.23\linewidth}
        \centering
        \includegraphics[width=1\linewidth]{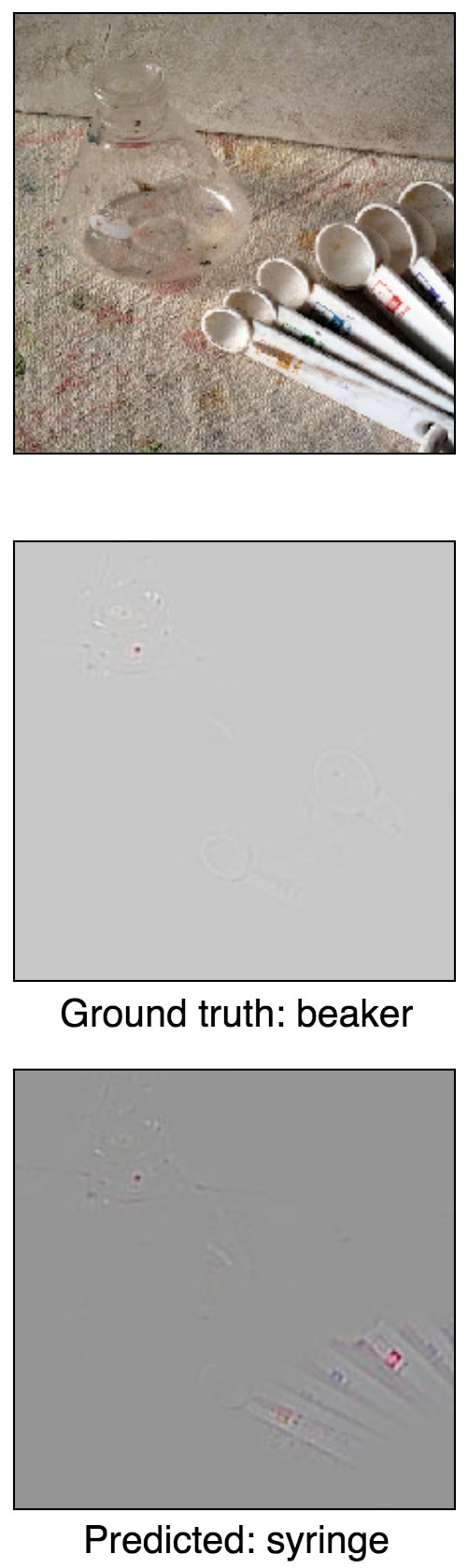}
        \caption{}
        \label{fig:failure_syringe}
    \end{subfigure}
    \begin{subfigure}[b]{0.23\linewidth}
        \centering
        \includegraphics[width=1\linewidth]{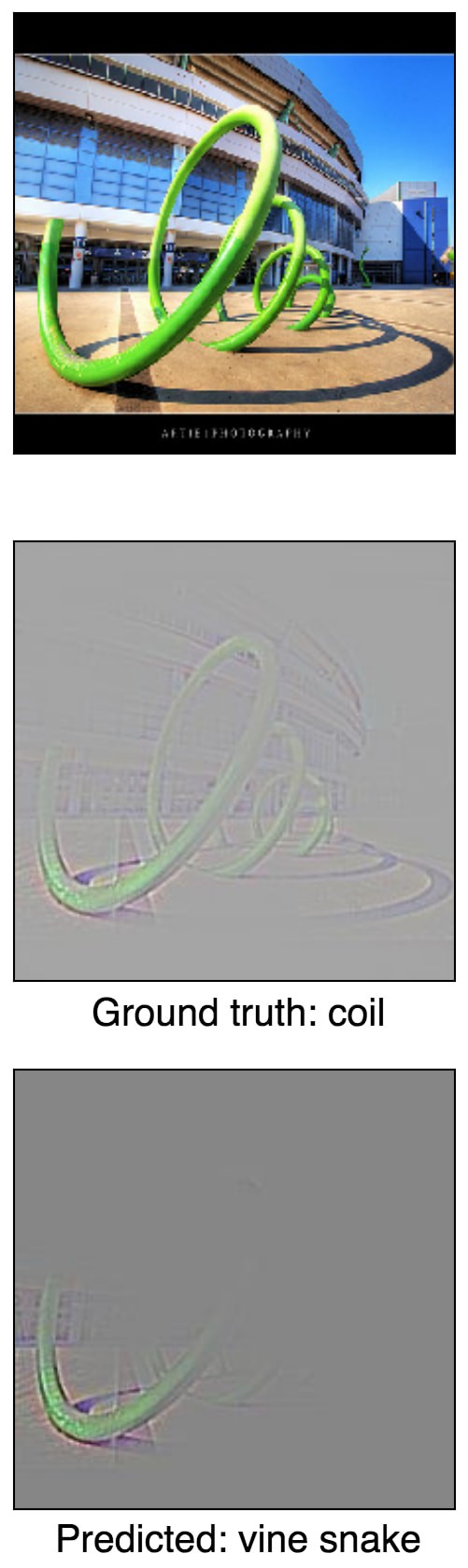}
        \caption{}
        \label{fig:failure_snake}
    \end{subfigure}
    \vspace{14pt}
     \caption{
In these cases the model (VGG-16) failed to predict the correct class in its top 1 (a and d) and top 5 (b and c) predictions. Humans would find it hard to explain some of these predictions without looking at the visualization for the predicted class. But with  \gcam{}, these mistakes seem justifiable.
    }
    \label{fig:failures}
    \end{center}
    \vspace{-5pt}
\end{figure}

In order to see what mistakes a network is making, we first get a list of examples
that the network (VGG-16) fails to classify correctly.
For these misclassified examples, we use \cgb{} to visualize both the correct and the predicted class.
As seen in \figref{fig:failures}, some failures are due to ambiguities inherent
in ImageNet classification. We can also see that \emph{seemingly unreasonable
predictions have reasonable explanations}, an
observation also made in HOGgles~\cite{vondrick_iccv13}.
    A major advantage of \cgb{} visualizations over other methods is that due to
    its high-resolution and ability to be class-discriminative, it readily enables these analyses.

\vspace{-8pt}
\subsection{Effect of adversarial noise on VGG-16}\label{sec:adversarial_noise}

\begin{figure}[ht!]
\vspace{-2pt}
  \begin{center}
    \begin{subfigure}[b]{0.32\linewidth}
        \centering
        \includegraphics[width=1\linewidth]{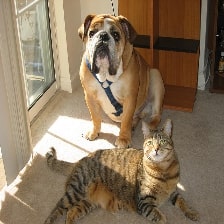}
        \tiny{Boxer: 0.4   Cat: 0.2} %
		\caption{\tiny{Original image}}

        \label{fig:orig_cat_dog}
    \end{subfigure}
    \begin{subfigure}[b]{0.32\linewidth}
        \centering
        \includegraphics[width=1\linewidth]{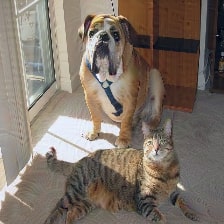}
        \tiny{Airliner: 0.9999\\
        }
		\caption{\tiny{Adversarial image}}
        \label{fig:adv_cat_dog_airliner}
    \end{subfigure}
    \begin{subfigure}[b]{0.32\linewidth}
        \centering
        \includegraphics[width=1\linewidth]{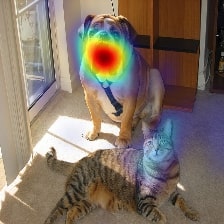}
        \tiny{Boxer: 1.1e-20}
		\caption{\tiny{\gcam{} ``Dog''}}
        \label{fig:gcam_dog_airliner}
    \end{subfigure}
    
    \begin{subfigure}[b]{0.32\linewidth}
    \vspace{10pt}
        \centering
        \includegraphics[width=1\linewidth]{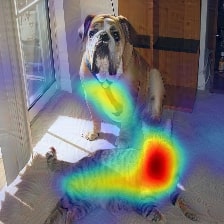}
        \tiny{Tiger Cat: 6.5e-17\\}
		\caption{\tiny{\gcam{} ``Cat''}}
        \label{fig:gcam_cat_airliner}
	\end{subfigure}
    \begin{subfigure}[b]{0.33\linewidth}
    \vspace{10pt}
        \centering
        \includegraphics[width=1\linewidth]{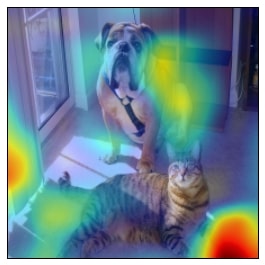}
        \tiny{Airliner: 0.9999\\}
		\caption{\tiny{\gcam{} ``Airliner''}}
        \label{fig:gcam_airliner_airliner}
	\end{subfigure}
    \begin{subfigure}[b]{0.33\linewidth}
    \vspace{10pt} 
        \centering
        \includegraphics[width=1\linewidth]{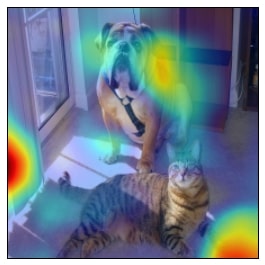}
        \tiny{Space shuttle: 1e-5\\}
		\caption{\tiny{\gcam{} ``Space Shuttle''}}
        \label{fig:gcam_spaceshuttle_airliner}
	\end{subfigure}
	\vspace{14pt}
	\caption{\rp{
	  (a-b) Original image and the generated adversarial image for
	  category ``airliner''. (c-d) \gcam{} visualizations for the
	  original categories ``tiger cat'' and ``boxer (dog)'' along with their confidence. Despite the network being completely fooled into predicting the dominant category label of
	  ``airliner'' with high confidence (>0.9999), \gcam{} can localize the original categories accurately. \rpi{(e-f) \gcam{} for the top-2 predicted classes ``airliner'' and ``space shuttle'' seems to highlight the background.}}
    }
    \label{fig:adversarial_main}
    \end{center}
\end{figure}

Goodfellow \etal~\cite{goodfellow2015explaining} demonstrated the
vulnerability of current deep networks to adversarial examples, which
are slight imperceptible perturbations of input images that fool the network into
misclassifying them with high confidence. We generate adversarial
images for an ImageNet-pretrained VGG-16 model such that it assigns high probability
($>0.9999$) to a category that is not present in the image and low probabilities
to categories that are present. We then compute
\gcam{} visualizations for the categories that are present.
As shown in \reffig{fig:adversarial_main}, despite the network
being certain about the absence of these categories (`tiger cat' and `boxer'),
\gcam{} visualizations can correctly localize them.
This shows that \gcam{} is fairly robust to adversarial noise.

\vspace{-20pt}
\subsection{Identifying bias in dataset}\label{sec:bias}
In this section, we demonstrate another use of \gcam{}: identifying and
reducing bias in training datasets. Models trained on biased datasets may not
generalize to real-world scenarios, or worse, may perpetuate biases and stereotypes
(\wrt gender, race, age, \etc). %
We finetune an ImageNet-pretrained VGG-16 model for a ``doctor'' \vs ``nurse''
binary classification task. We built our training and validation splits using
the top $250$ relevant images (for each class) from a popular image search engine.
And the test set was controlled to be balanced in its distribution
of genders across the two classes. Although the trained model achieves good validation
accuracy, it does not generalize well ($82$\% test accuracy).

\begin{figure}[htp]
 \centering
 \includegraphics[width=1\linewidth]{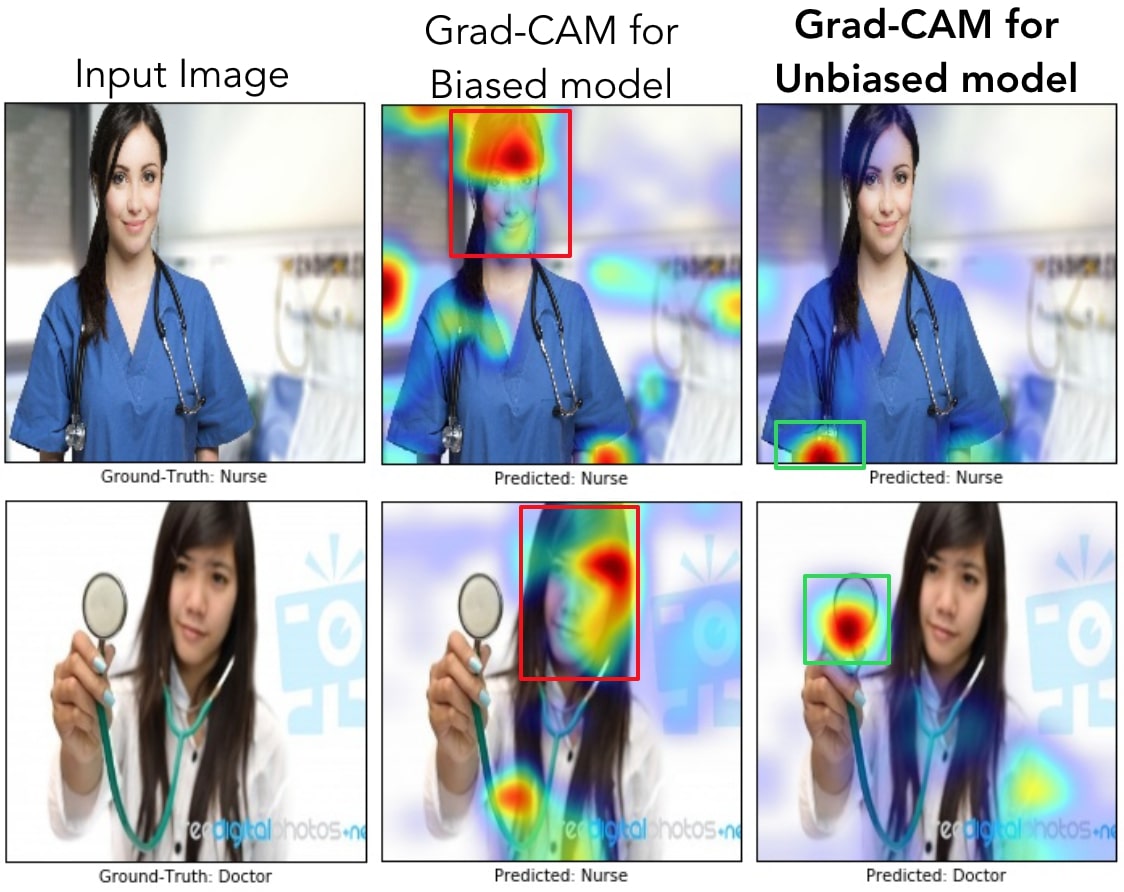}
 \caption{In the first row, we can see that even though both models made the right decision, the biased model (model1) was looking at the face of the person to decide if the person was a nurse, whereas the unbiased model was looking at the short sleeves to make the decision. For the example image in the second row, the biased model made the wrong prediction (misclassifying a doctor as a nurse) by looking at the face and the hairstyle, whereas the unbiased model made the right prediction looking at the white coat, and the stethoscope.
 }
 \label{fig:bias_qual}
\end{figure}

\gcam{} visualizations of the model predictions (see the red box\footnote{The green and red boxes are drawn manually to highlight correct and incorrect focus of the model.} regions in the middle column of \reffig{fig:bias_qual})
revealed that the model had learned to look at the person's face / hairstyle to distinguish nurses from doctors, thus learning a gender stereotype.
Indeed, the model was misclassifying several female doctors to be a nurse and male nurses to be a doctor. Clearly, this is problematic. Turns out the image search results were gender-biased (78\% of images for doctors were men, and 93\% images for nurses were women).

Through these intuitions gained from \gcam{} visualizations,
we reduced bias in the training set by adding in images of male nurses and female doctors,
while maintaining the same number of images per class as before.
The re-trained model not only generalizes better ($90$\% test accuracy),
but also looks at the right regions (last column of \reffig{fig:bias_qual}).
This experiment demonstrates a proof-of-concept that \gcam{} can help detect and
remove biases in datasets, which is important not just for better generalization,
but also for fair and ethical outcomes as more algorithmic decisions are made in society.

\vspace{-10pt}
\section{Textual Explanations with \gcam{}}\label{sec:text_exp}

\begin{figure*}[h]
\begin{center}
\begin{subfigure}[t]{\columnwidth}
\includegraphics[scale=0.135]{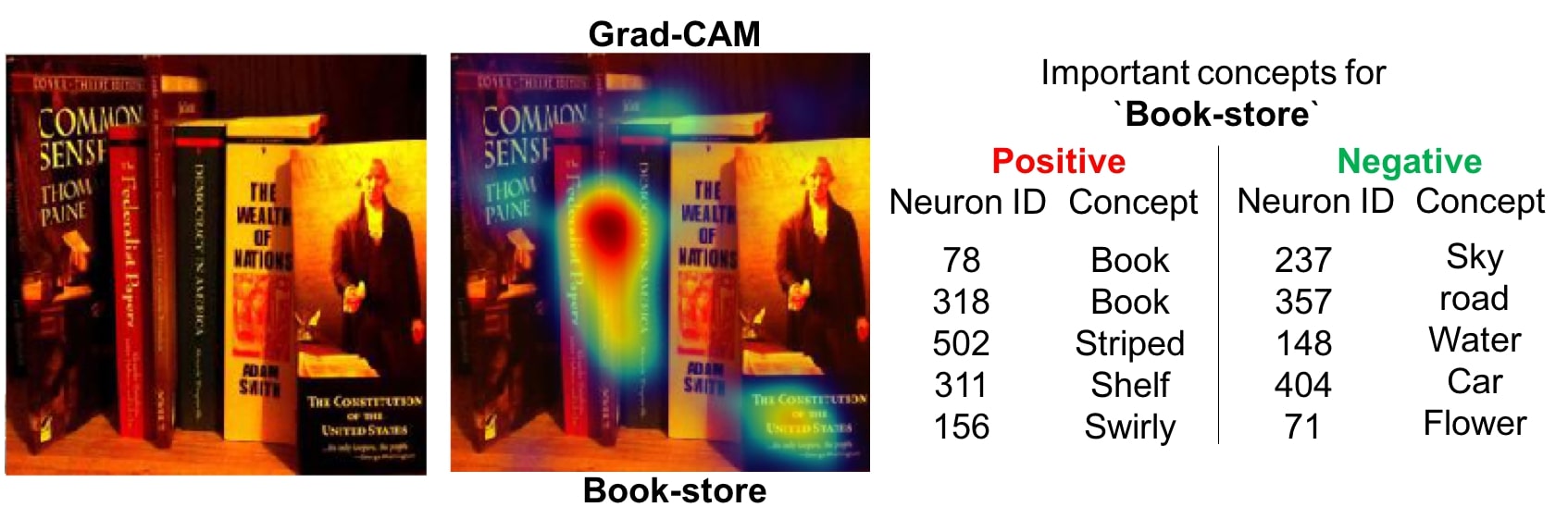}\caption{}
\vspace{10pt}
\end{subfigure}
\begin{subfigure}[t]{\columnwidth}
\includegraphics[scale=0.135]{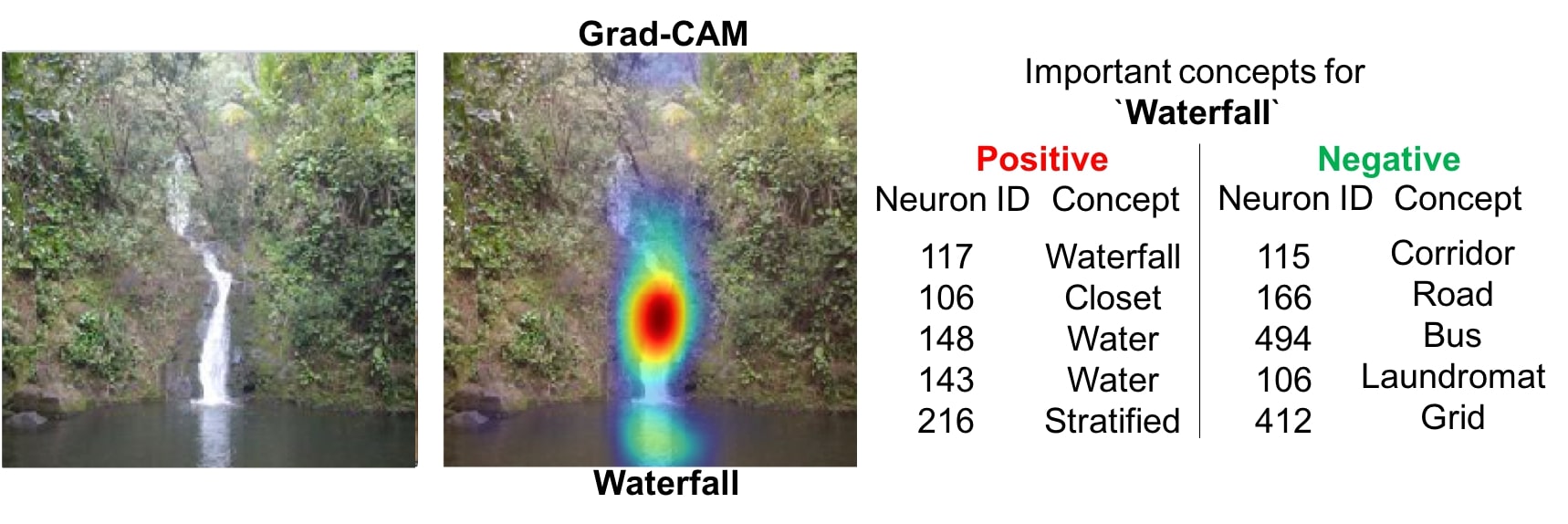}\caption{}
\vspace{10pt}
\end{subfigure}
\begin{subfigure}[t]{\columnwidth}
\includegraphics[scale=0.135]{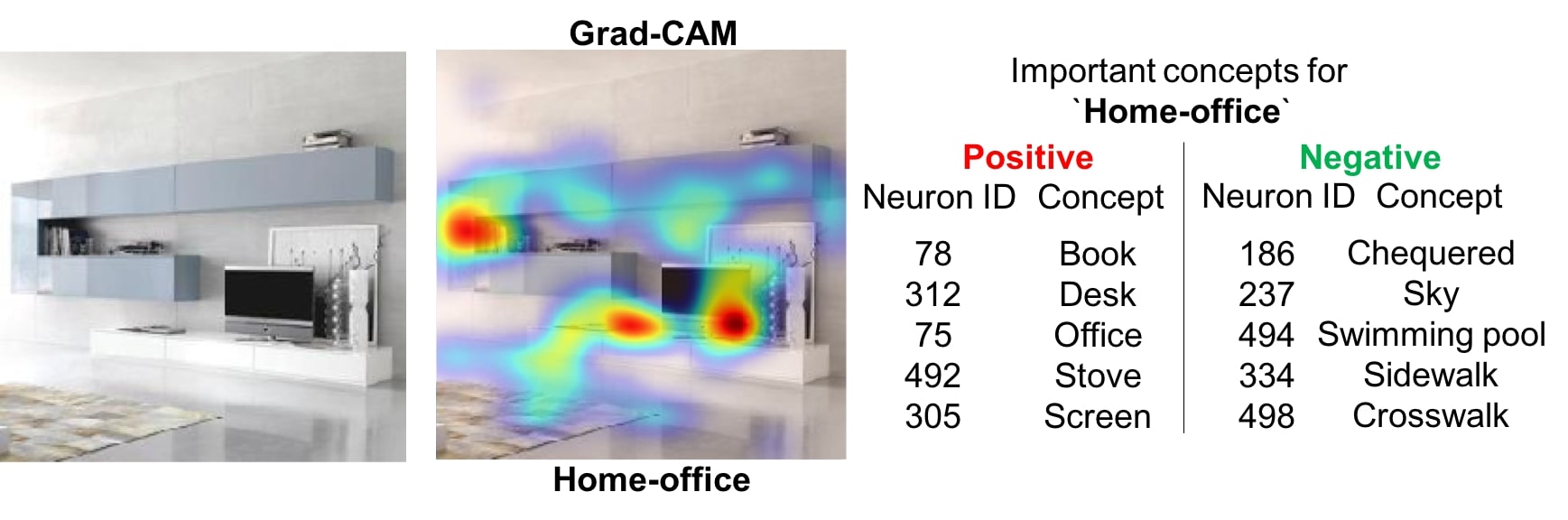}\caption{}
\vspace{10pt}
\end{subfigure}
\begin{subfigure}[t]{\columnwidth}
\includegraphics[scale=0.135]{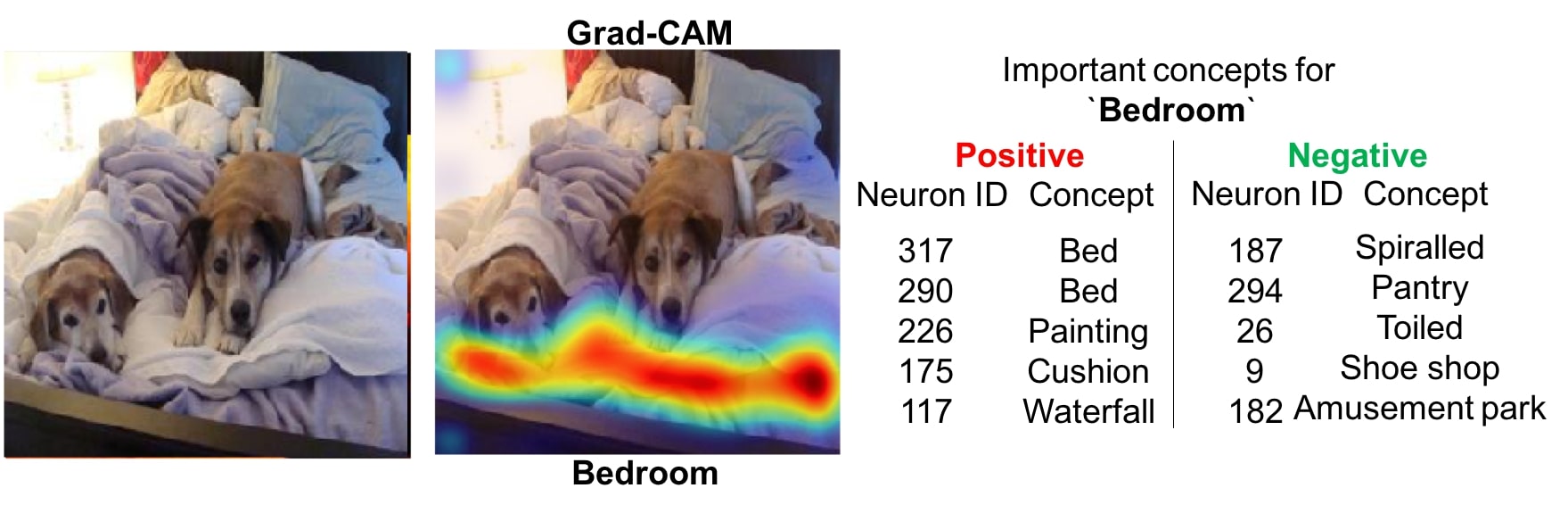}\caption{}
\vspace{10pt}
\end{subfigure}
\begin{subfigure}[t]{\columnwidth}
\includegraphics[scale=0.135]{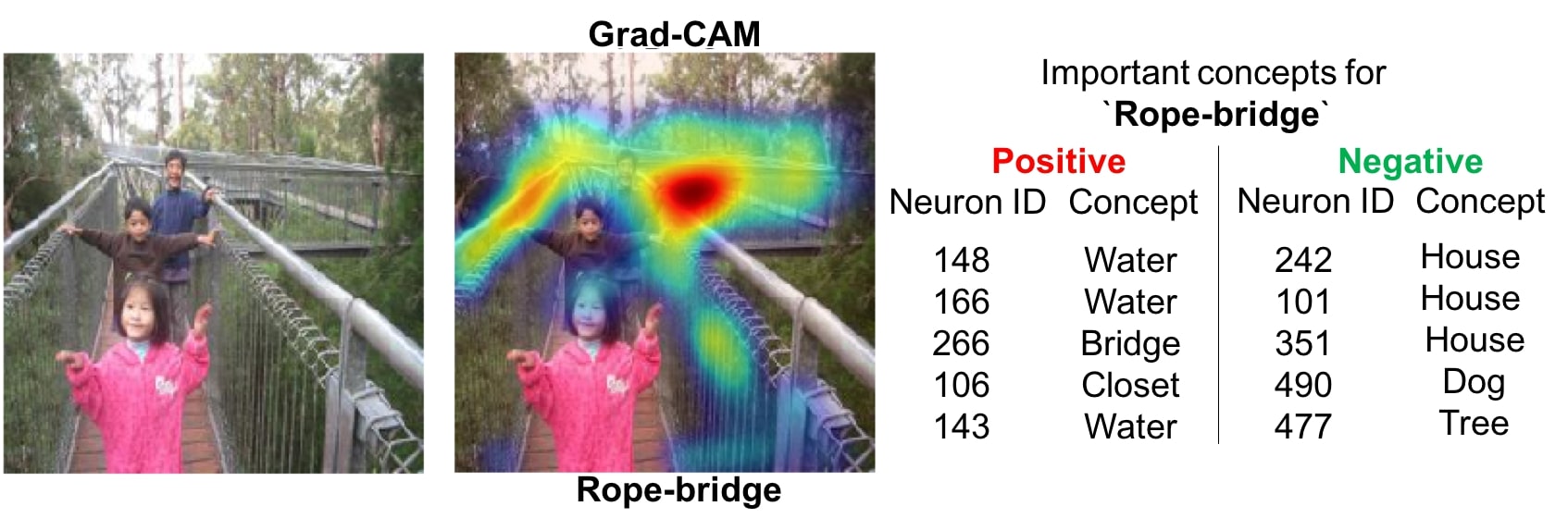}\caption{}
\vspace{10pt}
\end{subfigure}
\begin{subfigure}[t]{\columnwidth}
\includegraphics[scale=0.135]{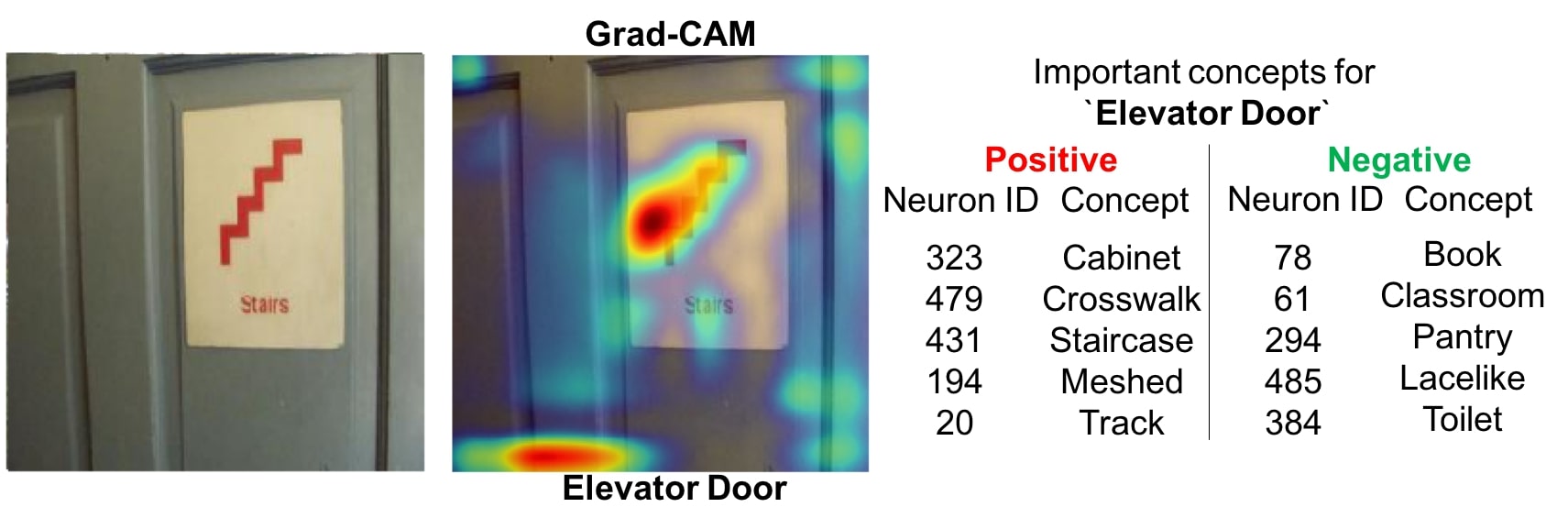}\caption{}
\vspace{10pt}
\end{subfigure}
  \caption{\ijcv{Examples showing visual explanations and textual explanations for VGG-16 trained on Places365 dataset~\cite{zhou2017places}. For textual explanations we provide the most important neurons for the predicted class along with their names. Important neurons can be either be persuasive (positive importance) or inhibitive (negative importance). The first 2 rows show success cases, and the last row shows 2 failure cases. We see that in (a), the important neurons computed by \eqref{eq:alpha1} look for concepts such as book and shelf which are indicative of class `Book-store' which is fairly intuitive. %
  }}
  \label{fig:text_explanations}
\end{center}
\end{figure*}

Equation. \eqref{eq:alpha1} gives a way to obtain neuron-importance, $\alpha{}$, for each neuron in a convolutional layer for a particular class.
There have been hypotheses presented in the literature \cite{Zhou2014ObjectDE,zeiler_eccv14} that neurons act as concept `detectors'.
Higher positive values of the neuron importance indicate that the presence of that concept leads to an increase in the class score,
whereas higher negative values indicate that its absence leads to an increase in the score for the class.

Given this intuition, let's examine a way to generate textual explanations.
In recent work, Bau~\etal~\cite{netdissect} proposed an approach to automatically name
neurons in any convolutional layer of a trained network.
These names indicate concepts that the neuron looks for in an image.
Using their approach. we first obtain neuron names for the last convolutional layer. %
Next, we sort and obtain the top-5 and bottom-5 neurons based on their class-specific importance scores, $\alpha_k$.
The names for these neurons can be used as text explanations.

\reffig{fig:text_explanations} shows some examples of visual and textual explanations for the
image classification model (VGG-16) trained on the Places365 dataset~\cite{zhou2017places}.
In (a), the positively important neurons computed by \eqref{eq:alpha1} look for intuitive
concepts such as book and shelf that are indicative of the class `Book-store'.
Also note that the negatively important neurons look for concepts such as sky, road, water and car which don't occur in `Book-store' images.
In (b), for predicting `waterfall', both visual and textual explanations highlight
`water' and `stratified' which are descriptive of `waterfall' images.
(e) is a failure case due to misclassification as the network predicted
`rope-bridge' when there is no rope, but still the important concepts (water
and bridge) are indicative of the predicted class.
In (f), while \gcam{} correctly looks at the door and the staircase on the
paper to predict `Elevator door', the neurons detecting doors did not pass the
IoU threshold\footnote{Area of overlap between ground truth concept annotation and neuron activation over area of their union. More details of this metric can be found in \cite{netdissect}} of 0.05 (chosen in order to suppress the noise in the neuron names),
and hence are not part of the textual explanations. More qualitative examples
can be found in the \refsec{sec:sup_text_exp}.

\begin{figure*}[ht!]
    \centering
  \begin{subfigure}[t]{0.49\textwidth}
    \includegraphics[width=\textwidth]{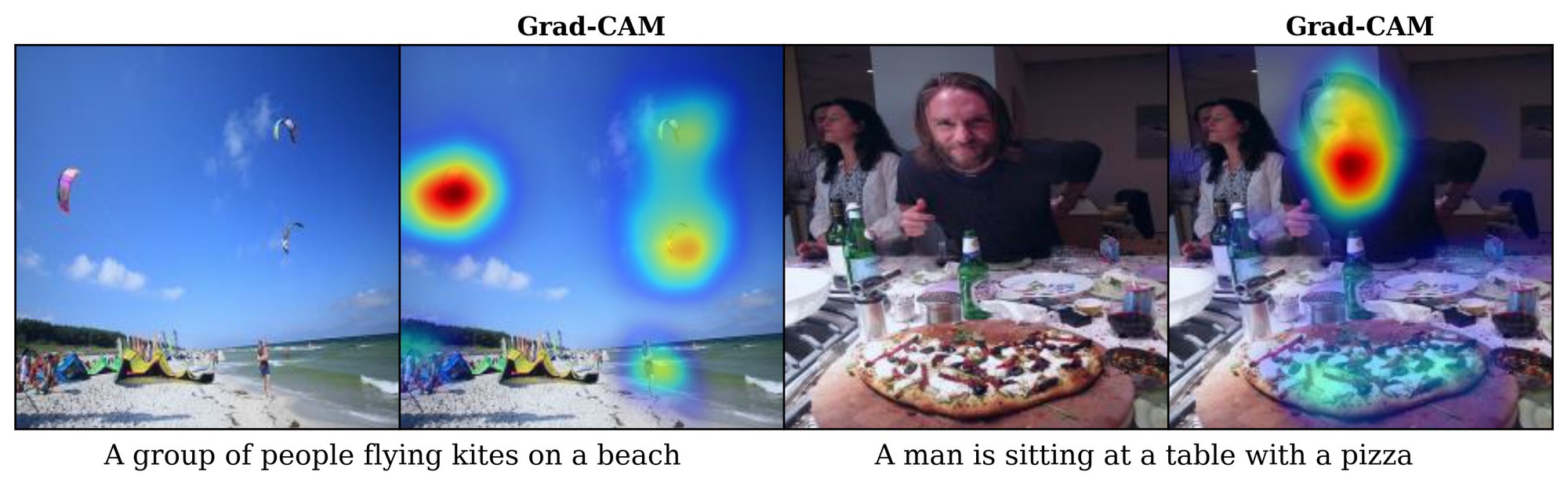}
    \caption{\scriptsize{Image captioning explanations}}
  \label{fig:captioning}
  \end{subfigure}
  ~
    \begin{subfigure}[t]{0.49\textwidth}
    \includegraphics[width=\textwidth]{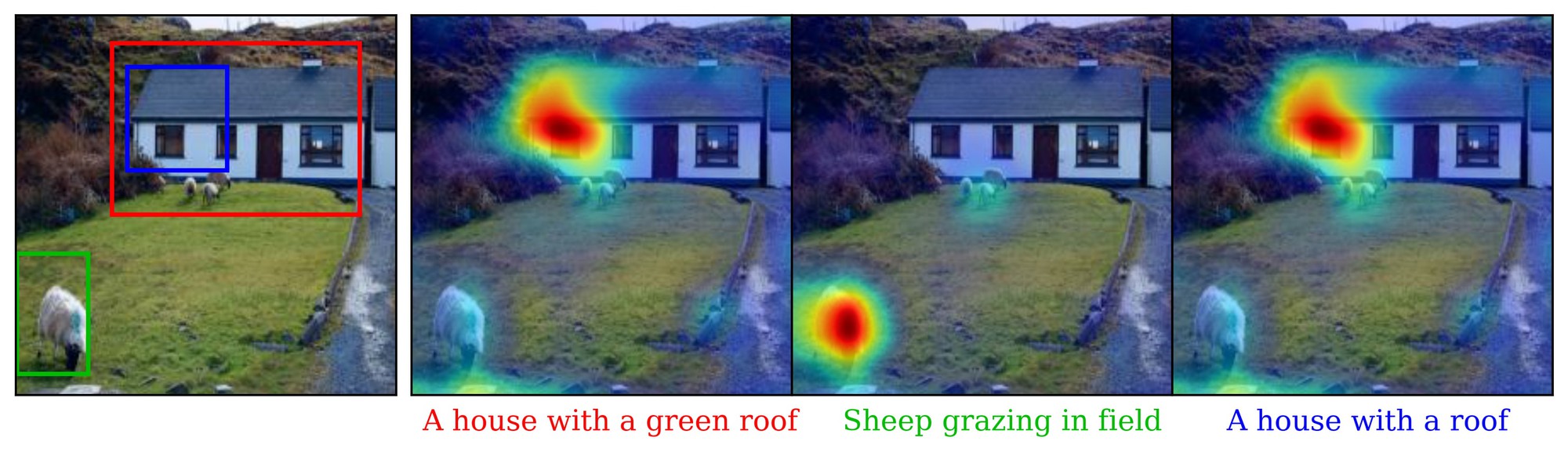}
    \caption{\scriptsize{Comparison to DenseCap}}
   \label{fig:densecap}
   \end{subfigure}
   \vspace{14pt}
    \caption{Interpreting image captioning models: We use our class-discriminative localization technique, \gcam{} to find spatial support regions for captions in images. \reffig{fig:captioning} Visual explanations from image captioning model~\cite{karpathy2015deep} highlighting image regions considered to be important for producing the captions. \reffig{fig:densecap} \gcam{} localizations of a \emph{global} or \emph{holistic} captioning model for captions generated by a dense
    captioning model~\cite{johnson_cvpr16} for the three bounding box proposals marked on the left. We can see that we get back \gcam{} localizations (right) that agree with those bounding boxes -- even though the captioning model and Grad-CAM techniques do not use any bounding box annotations.}
\end{figure*}

\begin{figure*}[h]
\begin{center}
\begin{subfigure}[t]{\columnwidth}
\hspace{-20pt}
\includegraphics[scale=0.25]{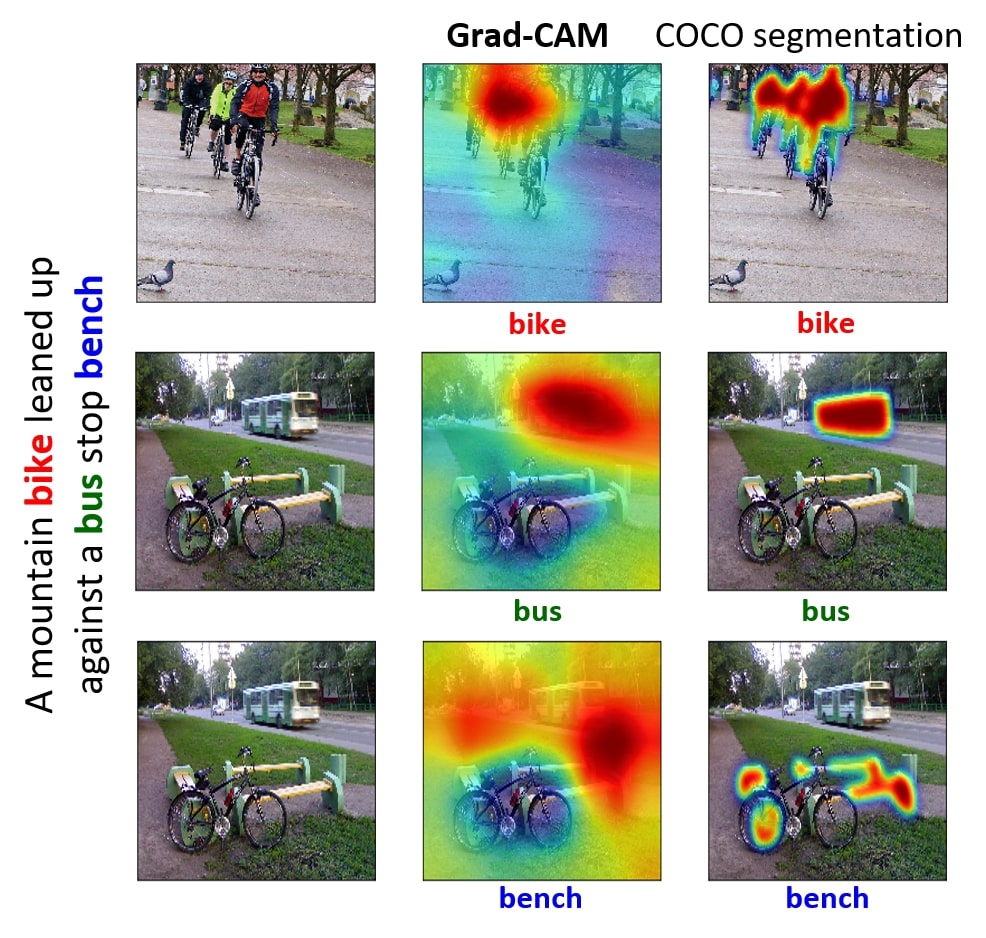}\caption{}
\vspace{10pt}
\end{subfigure}
\begin{subfigure}[t]{\columnwidth}
\includegraphics[scale=0.25]{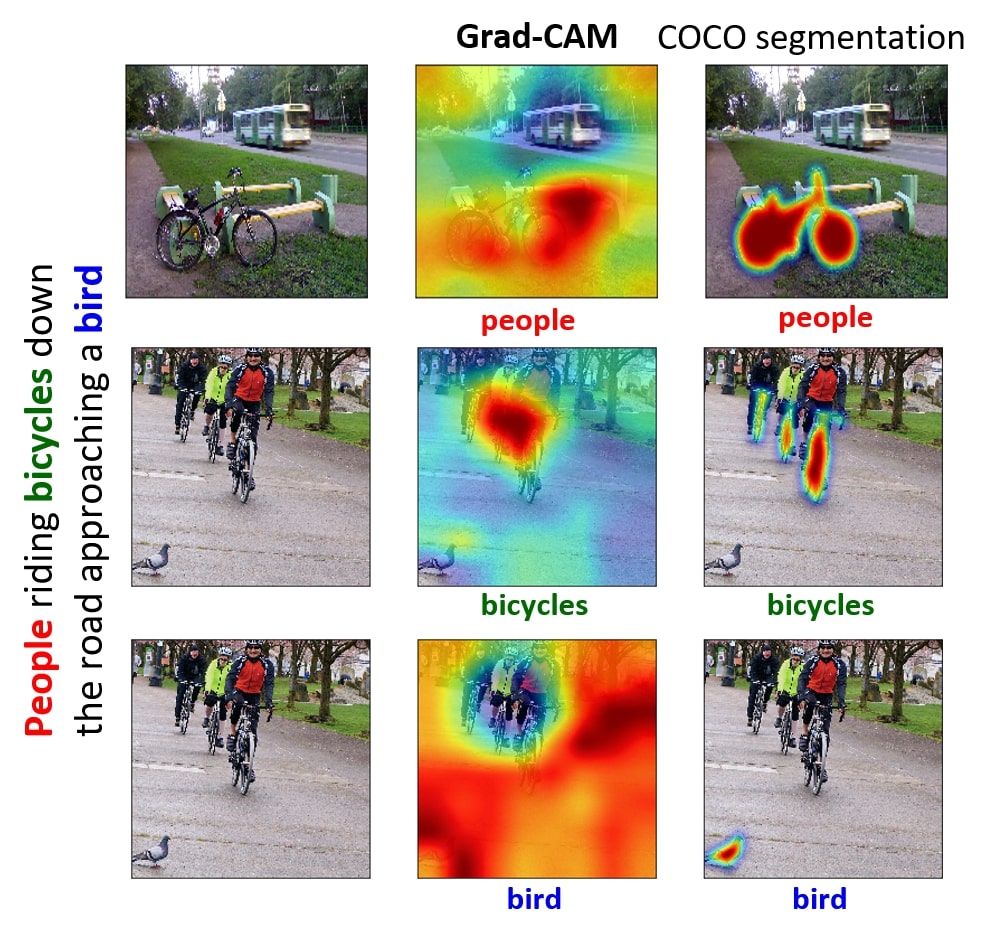}\caption{}
\vspace{10pt}

\end{subfigure}
\caption{\rpi{Qualitative Results for our word-level captioning experiments: (a) Given the image on the left and the caption, we visualize \gcam{} maps for the visual words ``bike", ``bench" and ``bus". Note how well the \gcam{} maps correlate with the COCO segmentation maps on the right column. (b) shows a similar example where we visualize \gcam{} maps for the visual words ``people", ``bicycle" and ``bird".}}
\label{fig:captioning_word}
\end{center}
\end{figure*}

\vspace{-10pt}
\section{\gcam{} for Image Captioning and VQA}
Finally, we apply \gcam{} to vision \& language tasks such as image captioning~\cite{chen2015microsoft,johnson_cvpr16,vinyals_cvpr15}
and Visual Question Answering (VQA)~\cite{antol2015vqa,gao2015you,malinowski_iccv15,ren_nips15}.
We find that \gcam{} leads to interpretable visual explanations for these tasks
as compared to baseline visualizations which do not change noticeably across changing predictions.
Note that existing visualization techniques either are not class-discriminative
(\gb{}, \dec{}), or simply cannot be used for these tasks/architectures,
or both (CAM, c-MWP).

\vspace{-10pt}
\subsection{Image Captioning}\label{sec:nic}

  In this section, we visualize spatial support for an image captioning model using \gcam{}.
  We build \gcam{} on top of the publicly available neuraltalk2\footnote{\url{https://github.com/karpathy/neuraltalk2}} implementation~\cite{karpathy2015deep} that uses a finetuned VGG-16 CNN for images and an LSTM-based language model. Note that this model does not have an explicit attention mechanism.
  Given a caption, we compute the gradient of its log probability~\wrt units in the last convolutional layer of the CNN ($conv5\_3$ for VGG-16) and generate \gcam{} visualizations as described in \secref{sec:approach}.
See \reffig{fig:captioning}.
In the first example, \gcam{} maps for the generated caption localize
every occurrence of both the kites and people despite their relatively small size.
In the next example, \gcam{} correctly highlights the pizza and the man, but ignores the woman nearby, since `woman' is not mentioned in the caption. More examples are in \refsec{sec:sup_experiments}.%

\noindent \para{Comparison to dense captioning}
Johnson~\etal~\cite{johnson_cvpr16} recently introduced the Dense Captioning (DenseCap) task that requires a system to jointly localize and caption salient regions in a given image.
Their model consists of a Fully Convolutional Localization Network (FCLN)
that produces bounding boxes for regions of interest
and an LSTM-based language model that generates associated captions, all in a
single forward pass.
Using DenseCap, we generate \iccv{5 region-specific captions per image with associated ground truth bounding boxes.
\gcam{} for a whole-image captioning model (neuraltalk2) should localize the bounding box
the region-caption was generated for, which is shown in \reffig{fig:densecap}.
We quantify this by computing the ratio of mean activation inside \vs
outside the box. Higher ratios are better because they indicate stronger
attention to the region the caption was generated for.
Uniformly highlighting the whole image results in a baseline ratio of $1.0$ whereas
\gcam{} achieves $3.27$ $\pm$ $0.18$. Adding high-resolution detail gives an improved
baseline of $2.32$ $\pm$ 0.08 (\gb{}) and the best localization at $6.38$ $\pm$ 0.99 (\cgb{}).
Thus, \gcam{} is able to localize regions in the image that the
DenseCap model describes, even though the holistic captioning model was never
trained with bounding-box annotations.}

\vspace{-30pt}
\rpi{\subsubsection{Grad-CAM for individual words of caption}}

\rpi{In our experiment we use the Show and Tell model \cite{vinyals_cvpr15} pre-trained on MSCOCO without fine-tuning through the visual representation obtained from Inception \cite{szegedy2016rethinking} architecture. 
In order to obtain \gcam{} map for individual words in the ground-truth caption we one-hot encode each of the visual words at the corresponding time-steps and compute the neuron importance score using Eq. \eqref{eq:alpha1} and combine with the convolution feature maps using Eq. \eqref{eq:gcam}. 
}

\vspace{5pt}
\noindent
\rpi{\textbf{Comparison to Human Attention}~
We manually created an object category to word mapping that maps object categories
like $<$person$>$ to a list of potential fine-grained labels like
[“child”, “man”, "woman", ...]. 
We map a total of 830 visual words existing in COCO captions to 80 COCO categories.
We then use the segmentation annotations for the 80 categories as human attention for this subset of matching words.
}

\rpi{We then use the pointing evaluation from \cite{zhang2016top}. 
For each visual word from the caption, we generate the \gcam{} map and then extract the maximally activated point. 
We then evaluate if the point lies within the human attention mapsegmentation for the corresponding COCO category, thereby counting it as a hit or a miss. 
The pointing accuracy is then calculated as \\$Acc = \frac{\#Hits}{\#Hits+\#Misses}$. 
We perform this experiment on 1000 randomly sampled images from COCO dataset and obtain an accuracy of 30.0\%. 
Some qualitative examples can be found in \reffig{fig:captioning_word}. 
}

\vspace{10pt}
\subsection{Visual Question Answering}\label{sec:vqa}

\begin{figure}[t]
    \begin{subfigure}[t]{\columnwidth}
    \includegraphics[scale=0.315]{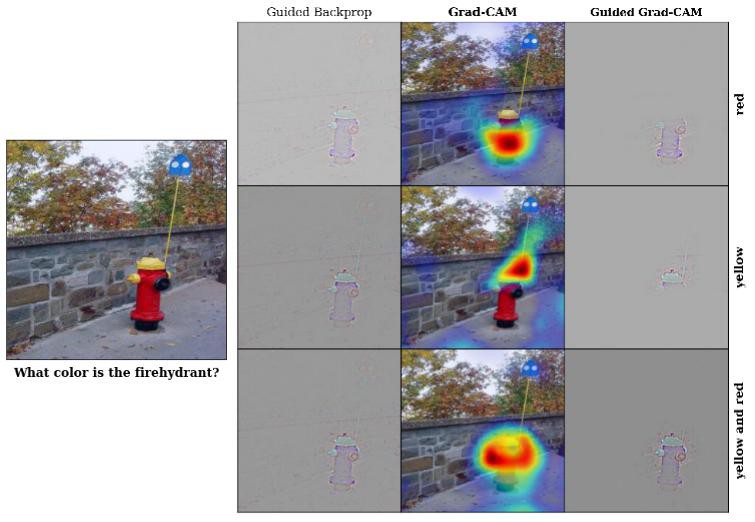}
    \vspace{5pt}
    \caption{\scriptsize{Visualizing VQA model from \cite{Lu2015}}}
    \vspace{35pt}
    \label{fig:vqa_main_fig}
    \end{subfigure}
    \begin{subfigure}[t]{\columnwidth}
    \includegraphics[scale=0.225]{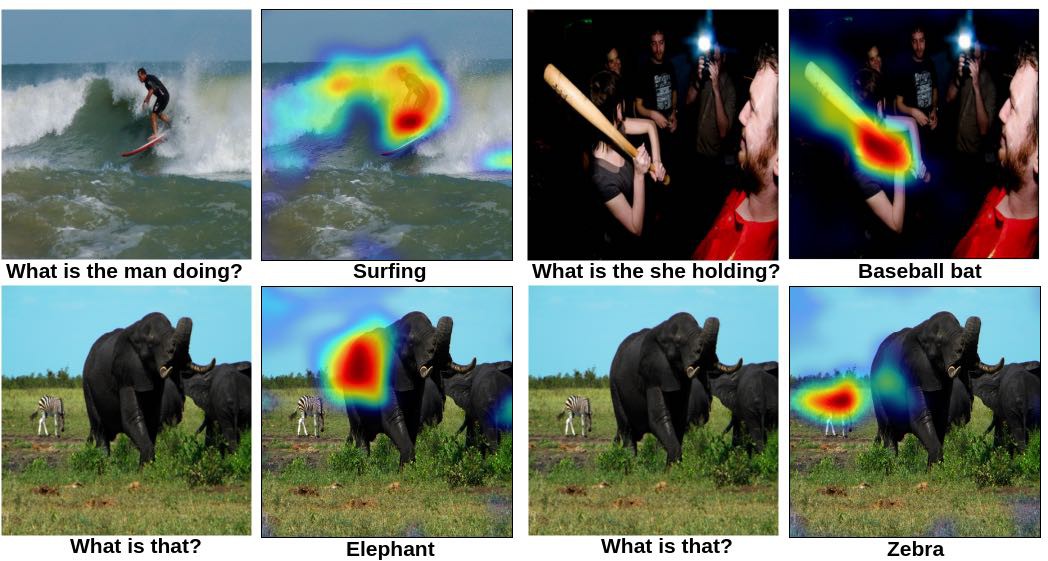}
    \vspace{10pt}
    \caption{\scriptsize{Visualizing ResNet based Hierarchical co-attention VQA model from \cite{Lu2016}}}
    
        \label{fig:vqa_residual_main}
    \end{subfigure}
	\vspace{14pt}
    \caption{Qualitative Results for our VQA experiments: (a) Given the image on the left and the question ``What color is the firehydrant?'', we visualize \gcam{}s and \cgb{}s for the answers ``red", ``yellow" and ``yellow and red".
        \gcam{} visualizations are highly interpretable and help explain any target prediction -- for ``red'', the model focuses on the bottom red part of the firehydrant; when forced to answer ``yellow'', the model concentrates on it`s top yellow cap, and when forced to answer ``yellow and red", it looks at the whole firehydrant! \rp{ (b) Our approach is capable of providing interpretable explanations even for complex models.}
    }
	\label{fig:vqa}
\end{figure}

Typical VQA pipelines ~\cite{antol2015vqa,gao2015you,malinowski_iccv15,ren_nips15}
consist of a CNN to process images and an RNN language model for questions.
The image and the question representations are fused to predict the answer,
typically with a $1000$-way classification ($1000$ being the size of the answer space).
Since this is a classification problem, we pick an answer (the score $y^c$ in \eqref{eq:scores})
and use its score to compute \gcam{} visualizations over the image to explain the answer.
Despite the complexity of the task, involving both visual and textual components,
the explanations (of the VQA model from Lu~\etal~\cite{Lu2015}) described in \reffig{fig:vqa} are surprisingly intuitive and informative.
\iccv{We quantify the performance of \gcam{} via correlation with occlusion maps,
as in \secref{sec:occ}. \gcam{} achieves a rank correlation (with occlusion maps)
of 0.60 $\pm$ 0.038 whereas \gb{} achieves 0.42 $\pm$ 0.038, indicating higher
faithfulness of our \gcam{} visualization.}

\noindent \para{Comparison to Human Attention}
Das~\etal~\cite{vqahat} collected human attention maps for a subset of the VQA dataset~\cite{antol2015vqa}.
These maps have high intensity where humans looked in the image in order to answer a visual question.
Human attention maps are compared to \gcam{} visualizations for the VQA model from~\cite{Lu2015}
on 1374 val question-image (QI) pairs from ~\cite{antol2015vqa} using the rank correlation evaluation protocol as in ~\cite{vqahat}.
\gcam{} and human attention maps have a correlation of 0.136, which is higher than chance or random attention maps (zero correlation).
This shows that despite not being trained on grounded image-text pairs, even non-attention based
CNN + LSTM based VQA models are surprisingly good at localizing regions for predicting a particular answer.

\noindent \para{Visualizing ResNet-based VQA model with co-attention}
    Lu~\etal~\cite{Lu2016} use a 200 layer ResNet~\cite{he_cvpr15} to encode the image, and jointly learn a hierarchical attention mechanism on the question and image. \reffig{fig:vqa_residual_main} shows \gcam{} visualizations for this network.
    \mac{As we visualize deeper layers of the ResNet, we see small changes in \gcam{}
    for most adjacent layers and larger changes between layers that involve dimensionality reduction.}
    \rpi{More visualizations for ResNets can be found in \refsec{sec:sup_resnet_analysis}}.
    To the best of our knowledge, we are the first to visualize decisions from ResNet-based models.

\vspace{-10pt}
\section{Conclusion}
\ad{
    In this work, we proposed a novel class-discriminative localization technique --
    Gradient-weighted Class Activation Mapping (\gcam{}) -- for making \emph{any}
    CNN-based model more transparent by producing visual explanations.
    Further, we combined \gcam{} localizations with existing high-resolution visualization
    techniques to obtain the best of both worlds -- high-resolution and class-discriminative
    \cgb{} visualizations.
}
Our visualizations outperform existing approaches on both axes -- interpretability
and faithfulness to original model. Extensive human studies reveal that our
visualizations can discriminate between classes more accurately, better expose
the trustworthiness of a classifier, and help identify biases in datasets.
Further, we devise a way to identify important neurons through Grad-CAM and provide a way to obtain textual explanations for model decisions.
Finally, we show the broad applicability of \gcam{} to various off-the-shelf
architectures for tasks such as image classification, image captioning and
visual question answering. 
We believe that a true AI system should not only be intelligent, but also be able
to reason about its beliefs and actions for humans to trust and use it.
Future work includes explaining decisions made by deep networks in domains such
as reinforcement learning, natural language processing and video applications.

\vspace{-10pt}
\section{Acknowledgements}

This work was funded in part by NSF CAREER awards to
DB and DP, DARPA XAI grant to DB and DP, ONR YIP awards to DP and DB, ONR Grant
N00014-14-1-0679 to DB, a Sloan Fellowship to DP, ARO
YIP awards to DB and DP, an Allen Distinguished Investigator
award to DP from the Paul G. Allen Family Foundation,
ICTAS Junior Faculty awards to DB and DP, Google
Faculty Research Awards to DP and DB, Amazon Academic
Research Awards to DP and DB, AWS in Education
Research grant to DB, and NVIDIA GPU donations to DB. The
views and conclusions contained herein are those of the authors
and should not be interpreted as necessarily representing
the official policies or endorsements, either expressed or
implied, of the U.S. Government, or any sponsor.

\vspace{-10pt}
\begin{appendices}

\section{Appendix Overview}

In the appendix, we provide:
 \begin{enumerate}[I]
\setlength{\itemsep}{1pt}
  \setlength{\parskip}{0pt}
  \setlength{\parsep}{0pt}
 \item - Ablation studies evaluating our design choices
 \item - More qualitative examples for image classification, captioning and VQA
 \item - More details of Pointing Game evaluation technique
 \item - Qualitative comparison to existing visualization techniques
 \item - More qualitative examples of textual explanations
 \end{enumerate}

\vspace{-15pt}
\section{Ablation studies} \label{sec:ablation}

We perform several ablation studies to explore and validate our design choices for computing \gcam{} visualizations. 
This includes visualizing different layers in the network, understanding importance of ReLU in \eqref{eq:gcam}, analyzing different types of gradients (for ReLU backward pass), and different gradient pooling strategies.

\vspace{-8pt}
\subsection*{1. \gcam{} for different layers}

\begin{figure*}[t!]
	\centering
    \includegraphics[width=1\textwidth]{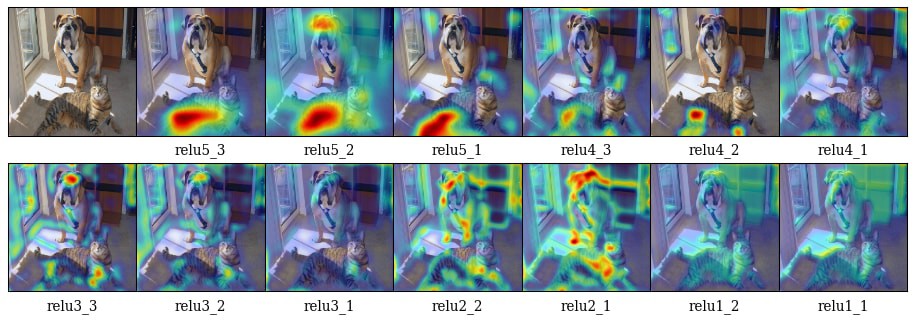}\\
    \vspace{-5pt}
		\label{fig:vgg16conv}
        \vspace{5pt}
		\caption{ \gcam{} at different convolutional layers for the `tiger cat' class.
            This figure analyzes how localizations change qualitatively as we perform \gcam{} with respect to different feature maps in a CNN (VGG16~\cite{simonyan_arxiv14}).
	We find that the best looking visualizations are often obtained after the deepest convolutional layer in the network, and localizations get progressively worse at shallower layers.
This is consistent with our intuition described in Section 3 of main paper, that deeper convolutional layer capture more semantic concepts. %
}
	\label{fig:conv_layers}
\end{figure*}

We show \gcam{} visualizations for the ``tiger-cat'' class at different convolutional
layers in AlexNet and VGG-16.
As expected, the results from \reffig{fig:conv_layers} show that localization
becomes progressively worse as we move to earlier convolutional layers.
This is because later convolutional layers better capture high-level semantic information
while retaining spatial information than earlier layers, that have smaller
receptive fields and only focus on local features.

\begin{figure*}
 \centering
 \includegraphics[width=1\linewidth]{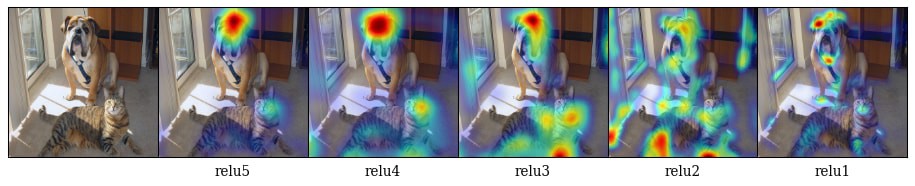}
 \caption{\gcam{} localizations for ``tiger cat'' category for different rectified convolutional layer feature maps for AlexNet.}
 \label{fig:alex_conv}
\end{figure*}

\vspace{-5pt}
\subsection*{2. Design choices}

\vspace{-5pt}
\begin{table}[h!]
\centering
\captionsetup{belowskip=0pt,aboveskip=4pt}
    {
        \begin{tabular}{c@{\hskip 0.6in} c}\toprule
            \textbf{Method} & \textbf{Top-1 Loc error} \\
            \midrule

            \gcam{}                                                  & 59.65          \\
            \midrule
            \gcam{} without ReLU in Eq.1                             & 74.98          \\
            \gcam{} with Absolute gradients                          & 58.19          \\
            \gcam{} with GMP gradients                               & 59.96          \\
            \gcam{} with Deconv ReLU                                 & 83.95          \\
            \gcam{} with Guided ReLU                                 & 59.14          \\
            \bottomrule
        \end{tabular}

    }

    \caption{Localization results on ILSVRC-15 val for the ablations. Note that
    this evaluation is over 10 crops, while visualizations are single crop.}

    \label{table:locresablation}
\end{table}

We evaluate different design choices via top-1 localization errors on the ILSVRC-15
val set~\cite{imagenet_cvpr09}. See \reftab{table:locresablation}.
\subsubsection*{2.1. Importance of ReLU in \eqref{eq:scores}}

Removing ReLU (\eqref{eq:scores}) increases error by 15.3\%.
Negative values in \gcam{} indicate confusion between multiple occurring classes.

\subsubsection*{2.2. Global Average Pooling \vs Global Max Pooling}

Instead of Global Average Pooling (GAP) the incoming gradients to the convolutional layer,
we tried Global Max Pooling (GMP). We observe that using GMP lowers the localization ability
of \gcam{}. An example can be found in \reffig{fig:gmp} below. This may be due to the fact
that max is statistically less robust to noise compared to the averaged gradient.

\begin{figure}[h]

     \centering
     \includegraphics[width=1\linewidth]{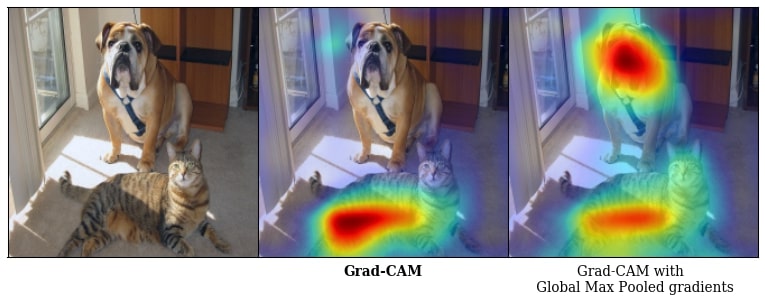}
     \caption{\gcam{} visualizations for ``tiger cat'' category with Global Average Pooling and Global Max Pooling.}
     \label{fig:gmp}

   \end{figure}

\subsubsection*{2.3. Effect of different ReLU on \gcam{}}

We experiment with Guided-ReLU~\cite{springenberg_arxiv14} and Deconv-ReLU~\cite{zeiler_eccv14}
as modifications to the backward pass of ReLU.

\begin{figure}[h]

     \centering
     \includegraphics[width=1\linewidth]{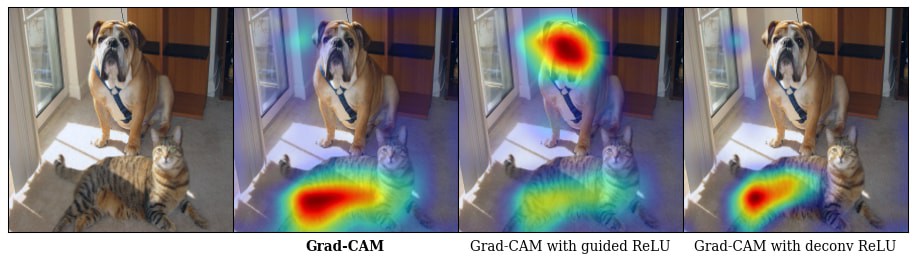}
     \caption{\gcam{} visualizations for ``tiger cat'' category for different modifications to the ReLU backward pass. The best results are obtained when we use the actual gradients during the computation of \gcam{}.}
     \label{fig:relu}
   \end{figure}

\textbf{Guided-ReLU:} Springenberg~\etal\cite{springenberg_arxiv14} introduced Guided Backprop,
where the backward pass of ReLU is modified to only pass positive gradients
to regions of positive activations.
Applying this change to the computation of \gcam{} introduces a drop in the
class-discriminative ability as can be seen in \figref{fig:relu},
but it marginally improves localization performance as can be seen in \reftab{table:locresablation}.

\textbf{Deconv-ReLU:} In \dec{}~\cite{zeiler_eccv14}, Zeiler and Fergus introduced a
modification to the backward pass of ReLU to only pass positive gradients.
Applying this modification to the computation of \gcam{} leads to worse results
(\figref{fig:relu}). This indicates that negative gradients also carry
important information for class-discriminativeness.

\vspace{-5pt}
\section{Qualitative results for vision and language tasks}\label{sec:sup_experiments}

In this section we provide more qualitative results for \gcam{} and \cgb{} applied to the task of image classification, image captioning and VQA.
\vspace{-5pt}
\subsection*{1. Image Classification}

\begin{figure*}
 \centering
 \includegraphics[width=1\linewidth]{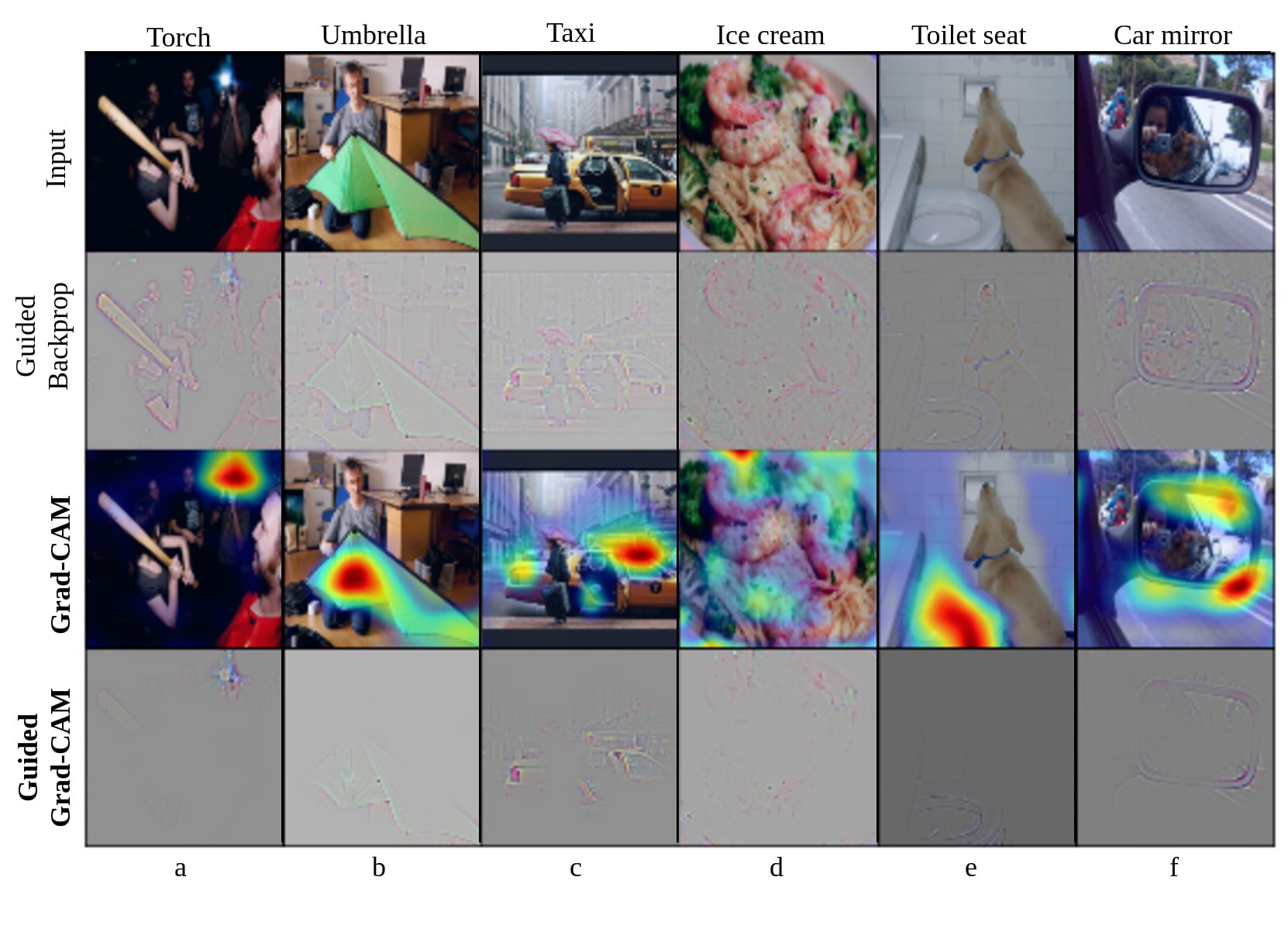}
 \caption{Visualizations for randomly sampled images from the COCO validation dataset.
 Predicted classes are mentioned at the top of each column.}
 \label{fig:classification}
\end{figure*}

We use \gcam{} and \cgb{} to visualize the regions of the image that provide support for a particular prediction.
The results reported in \reffig{fig:classification} correspond to the VGG-16 \cite{simonyan_arxiv14} network trained on ImageNet.

\reffig{fig:classification} shows randomly sampled examples from COCO~\cite{Lin_2014} validation set.
COCO images typically have multiple objects per image and \gcam{} visualizations show precise localization to support the model's prediction.

\cgb{} can even localize tiny objects. For example our approach correctly localizes the predicted class ``torch'' (\reffig{fig:classification}.a) inspite of its size and odd location in the image. Our method is also class-discriminative -- it places attention \emph{only} on the ``toilet seat'' even when a popular ImageNet category ``dog'' exists in the image (\reffig{fig:classification}.e).

We also visualized Grad-CAM, Guided Backpropagation (GB), Deconvolution (DC), GB + \gcam{} (\cgb{}), DC + \gcam{} (\cdec{}) for images from the ILSVRC13 detection val set that have at least 2 unique object categories each.
The visualizations for the mentioned class can be found in the following links.

``computer keyboard, keypad'' class: \\http://i.imgur.com/QMhsRzf.jpg

``sunglasses, dark glasses, shades'' class: \\ http://i.imgur.com/a1C7DGh.jpg

\vspace{-10pt}
\subsection*{2. Image Captioning}

\begin{figure*}
     \centering
     \includegraphics[width=0.83\linewidth]{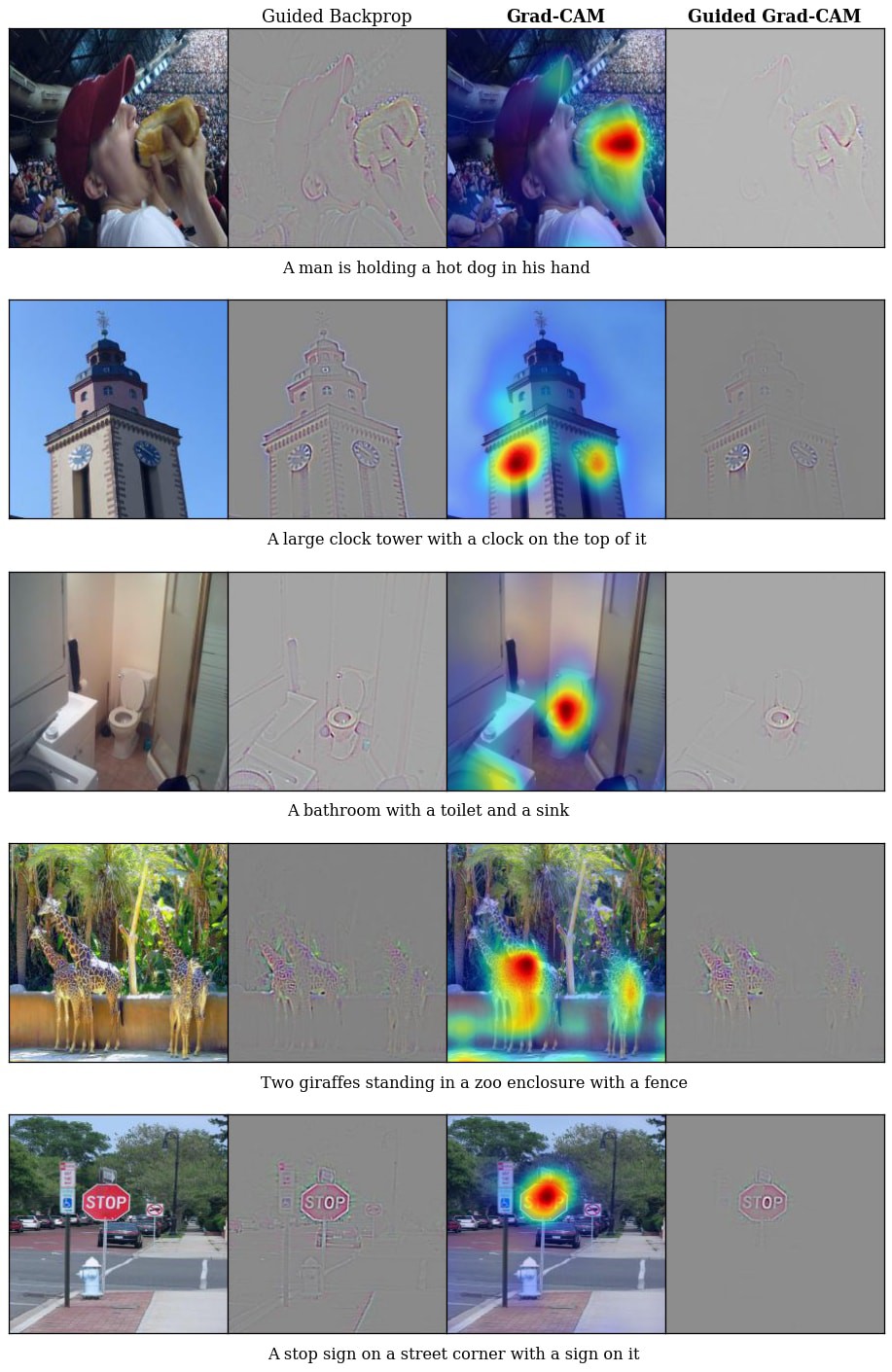}
     \caption{\gb{}, \gcam{}  and \cgb{}  visualizations for the captions produced by the Neuraltalk2 image captioning model.}
     \label{fig:sup_captioning}
\end{figure*}

\begin{figure*}
     \centering
     \includegraphics[scale=0.30]{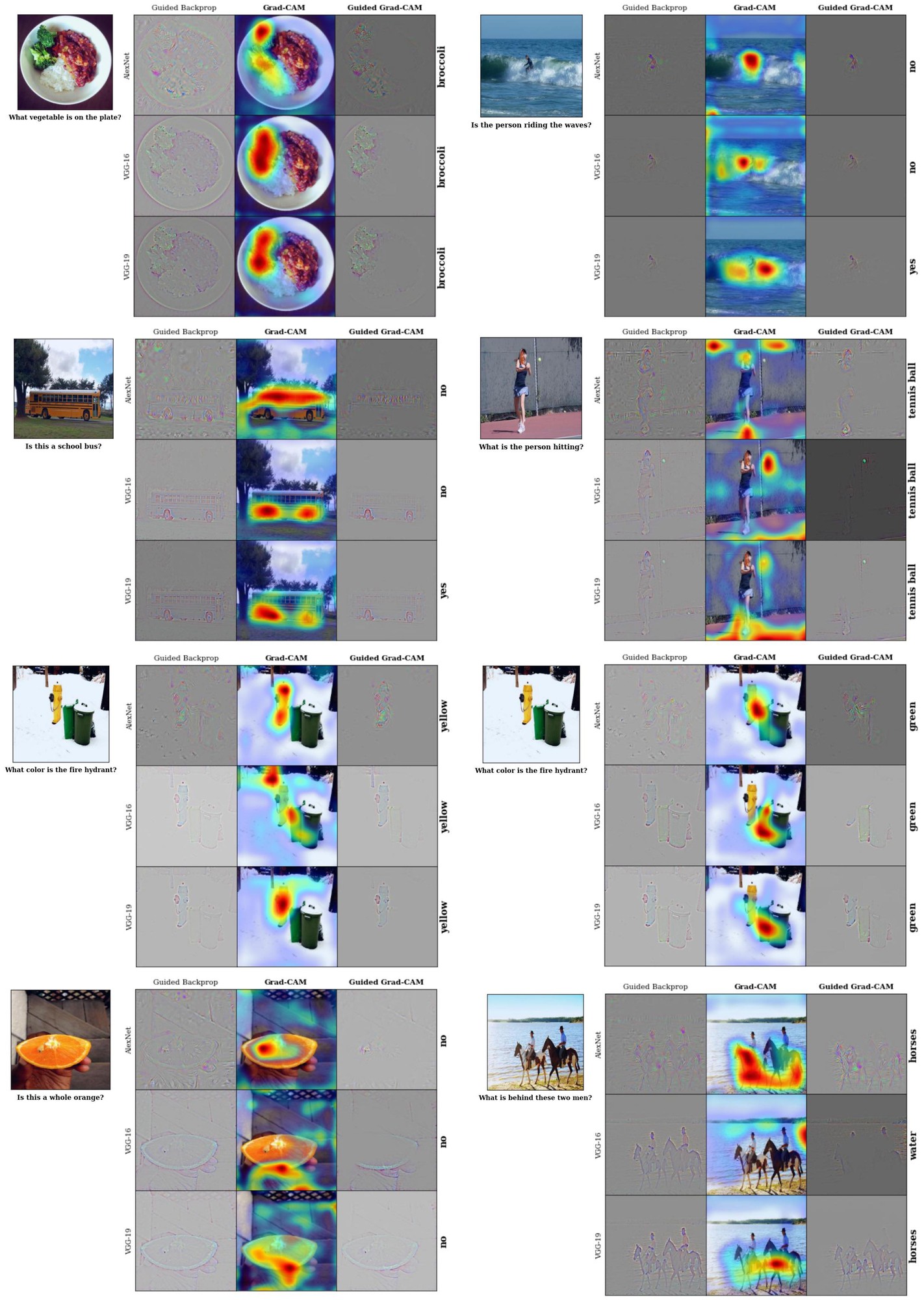}
     \caption{\gb{}, \gcam{} and \cgb{}  visualizations for the answers from a VQA model.
     For each image-question pair, we show visualizations for AlexNet, VGG-16 and VGG-19.
     Notice how the attention changes in row 3, as we change the answer from \emph{Yellow} to \emph{Green}.}
     \label{fig:vqa_supp}
\end{figure*}

\begin{figure*}[htp]
 \centering
 \includegraphics[width=0.65\linewidth]{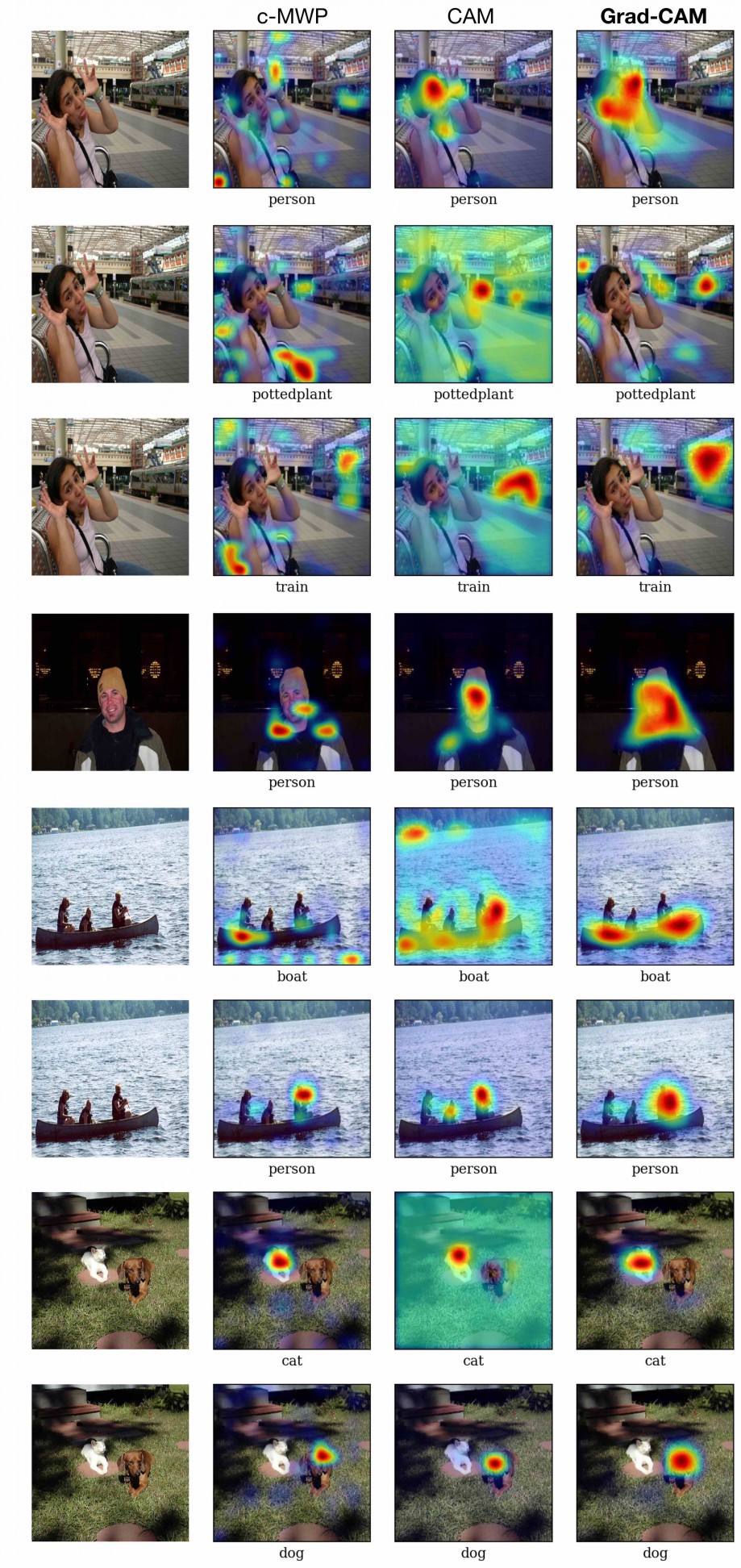}
 \caption{Visualizations for ground-truth categories (shown below each image) for images sampled from the PASCAL~\cite{pascal-voc-2007} validation set.}
 \label{fig:gcam_pascal}
\end{figure*}

We use the publicly available Neuraltalk2 code and model\footnote{\url{https://github.com/karpathy/neuraltalk2}} for our image captioning experiments.
The model uses VGG-16 to encode the image.
The image representation is passed as input at the first time step to an LSTM that generates a caption for the image.
The model is trained end-to-end along with CNN finetuning using the COCO~\cite{Lin_2014} Captioning dataset.
We feedforward the image to the image captioning model to obtain a caption. \rp{We use \gcam{} to get a coarse localization and combine it with \gb{} to get a high-resolution visualization that highlights regions in the image that provide support for the generated caption.} %

\subsection*{3. Visual Question Answering (VQA)}

\rp{We use \gcam{} and \cgb{} to explain why \rp{a} publicly available VQA model \cite{Lu2015} answered what it answered.}

The VQA model by Lu~\etal uses a standard CNN followed by a fully connected layer to transform the image to 1024-dim to match the \rp{LSTM} embeddings of the question. Then the transformed image and LSTM embeddings are pointwise multiplied to get a combined representation of the image and question and a multi-layer perceptron is trained on top to predict one among 1000 answers.
We show visualizations for the VQA model trained with 3 different CNNs - AlexNet~\cite{krizhevsky_nips12}, VGG-16 and VGG-19~\cite{simonyan_arxiv14}.
Even though the CNNs were not finetuned for the task of VQA, it is interesting to see how our approach can serve as a tool to understand these networks better by providing a localized high-resolution visualization of the regions the model is looking at. Note that these networks were trained with no explicit attention mechanism enforced.

Notice in the first row of \reffig{fig:vqa_supp}, for the question, ``\emph{Is the person riding the waves?}'', the VQA model with AlexNet and VGG-16 answered ``No'', as they concentrated on the person mainly, and not the waves.
On the other hand, VGG-19 correctly answered ``Yes'', and it looked at the regions around the man in order to answer the question.
In the second row, for the question, ``\emph{What is the person hitting?}'', the VQA model trained with AlexNet answered ``Tennis ball'' just based on context without looking at the ball.
Such a model might be risky when employed in real-life scenarios.
It is difficult to determine the trustworthiness of a model just based on the predicted answer.
Our visualizations provide an accurate way to explain the model's predictions and help in determining which model to trust, without making any architectural changes or sacrificing accuracy.
Notice in the last row of \reffig{fig:vqa_supp}, for the question, ``\emph{Is this a whole orange?}'', the model looks for regions around the orange to answer ``No''.

\begin{figure*}[h]
\begin{center}
\begin{subfigure}[t]{\columnwidth}
\includegraphics[scale=0.135]{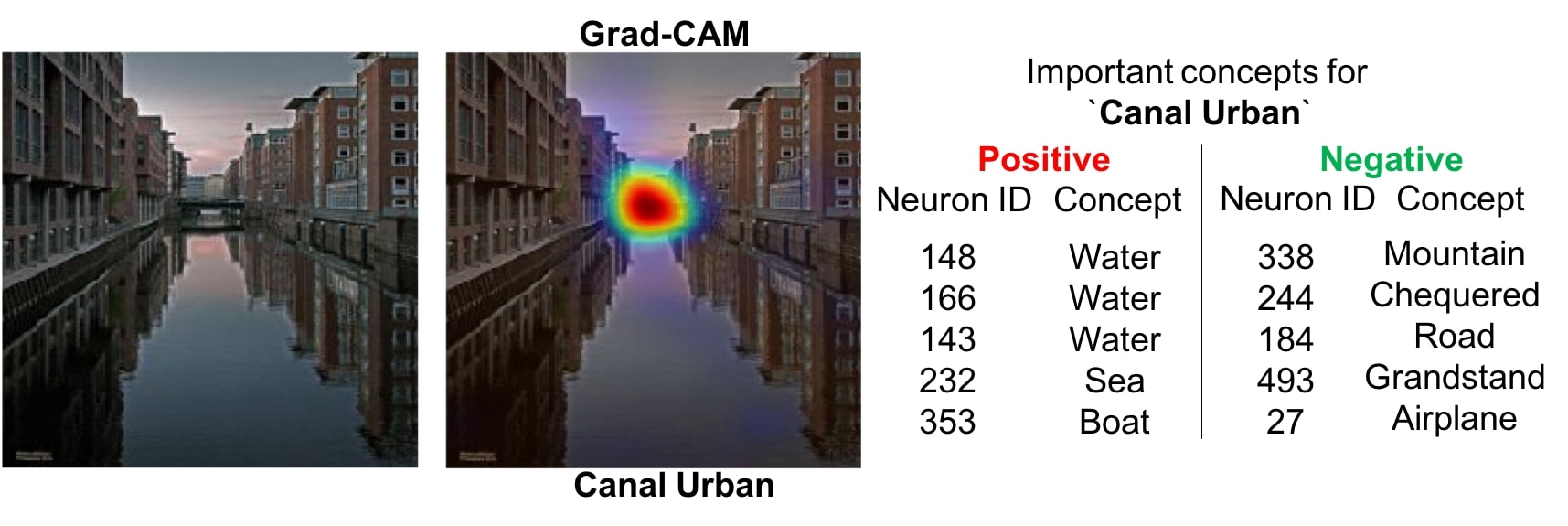}\caption{}
\vspace{10pt}
\end{subfigure}
\begin{subfigure}[t]{\columnwidth}
\includegraphics[scale=0.135]{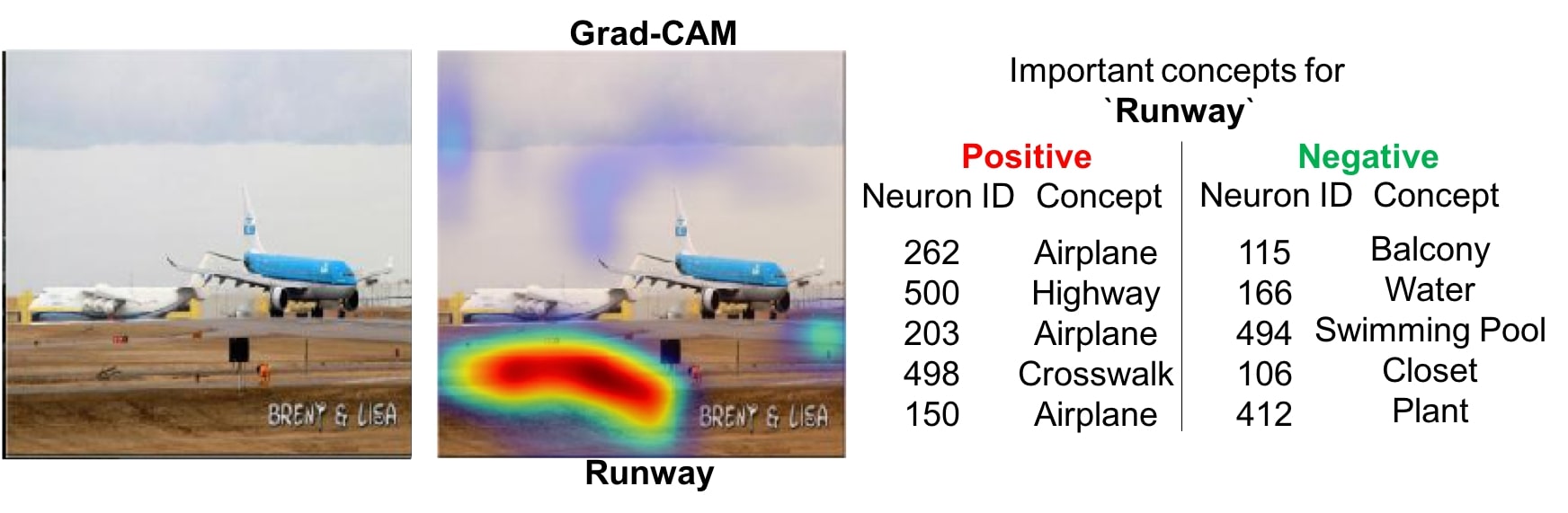}\caption{}
\vspace{10pt}
\end{subfigure}
\begin{subfigure}[t]{\columnwidth}
\includegraphics[scale=0.135]{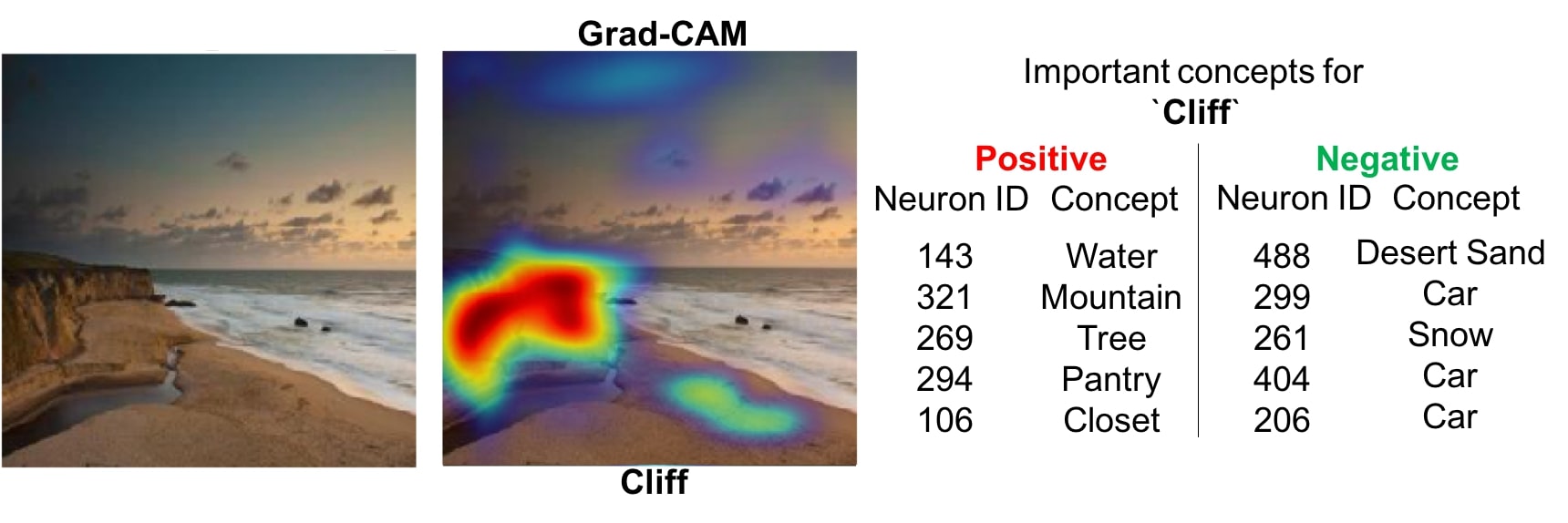}\caption{}
\vspace{10pt}
\end{subfigure}
\begin{subfigure}[t]{\columnwidth}
\includegraphics[scale=0.135]{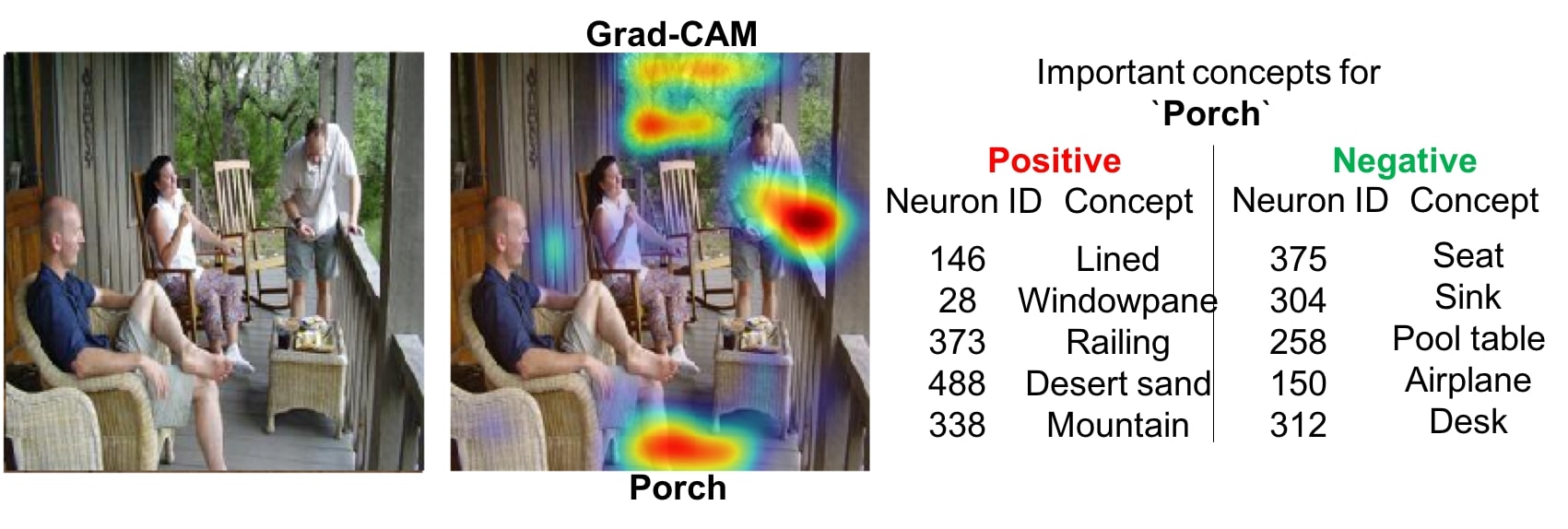}\caption{}
\vspace{10pt}
\end{subfigure}
\begin{subfigure}[t]{\columnwidth}
\includegraphics[scale=0.135]{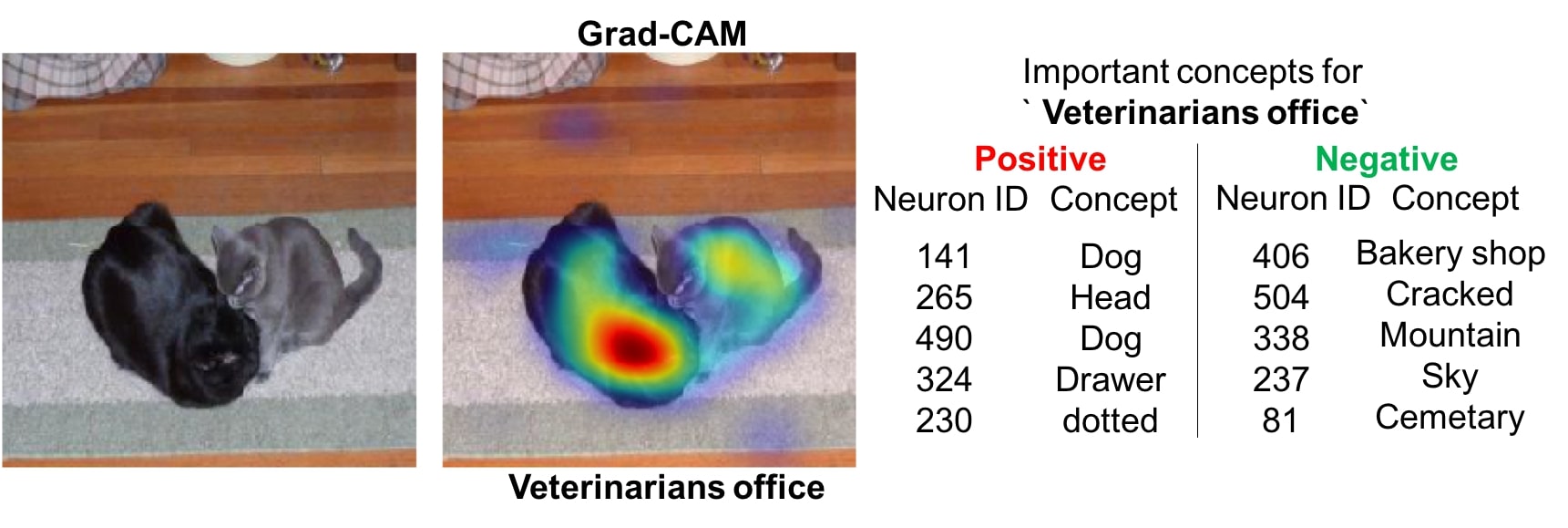}\caption{}
\vspace{10pt}
\end{subfigure}
\begin{subfigure}[t]{\columnwidth}
\includegraphics[scale=0.135]{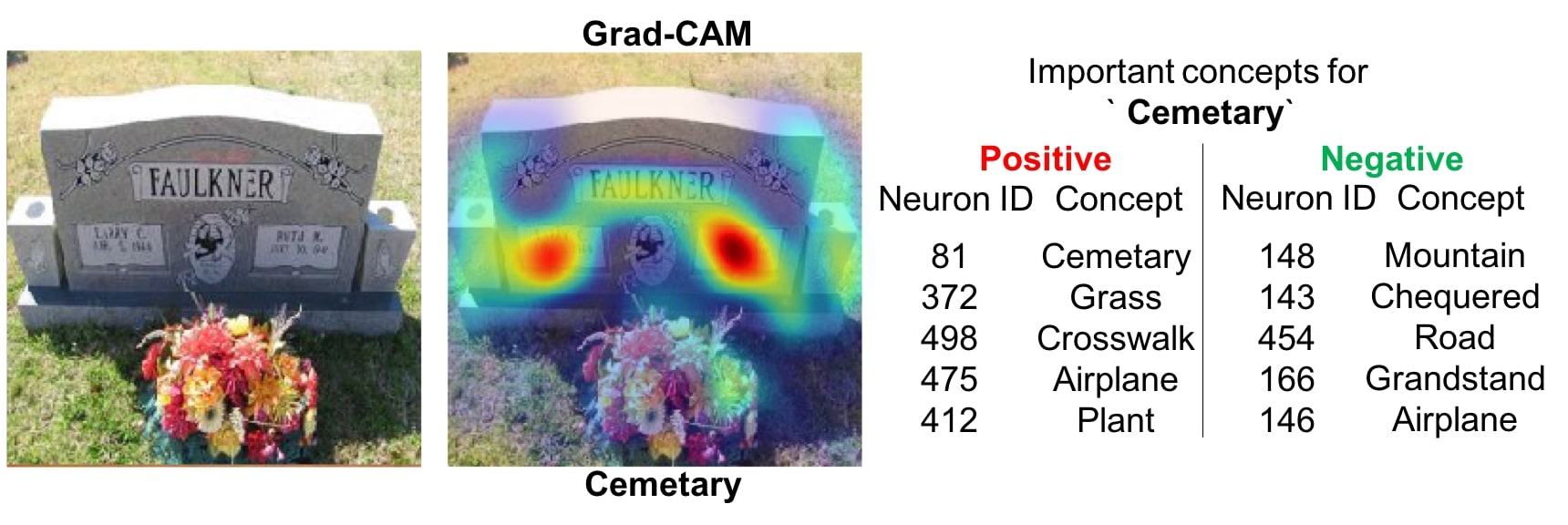}\caption{}
\vspace{10pt}
\end{subfigure}
\begin{subfigure}[t]{\columnwidth}
\includegraphics[scale=0.135]{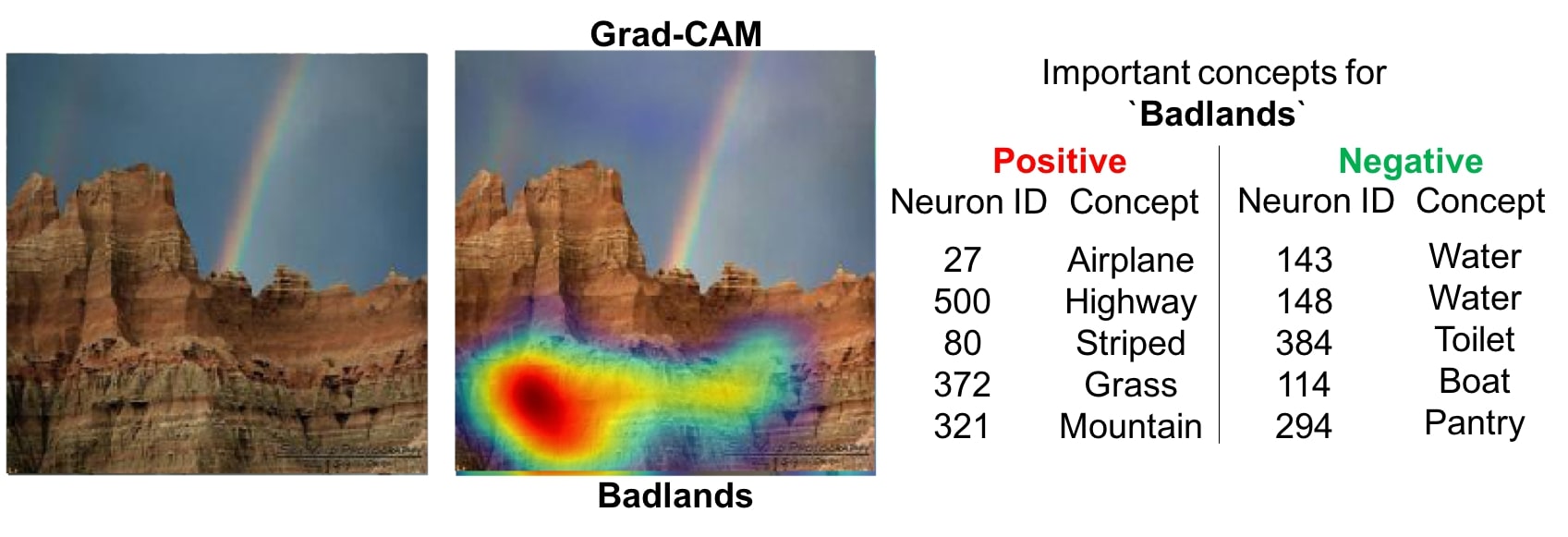}\caption{}
\vspace{10pt}
\end{subfigure}
\begin{subfigure}[t]{\columnwidth}
\includegraphics[scale=0.135]{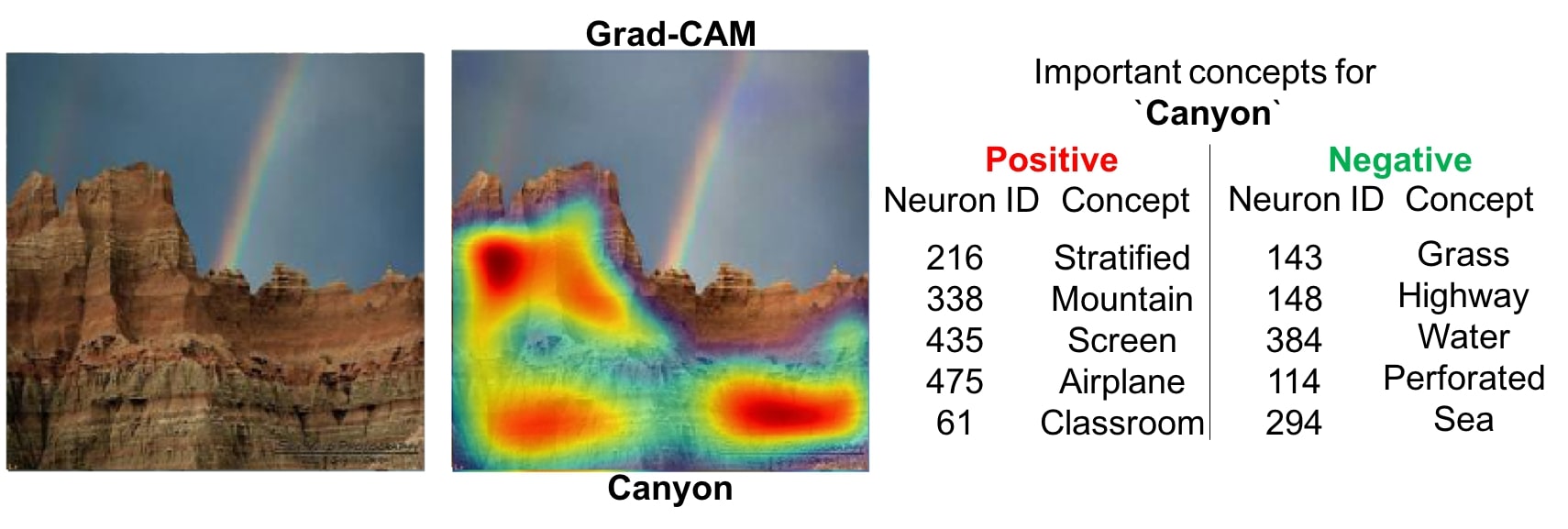}\caption{}
\vspace{10pt}
\end{subfigure}
\begin{subfigure}[t]{\columnwidth}
\includegraphics[scale=0.135]{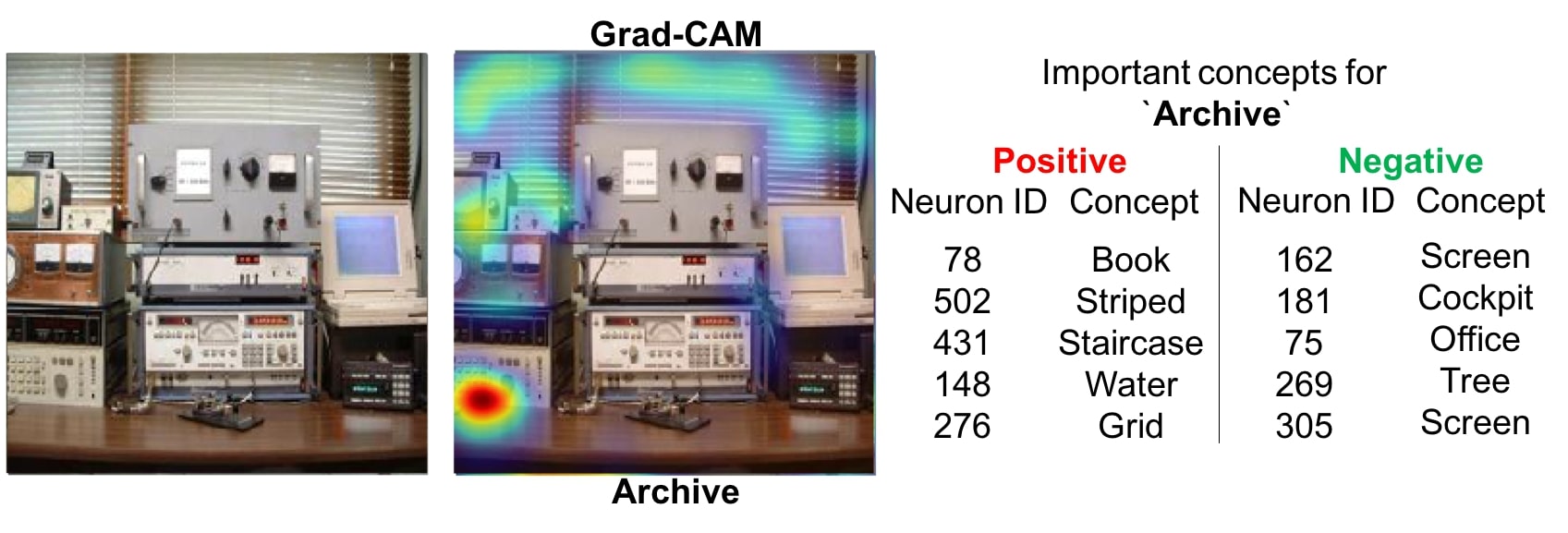}\caption{}
\vspace{10pt}
\end{subfigure}
\begin{subfigure}[t]{\columnwidth}
\includegraphics[scale=0.135]{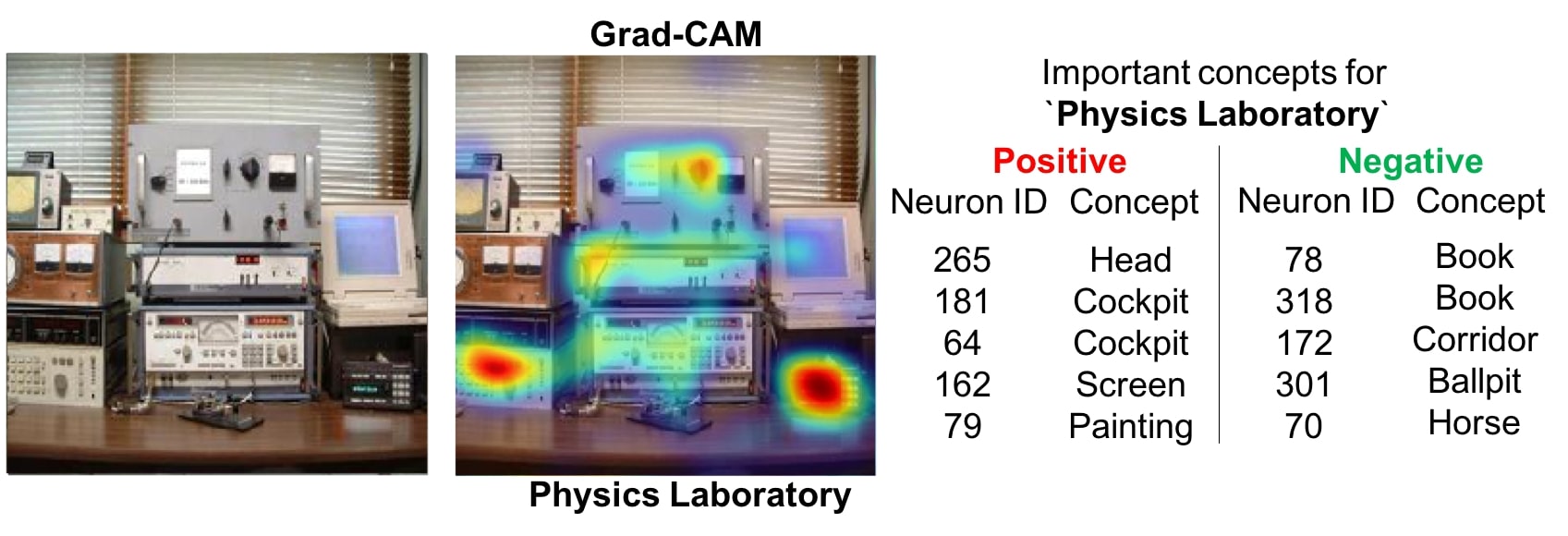}\caption{}
\vspace{10pt}
\end{subfigure}
  \caption{More Qualitative examples showing visual explanations and textual explanations for VGG-16 trained on Places365 dataset (\cite{zhou2017places}). For textual explanations we provide the most important neurons for the predicted class along with their names. Important neurons can be either be persuasive (positively important) or inhibitive (negatively important). 
  The first 3 rows show positive examples, and the last 2 rows show failure cases.}
  \label{fig:sup_text_explanations}
\end{center}
\end{figure*}

\vspace{-10pt}
\section{More details of Pointing Game}\label{sec:sup_pointing}

In \cite{zhang2016top}, the pointing game was setup to evaluate the discriminativeness of different attention maps for localizing ground-truth categories. In a sense, this evaluates the precision of a visualization, \ie how often does the attention map intersect the segmentation map of the ground-truth category.
This does not evaluate how often the visualization technique produces maps which do not correspond to the category of interest. 

Hence we propose a modification to the pointing game to evaluate visualizations of the top-5 predicted category.
In this case the visualizations are given an additional option to reject any of the top-5 predictions from the CNN classifiers.
For each of the two visualizations, \gcam{} and c-MWP, we choose a threshold on the max value of the visualization, that can be used to determine if the category being visualized exists in the image.

We compute the maps for the top-5 categories, and based on the maximum value in the map, we try to classify if the map is of the GT label or a category that is absent in the image.
As mentioned in Section 4.2 of the main paper, we find that our approach \gcam{} outperforms c-MWP by a significant margin (70.58\% vs 60.30\% on VGG-16).

\vspace{-10pt}
\section{Qualitative comparison to Excitation Backprop (c-MWP) and CAM}\label{sec:sup_comparison}
\vspace{-5pt}
In this section we provide more qualitative results comparing \gcam{} with CAM~\cite{zhou_cvpr16} and c-MWP~\cite{zhang2016top} on Pascal~\cite{pascal-voc-2007}. %

We compare \gcam{}, CAM and c-MWP visualizations from ImageNet trained VGG-16 models finetuned on PASCAL VOC 2012 dataset.
While \gcam{} and c-MWP visualizations can be directly obtained from existing models, CAM requires an architectural change, and requires re-training, which leads to loss in accuracy.
Also, unlike \gcam{}, c-MWP and CAM can only be applied for image classification networks.
Visualizations for the ground-truth categories can be found in \reffig{fig:gcam_pascal}.

\begin{figure*}
     \centering
        \begin{subfigure}[t]{0.49\textwidth}
        \begin{center}
     \includegraphics[width=\linewidth]{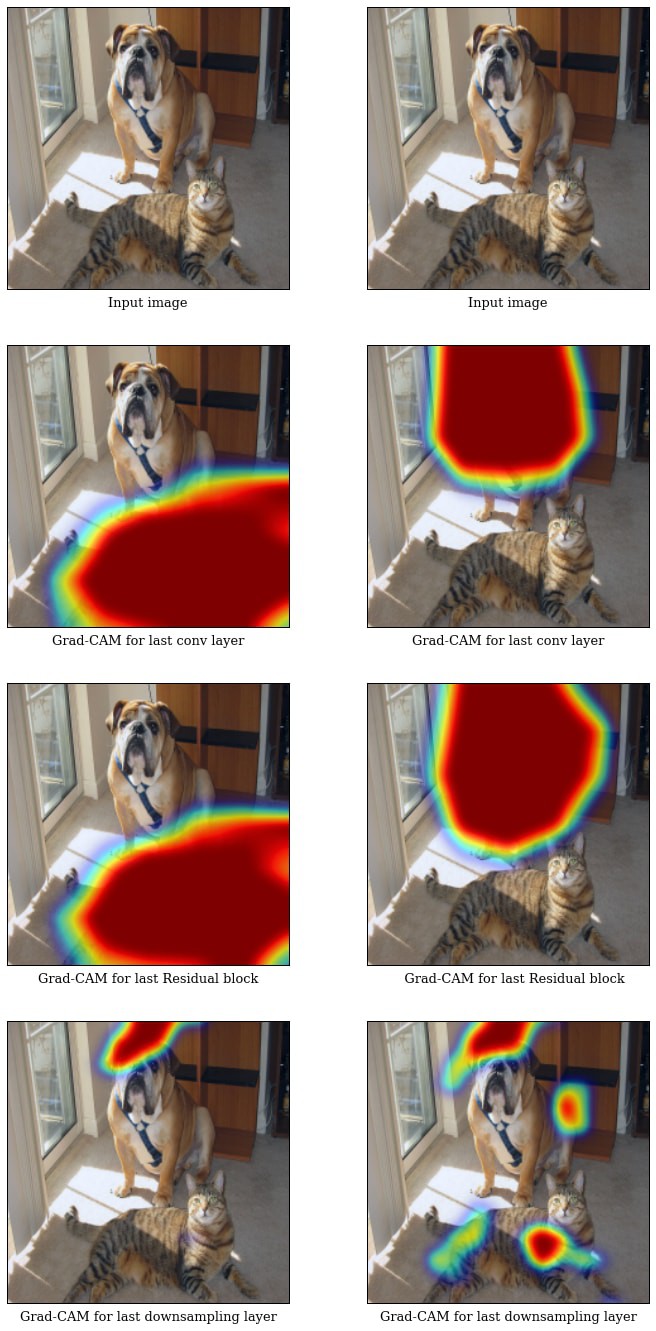}
     \caption{\gcam{}  visualizations for the ResNet-200 layer architecture for 'tiger cat'(left) and 'boxer'(right) category.}
 \end{center}
    \end{subfigure}
    \begin{subfigure}[t]{0.49\textwidth}
        \begin{center}
     \includegraphics[width=\linewidth]{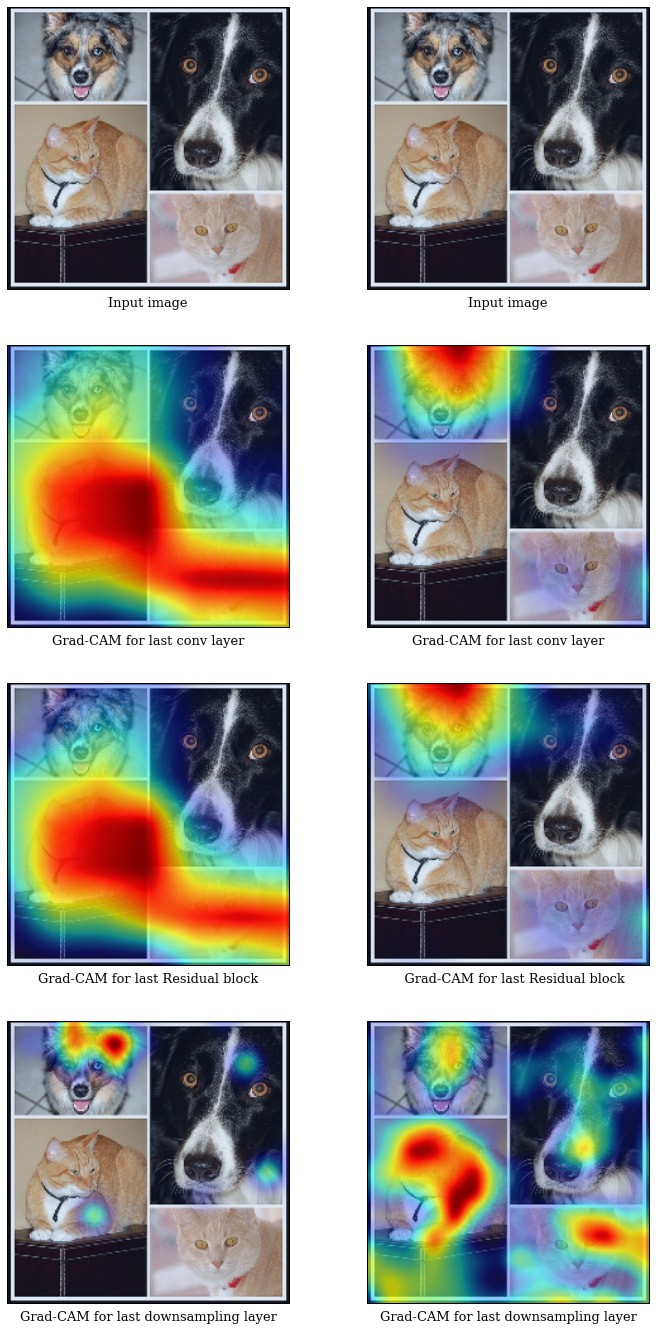}
     \caption{\gcam{}  visualizations for the ResNet-200 layer architecture for 'tabby cat'(left) and 'boxer'(right) category.}
        \end{center}
    \end{subfigure}
    \vspace{18pt}
    \caption{We observe that the discriminative ability of \gcam{} significantly reduces as we encounter the downsampling layer.}
     \label{fig:sup_resnet}
\end{figure*}

\vspace{-10pt}
\section{Visual and Textual explanations for Places dataset}\label{sec:sup_text_exp}
\vspace{-15pt}
\reffig{fig:sup_text_explanations} shows more examples of visual and textual explanations (\refsec{sec:text_exp}) for the image classification model (VGG-16) trained on Places365 dataset (\cite{zhou2017places}).

\vspace{-18pt}
\section{Analyzing Residual Networks}\label{sec:sup_resnet_analysis}
\vspace{-5pt}

In this section, we perform Grad-CAM on Residual Networks (ResNets). In particular, we analyze the 200-layer architecture trained on ImageNet\footnote{We use the 200-layer ResNet architecture from \url{https://github.com/facebook/fb.resnet.torch.}}.

Current ResNets~\cite{he_cvpr15} typically consist of residual blocks.
One set of blocks use identity skip connections (shortcut connections between two layers having identical output dimensions).
These sets of residual blocks are interspersed with downsampling modules that alter dimensions of propagating signal.
As can be seen in \reffig{fig:sup_resnet} our visualizations applied on the last convolutional layer can correctly localize the cat and the dog.
\gcam{} can also visualize the cat and dog correctly in the residual blocks of the last set.
However, as we go towards earlier sets of residual blocks with different spatial resolution, we see that \gcam{} fails to localize the category of interest (see last row of \reffig{fig:sup_resnet}).
We observe similar trends for other ResNet architectures (18 and 50-layer).

\end{appendices}

\vspace{-5pt}
{\footnotesize
\bibliographystyle{ieee}
\bibliography{strings,main}

\begin{thebibliography}{10}\itemsep=-1pt

\bibitem{agrawal2016analyzing}
A.~Agrawal, D.~Batra, and D.~Parikh.
\newblock {Analyzing the Behavior of Visual Question Answering Models}.
\newblock In {\em EMNLP}, 2016.

\bibitem{agrawal2015cloudcv}
H.~Agrawal, C.~S. Mathialagan, Y.~Goyal, N.~Chavali, P.~Banik, A.~Mohapatra,
  A.~Osman, and D.~Batra.
\newblock {CloudCV: Large Scale Distributed Computer Vision as a Cloud
  Service}.
\newblock In {\em Mobile Cloud Visual Media Computing}, pages 265--290.
  Springer, 2015.

\bibitem{antol2015vqa}
S.~Antol, A.~Agrawal, J.~Lu, M.~Mitchell, D.~Batra, C.~Lawrence~Zitnick, and
  D.~Parikh.
\newblock {VQA: Visual Question Answering}.
\newblock In {\em ICCV}, 2015.

\bibitem{netdissect}
D.~Bau, B.~Zhou, A.~Khosla, A.~Oliva, and A.~Torralba.
\newblock Network dissection: Quantifying interpretability of deep visual
  representations.
\newblock In {\em Computer Vision and Pattern Recognition}, 2017.

\bibitem{bazzani2016self}
L.~Bazzani, A.~Bergamo, D.~Anguelov, and L.~Torresani.
\newblock Self-taught object localization with deep networks.
\newblock In {\em WACV}, 2016.

\bibitem{bengio2013representation}
Y.~Bengio, A.~Courville, and P.~Vincent.
\newblock Representation learning: A review and new perspectives.
\newblock {\em IEEE transactions on pattern analysis and machine intelligence},
  35(8):1798--1828, 2013.

\bibitem{chen2015microsoft}
X.~Chen, H.~Fang, T.-Y. Lin, R.~Vedantam, S.~Gupta, P.~Doll{\'a}r, and C.~L.
  Zitnick.
\newblock {Microsoft COCO captions: Data Collection and Evaluation Server}.
\newblock {\em arXiv preprint arXiv:1504.00325}, 2015.

\bibitem{cinbis2016weakly}
R.~G. Cinbis, J.~Verbeek, and C.~Schmid.
\newblock Weakly supervised object localization with multi-fold multiple
  instance learning.
\newblock {\em IEEE transactions on pattern analysis and machine intelligence},
  2016.

\bibitem{vqahat}
A.~Das, H.~Agrawal, C.~L. Zitnick, D.~Parikh, and D.~Batra.
\newblock {Human Attention in Visual Question Answering: Do Humans and Deep
  Networks Look at the Same Regions?}
\newblock In {\em EMNLP}, 2016.

\bibitem{embodiedqa}
A.~Das, S.~Datta, G.~Gkioxari, S.~Lee, D.~Parikh, and D.~Batra.
\newblock {E}mbodied {Q}uestion {A}nswering.
\newblock In {\em Proceedings of the IEEE Conference on Computer Vision and
  Pattern Recognition (CVPR)}, 2018.

\bibitem{visdial}
A.~Das, S.~Kottur, K.~Gupta, A.~Singh, D.~Yadav, J.~M. Moura, D.~Parikh, and
  D.~Batra.
\newblock {V}isual {D}ialog.
\newblock In {\em Proceedings of the IEEE Conference on Computer Vision and
  Pattern Recognition (CVPR)}, 2017.

\bibitem{visdial_rl}
A.~Das, S.~Kottur, J.~M. Moura, S.~Lee, and D.~Batra.
\newblock Learning cooperative visual dialog agents with deep reinforcement
  learning.
\newblock In {\em Proceedings of the IEEE International Conference on Computer
  Vision (ICCV)}, 2017.

\bibitem{guesswhat}
H.~de~Vries, F.~Strub, S.~Chandar, O.~Pietquin, H.~Larochelle, and A.~C.
  Courville.
\newblock Guesswhat?! visual object discovery through multi-modal dialogue.
\newblock In {\em Proceedings of the IEEE Conference on Computer Vision and
  Pattern Recognition (CVPR)}, 2017.

\bibitem{imagenet_cvpr09}
J.~Deng, W.~Dong, R.~Socher, L.-J. Li, K.~Li, and L.~Fei-Fei.
\newblock {ImageNet: A Large-Scale Hierarchical Image Database}.
\newblock In {\em CVPR}, 2009.

\bibitem{dosovitskiy_cvpr16}
A.~Dosovitskiy and T.~Brox.
\newblock {Inverting Convolutional Networks with Convolutional Networks}.
\newblock In {\em CVPR}, 2015.

\bibitem{erhan2009visualizing}
D.~Erhan, Y.~Bengio, A.~Courville, and P.~Vincent.
\newblock {Visualizing Higher-layer Features of a Deep Network}.
\newblock {\em University of Montreal}, 1341, 2009.

\bibitem{pascal-voc-2007}
M.~Everingham, L.~Van~Gool, C.~K.~I. Williams, J.~Winn, and A.~Zisserman.
\newblock The {PASCAL} {V}isual {O}bject {C}lasses {C}hallenge 2007 {(VOC2007)}
  {R}esults.
\newblock
  http://www.pascal-network.org/challenges/VOC/voc2007/workshop/index.html,
  2009.

\bibitem{fang2015captions}
H.~Fang, S.~Gupta, F.~Iandola, R.~K. Srivastava, L.~Deng, P.~Doll{\'a}r,
  J.~Gao, X.~He, M.~Mitchell, J.~C. Platt, et~al.
\newblock {From Captions to Visual Concepts and Back}.
\newblock In {\em CVPR}, 2015.

\bibitem{Gan_2015_CVPR}
C.~Gan, N.~Wang, Y.~Yang, D.-Y. Yeung, and A.~G. Hauptmann.
\newblock Devnet: A deep event network for multimedia event detection and
  evidence recounting.
\newblock In {\em CVPR}, 2015.

\bibitem{gao2015you}
H.~Gao, J.~Mao, J.~Zhou, Z.~Huang, L.~Wang, and W.~Xu.
\newblock {Are You Talking to a Machine? Dataset and Methods for Multilingual
  Image Question Answering}.
\newblock In {\em NIPS}, 2015.

\bibitem{girshick2014rcnn}
R.~Girshick, J.~Donahue, T.~Darrell, and J.~Malik.
\newblock {Rich Feature Hierarchies for Accurate Object Detection and Semantic
  Segmentation}.
\newblock In {\em CVPR}, 2014.

\bibitem{goodfellow2015explaining}
I.~J. Goodfellow, J.~Shlens, and C.~Szegedy.
\newblock Explaining and harnessing adversarial examples.
\newblock {\em stat}, 2015.

\bibitem{gordon2017iqa}
D.~Gordon, A.~Kembhavi, M.~Rastegari, J.~Redmon, D.~Fox, and A.~Farhadi.
\newblock Iqa: Visual question answering in interactive environments.
\newblock {\em arXiv preprint arXiv:1712.03316}, 2017.

\bibitem{he_cvpr15}
K.~He, X.~Zhang, S.~Ren, and J.~Sun.
\newblock Deep residual learning for image recognition.
\newblock In {\em CVPR}, 2016.

\bibitem{hoiem2012diagnosing}
D.~Hoiem, Y.~Chodpathumwan, and Q.~Dai.
\newblock {Diagnosing Error in Object Detectors}.
\newblock In {\em ECCV}, 2012.

\bibitem{jackson_expertsys}
P.~Jackson.
\newblock {\em Introduction to Expert Systems}.
\newblock Addison-Wesley Longman Publishing Co., Inc., Boston, MA, USA, 3rd
  edition, 1998.

\bibitem{jia2014caffe}
Y.~Jia, E.~Shelhamer, J.~Donahue, S.~Karayev, J.~Long, R.~Girshick,
  S.~Guadarrama, and T.~Darrell.
\newblock {Caffe: Convolutional Architecture for Fast Feature Embedding}.
\newblock In {\em ACM MM}, 2014.

\bibitem{JohnsCVPR2015}
E.~Johns, O.~Mac~Aodha, and G.~J. Brostow.
\newblock {Becoming the Expert - Interactive Multi-Class Machine Teaching}.
\newblock In {\em CVPR}, 2015.

\bibitem{johnson_cvpr16}
J.~Johnson, A.~Karpathy, and L.~Fei-Fei.
\newblock {DenseCap: Fully Convolutional Localization Networks for Dense
  Captioning}.
\newblock In {\em CVPR}, 2016.

\bibitem{karpathy_imagenet}
A.~Karpathy.
\newblock {What I learned from competing against a ConvNet on ImageNet}.
\newblock
  http://karpathy.github.io/2014/09/02/what-i-learned-from-competing-against-a-convnet-on-imagenet/,
  2014.

\bibitem{karpathy2015deep}
A.~Karpathy and L.~Fei-Fei.
\newblock Deep visual-semantic alignments for generating image descriptions.
\newblock In {\em CVPR}, 2015.

\bibitem{seed_eccv16}
A.~Kolesnikov and C.~H. Lampert.
\newblock Seed, expand and constrain: Three principles for weakly-supervised
  image segmentation.
\newblock In {\em ECCV}, 2016.

\bibitem{krizhevsky_nips12}
A.~Krizhevsky, I.~Sutskever, and G.~E. Hinton.
\newblock Imagenet classification with deep convolutional neural networks.
\newblock In {\em NIPS}, 2012.

\bibitem{lin2013network}
M.~Lin, Q.~Chen, and S.~Yan.
\newblock Network in network.
\newblock In {\em ICLR}, 2014.

\bibitem{Lin_2014}
T.-Y. Lin, M.~Maire, S.~Belongie, J.~Hays, P.~Perona, D.~Ramanan,
  P.~Doll{\'a}r, and C.~L. Zitnick.
\newblock Microsoft coco: Common objects in context.
\newblock In {\em ECCV}. 2014.

\bibitem{lipton_arxiv16}
Z.~C. {Lipton}.
\newblock {The Mythos of Model Interpretability}.
\newblock {\em ArXiv e-prints}, June 2016.

\bibitem{long2015fcn}
J.~Long, E.~Shelhamer, and T.~Darrell.
\newblock Fully convolutional networks for semantic segmentation.
\newblock In {\em CVPR}, 2015.

\bibitem{Lu2015}
J.~Lu, X.~Lin, D.~Batra, and D.~Parikh.
\newblock {Deeper LSTM and normalized CNN Visual Question Answering model}.
\newblock \url{https://github.com/VT-vision-lab/VQA_LSTM_CNN}, 2015.

\bibitem{Lu2016}
J.~Lu, J.~Yang, D.~Batra, and D.~Parikh.
\newblock Hierarchical question-image co-attention for visual question
  answering.
\newblock In {\em NIPS}, 2016.

\bibitem{mahendran16eccv}
A.~Mahendran and A.~Vedaldi.
\newblock Salient deconvolutional networks.
\newblock In {\em European Conference on Computer Vision}, 2016.

\bibitem{mahendran2016visualizing}
A.~Mahendran and A.~Vedaldi.
\newblock Visualizing deep convolutional neural networks using natural
  pre-images.
\newblock {\em International Journal of Computer Vision}, pages 1--23, 2016.

\bibitem{malinowski_iccv15}
M.~Malinowski, M.~Rohrbach, and M.~Fritz.
\newblock Ask your neurons: A neural-based approach to answering questions
  about images.
\newblock In {\em ICCV}, 2015.

\bibitem{oquab_cvpr14}
M.~Oquab, L.~Bottou, I.~Laptev, and J.~Sivic.
\newblock Learning and transferring mid-level image representations using
  convolutional neural networks.
\newblock In {\em CVPR}, 2014.

\bibitem{oquab_cvpr15}
M.~Oquab, L.~Bottou, I.~Laptev, and J.~Sivic.
\newblock Is object localization for free? – weakly-supervised learning with
  convolutional neural networks.
\newblock In {\em CVPR}, 2015.

\bibitem{pinheiro2015image}
P.~O. Pinheiro and R.~Collobert.
\newblock From image-level to pixel-level labeling with convolutional networks.
\newblock In {\em CVPR}, 2015.

\bibitem{ren_nips15}
M.~Ren, R.~Kiros, and R.~Zemel.
\newblock Exploring models and data for image question answering.
\newblock In {\em NIPS}, 2015.

\bibitem{lime_sigkdd16}
M.~T. Ribeiro, S.~Singh, and C.~Guestrin.
\newblock {"Why Should I Trust You?": Explaining the Predictions of Any
  Classifier}.
\newblock In {\em SIGKDD}, 2016.

\bibitem{niwt}
R.~R. Selvaraju, P.~Chattopadhyay, M.~Elhoseiny, T.~Sharma, D.~Batra,
  D.~Parikh, and S.~Lee.
\newblock Choose your neuron: Incorporating domain knowledge through
  neuron-importance.
\newblock In {\em Proceedings of the European Conference on Computer Vision
  (ECCV)}, pages 526--541, 2018.

\bibitem{hint}
R.~R. Selvaraju, S.~Lee, Y.~Shen, H.~Jin, S.~Ghosh, L.~Heck, D.~Batra, and
  D.~Parikh.
\newblock Taking a hint: Leveraging explanations to make vision and language
  models more grounded.
\newblock In {\em Proceedings of the International Conference on Computer
  Vision (ICCV)}, 2019.

\bibitem{silver2016mastering}
D.~Silver, A.~Huang, C.~J. Maddison, A.~Guez, L.~Sifre, G.~Van Den~Driessche,
  J.~Schrittwieser, I.~Antonoglou, V.~Panneershelvam, M.~Lanctot, et~al.
\newblock Mastering the game of go with deep neural networks and tree search.
\newblock {\em Nature}, 529(7587):484--489, 2016.

\bibitem{simonyan_arxiv13}
K.~Simonyan, A.~Vedaldi, and A.~Zisserman.
\newblock Deep inside convolutional networks: Visualising image classification
  models and saliency maps.
\newblock {\em CoRR}, abs/1312.6034, 2013.

\bibitem{simonyan_arxiv14}
K.~Simonyan and A.~Zisserman.
\newblock {Very Deep Convolutional Networks for Large-Scale Image Recognition}.
\newblock In {\em ICLR}, 2015.

\bibitem{springenberg_arxiv14}
J.~T. Springenberg, A.~Dosovitskiy, T.~Brox, and M.~A. Riedmiller.
\newblock {Striving for Simplicity: The All Convolutional Net}.
\newblock {\em CoRR}, abs/1412.6806, 2014.

\bibitem{szegedy2016rethinking}
C.~Szegedy, V.~Vanhoucke, S.~Ioffe, J.~Shlens, and Z.~Wojna.
\newblock Rethinking the inception architecture for computer vision.
\newblock In {\em Proceedings of the IEEE conference on computer vision and
  pattern recognition}, pages 2818--2826, 2016.

\bibitem{vinyals_cvpr15}
O.~Vinyals, A.~Toshev, S.~Bengio, and D.~Erhan.
\newblock Show and tell: A neural image caption generator.
\newblock In {\em CVPR}, 2015.

\bibitem{vondrick_iccv13}
C.~Vondrick, A.~Khosla, T.~Malisiewicz, and A.~Torralba.
\newblock {HOGgles: Visualizing Object Detection Features}.
\newblock {\em ICCV}, 2013.

\bibitem{zeiler_eccv14}
M.~D. Zeiler and R.~Fergus.
\newblock {Visualizing and understanding convolutional networks}.
\newblock In {\em ECCV}, 2014.

\bibitem{zhang2016top}
J.~Zhang, Z.~Lin, J.~Brandt, X.~Shen, and S.~Sclaroff.
\newblock {Top-down Neural Attention by Excitation Backprop}.
\newblock In {\em ECCV}, 2016.

\bibitem{zhou_cvpr16}
B.~Zhou, A.~Khosla, L.~A., A.~Oliva, and A.~Torralba.
\newblock {Learning Deep Features for Discriminative Localization.}
\newblock In {\em CVPR}, 2016.

\bibitem{Zhou2014ObjectDE}
B.~Zhou, A.~Khosla, {\`A}.~Lapedriza, A.~Oliva, and A.~Torralba.
\newblock Object detectors emerge in deep scene cnns.
\newblock {\em CoRR}, abs/1412.6856, 2014.

\bibitem{zhou2017places}
B.~Zhou, A.~Lapedriza, A.~Khosla, A.~Oliva, and A.~Torralba.
\newblock Places: A 10 million image database for scene recognition.
\newblock {\em IEEE Transactions on Pattern Analysis and Machine Intelligence},
  2017.

\end{thebibliography}
}





\end{document}